%% file: main.tex
\RequirePackage{fix-cm}
\documentclass{lalaworld}

\input{sec/config}
\newcommand{\ours}{{LaLa-World}}
\newcommand{\ourpolicy}{{LaLa-Policy}}
\newcommand{\WebsiteURL}{https://june-page.github.io/lala-world/}
\newcommand{\LaLaWorldCodeURL}{https://github.com/junha1125/LaLa-World\#LaLa-World}
\newcommand{\LaLaPolicyCodeURL}{https://github.com/junha1125/LaLa-World\#LaLa-Policy}
\newcommand{\resourceurl}[2]{\href{#1}{\nolinkurl{#2}}}
\newcommand{\webicon}{\raisebox{-1.5pt}{\includegraphics[height=1.05em]{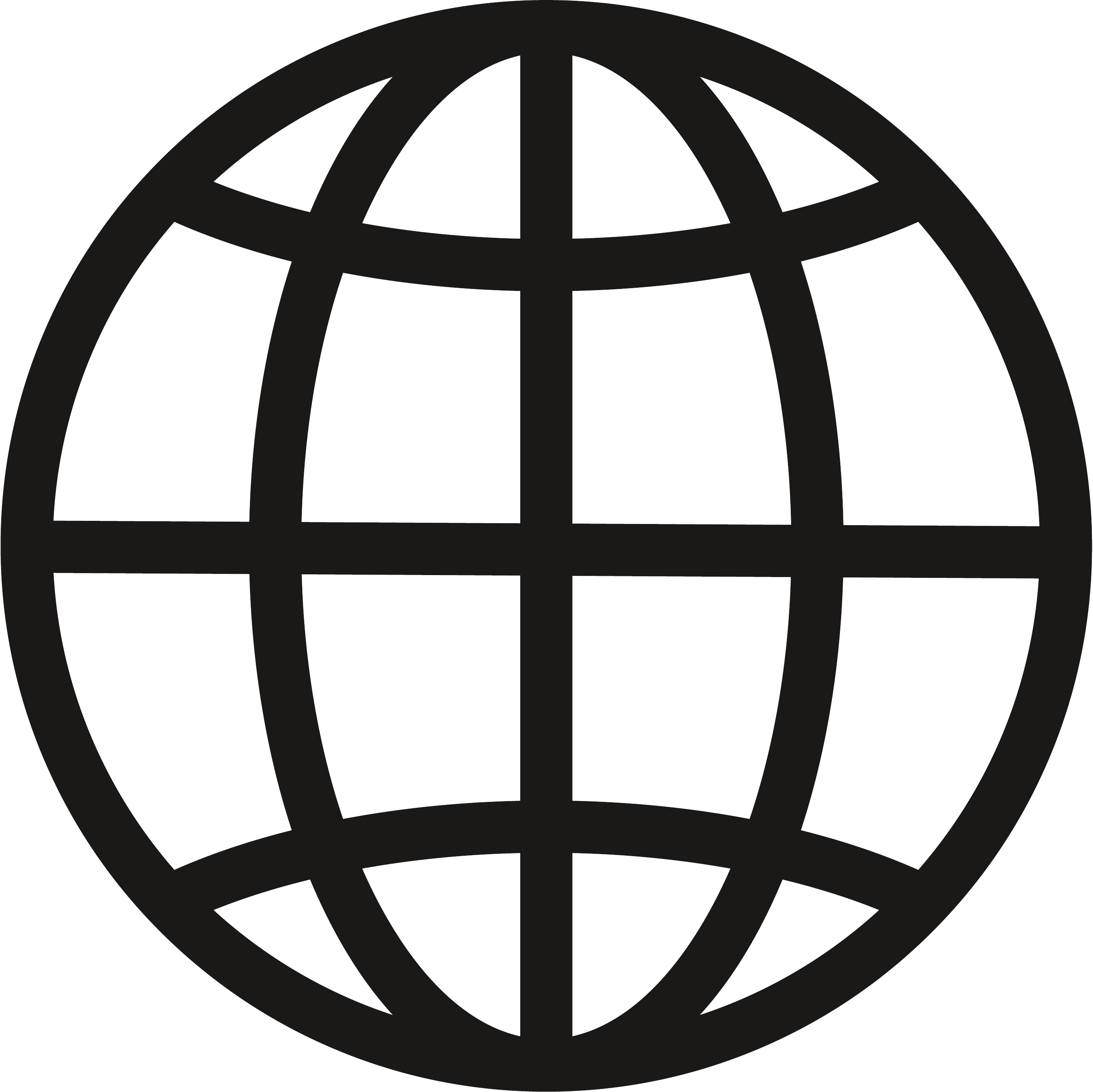}}}
\newcommand{\codeicon}{\raisebox{-1.5pt}{\includegraphics[height=1.05em]{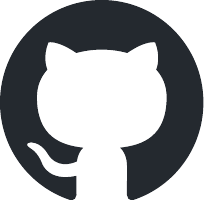}}}

\title{\ours: Scaling \textbf{La}tent \textbf{World} Models\\with Natural \textbf{La}nguage Actions}
\input{sec/author}
\hypersetup{
  pdftitle={LaLa-World: Scaling Latent World Models with Natural Language Actions},
  pdfauthor={Junha Song, Geonmo Gu, Jeongeun Park, Heejeong Nam, Lucas Maes, Byeongho Heo, Dongyoon Han, Sangdoo Yun}
}

\begin{document}
\maketitle

\begin{abstract}
This paper shows that JEPA-style latent world models (LWMs) can learn across diverse domains and data sources by adopting the language-conditioned interface of video generation models.
We establish this approach as an effective and efficient basis for planning with imagination and downstream policy fine-tuning.
To this end, we introduce \textbf{LaLa-World}, a \textbf{La}nguage-conditioned \textbf{La}tent \textbf{World} model that learns from simulation and robot trajectories together with broader image and video data.
Two designs make this work.
First, we describe both numerical actions and visual changes in language, enabling joint training of a single future predictor across heterogeneous datasets.
Second, we build the predictor on a language model, allowing it to interpret language conditions while predicting future latent states.
The resulting predictor achieves higher average MPC planning performance than existing LWMs trained separately for each domain or task.
Adding image and video data further improves average planning performance and robustness to action phrasing.
In a matched comparison with an RAEv2-based video generation model baseline, LaLa-World reaches comparable planning performance with $6\times$ less training cost and $13.9\times$ faster planning.
Beyond planning, the pretrained \ours\ serves as an initialization for \textbf{LaLa-Policy}, which jointly predicts actions and future visual representations.
LaLa-Policy performs competitively with Cosmos Policy on LIBERO and RoboCasa, with up to $5.5\times$ faster inference.
\end{abstract}

\begin{center}
  \renewcommand{\arraystretch}{1.5}
  \setlength{\tabcolsep}{4pt}
  \begin{tabular}{@{}rll@{}}
    \webicon & \textbf{Website} & \resourceurl{\WebsiteURL}{github.io/lala-world}\\
    \codeicon & \textbf{LaLa-World Code} & \resourceurl{\LaLaWorldCodeURL}{LaLa-World\#LaLa-World}\\
    \codeicon & \textbf{LaLa-Policy Code} & \resourceurl{\LaLaPolicyCodeURL}{LaLa-World\#LaLa-Policy}
  \end{tabular}
\end{center}

\clearpage
\tableofcontents
\clearpage

\section{Introduction}
A central capability of an intelligent agent is to anticipate how the world will change in response to its actions. 
Before pushing a cup across a table, we know where it will end up and whether it might fall. 
World models capture this capability by predicting future states under candidate actions, providing a basis for planning or decision-making~\citep{dyna,ha2018world,lecun2022path,dreamer}. 
A promising approach to making this predictive capability computationally efficient is to model how visual representations change under actions, without learning to reconstruct images at the pixel level.
This motivates our focus on JEPA-style latent world models (LWMs), such as V-JEPA 2-AC~\citep{v-jepa-2} and DINO-WM~\citep{dino-wm}.\footnote{Throughout this paper, LWMs refer to JEPA-style world models trained without DreamerV3's reconstruction loss~\citep{DreamerV3} or TD-MPC2's reward supervision~\citep{TD-MPC-2}.\vspace{-2em}}

\input{figure/teaser}

This efficiency of LWMs is particularly useful for planning with imagination in the latent space, which requires repeatedly evaluating candidate action sequences, by rolling out multiple candidate futures at low cost~\citep{TD-MPC}.
However, despite this advantage, the action-conditioned predictors in such LWMs have largely been trained on simulation or robotics datasets, where numerical actions have domain-specific dimensions and physical meanings. 
For example, \prompt{$[2,4]$} may encode horizontal and vertical displacement in one domain and \prompt{$[4]$} force in another, complicating the learning of shared dynamics. 
Furthermore, image and video data, which provide a rich source of observations of the world, often lack these numerical action annotations, making them difficult to use for training action-conditioned future predictors.
Therefore, scaling LWMs across diverse control actions and broader image and video data requires a shared interface that accommodates heterogeneous data. 

We suggest that language can serve as the shared interface. 
We introduce \ours, a language-conditioned LWM that learns from diverse control datasets and broader image and video data with a single future predictor.
Language conditioning already supports learning from diverse visual data in video generation~\citep{wan,cosmos-predict}.
We use language as a shared interface for numerical actions and visual changes
and can thereby scale LWM training with a single predictor across diverse contexts and actions.

As illustrated in \cref{fig:teaser}, \ours\ has two main designs. 
First, we convert numerical actions in simulation or robotics datasets into natural language, allowing them to be expressed in the same form, such as \prompt{move 2 left and 4 up} and \prompt{push with force 2 downward}. We combine these control trajectories with image and video transitions described in language to help the predictor relate descriptions to visual changes across diverse scenes. 
Second, we equip the future predictor with language understanding. Built on a language model, it interprets language conditions while predicting future latent states.
These designs allow one model to learn from different action spaces and data sources through a shared prediction objective.
\cref{fig:lang-planning} shows that a single \ours\ predictor serves multiple domains and predicts future states under qualitative actions such as \prompt{move slightly left}.

Our experiments show that our jointly trained predictor achieves higher average MPC planning performance than existing LWM baselines trained separately for each domain.
We also show that adding image and video data to training further improves planning performance and robustness to action phrasing. 
We compare \ours\ with an RAEv2-based generative world model~\citep{raev2}, whose predictor uses a diffusion Transformer (DiT).
With the same language conditions, training data, and visual encoder, \ours\ outperforms this RAEv2 baseline in planning with $6\times$ less training cost and $13.9\times$ faster inference.

\input{figure/lang-planning}

Beyond planning, we investigate whether the pretrained LWM can serve as a useful initialization for policy fine-tuning.
Cosmos Policy~\citep{cosmos-policy} demonstrates this use of a pretrained video model by adapting Cosmos-Predict2~\citep{cosmos-predict2} to predict robot actions together with future \textit{frames}.
Similarly, we fine-tune \ours\ into \ourpolicy\ for joint prediction of actions and future visual \textit{representations}.
On LIBERO and RoboCasa, \ourpolicy\ performs competitively with Cosmos Policy while reducing average inference time per trajectory by a factor of up to $5.5$.
We hope these results draw greater attention to latent world models, showing that even without visually compelling outputs, they can offer practical advantages over generative world models in planning and downstream policy learning.

We summarize our contributions as follows:

\begin{itemize}[leftmargin=1.2em, labelsep=0.3em, itemsep=0.0em, topsep=-0.2em]
    \item We introduce \ours, a latent world model that uses language as a shared interface for joint training on control trajectories and broader image and video data.
    \item  We show that a single \ours\ predictor learns dynamics across domains and outperforms existing LWMs in MPC planning.
    \item We compare \ours\ with an RAEv2-based generative world model under matched settings, demonstrating higher planning performance and greater training and inference efficiency.
    \item We fine-tune the pretrained \ours\ into \ourpolicy, which performs competitively with existing policy models while achieving faster inference than Cosmos Policy.
\end{itemize}

\section{Background and Motivation}
\label{sec:background}

World models aim to predict how an environment evolves under a given condition~\citep{ha2018world,lecun2022path, wan, cosmos-predict}. 
One approach to making future prediction computationally efficient is to predict changes in latent representations.
\hlv{Latent World Models} (LWMs), including V-JEPA 2-AC~\citep{v-jepa-2} and DINO-WM~\citep{dino-wm}, follow this approach.
These models commonly use numerical actions in the following action-conditioned formulation:
\begin{equation}
e_t = E_\theta(x_t), \qquad \hat{e}_{t+1} = P_\phi\left(e_{t-w+1:t},c_{t-w+1:t}\right), \quad c_t \in \mathbb{R}^k,
\label{eq:lwm}
\end{equation}
where $x_t$ and $c_t$ denote the observation and $k$-dimensional numerical action condition at time $t$, respectively, $E_\theta$ encodes observations into latent representations, and $P_\phi$ predicts the future latent $\hat{e}_{t+1}$ from a context window of length $w$. 
The action-conditioned predictors in these models are commonly trained on simulation or robotics trajectories and evaluated through planning in a domain-specific manner~\citep{pldm, leworldmodel, stableworldmodel}

\hlv{Video generation models} (\ie, generative world models) are trained to synthesize images or videos conditioned on language~\citep{sora,wan,cosmos-predict}. 
This language interface allows them to learn visual changes across domains and data sources. 
Through this training, these models acquire world knowledge that extends their capabilities beyond video generation. Their pretrained weights, for instance, serve as a prior for action prediction: Cosmos Policy fine-tunes Cosmos-Predict2 into a robot policy~\citep{cosmos-policy,cosmos-predict2}.

For \hlv{LWMs}, however, learning jointly from heterogeneous control datasets and broader visual data remains underexplored, as does their use in downstream policy learning.
We therefore explore an approach that retains the efficiency of LWMs while adopting the flexible language interface of generative world models.
The resulting model supports evaluation against existing LWMs in planning and against generative world models in both planning and policy fine-tuning.

\section{\ours}
\label{sec:method}

We introduce \textbf{\ours}, a \textbf{La}nguage-conditioned \textbf{La}tent \textbf{World} model whose predictor takes a language condition $z$ instead of the numerical action condition $c$ in \cref{eq:lwm}.
Two components are needed: transition data that describes the change from $x_{t}$ to $x_{t+1}$ in language, and a predictor that understands it.
\cref{sec:lang-data} covers the first, describing how we construct such language-conditioned data, and \cref{sec:model-arch-objective} covers the second with the architecture and training objective.

\input{figure/data-examples}

\subsection{Language-Conditioned Transition Data}
\label{sec:lang-data}

\paragraph{\hlr{Language unifies the action spaces of multiple domains.}}
Simulation and robotics datasets used to train LWMs record actions as numerical vectors. 
We convert these vectors into language conditions using fixed templates, as illustrated in \cref{fig:data-examples}, retaining each domain's numerical values with their direction and magnitude. 
Language thus expresses actions across domains in a common form, allowing their control data to be pooled to train a single predictor.
We apply this conversion to seven simulation and robotics datasets described in \cref{sec:mpc-setup}.
The actions in these domains may differ in physical meaning and dimensionality.
This raises the key question of whether a single future predictor can learn dynamics across these domains.
Details of the dataset composition and conversion templates are provided in \cref{sec:details-training-data}.

\paragraph{\hlr{Language enables action-conditioned training on broader image and video data.}}
We also train the same future predictor on image and video transitions without numerical action annotations, using descriptions of visual change as conditions.
Such data matter because the converted control datasets describe actions with fixed templates, which exposes the predictor to only a narrow range of language.
In addition, descriptions of visual change provide language conditions for these transitions, extending training beyond the control trajectories. 
We therefore supplement them with captioned videos from OpenVid~\citep{Openvid} and image-editing pairs with instructions from OmniEdit~\citep{Omniedit} and ShareGPT~\citep{sharegpt}.
In these data, the caption describing the change between frames or the editing instruction serves directly as the language condition, so no numerical action annotation is needed.
The predictor thereby sees a wider range of language and visual change.

\subsection{Model Architecture and Objective}
\label{sec:model-arch-objective}

As shown in \cref{fig:teaser}(b), \ours\ follows the encoder-predictor framework of existing LWMs, with simple adaptations for language conditioning. Below, we describe the observation encoder (\ie, vision encoder), transition predictor (\ie, future predictor), and training objective.

\paragraph{\hly{Encoder.}}
LWMs use an observation encoder to map each RGB frame $x_t$ to a latent representation $e_t$~\citep{v-jepa-2, jepa-wms}. For this role, vision transformers pretrained with self-supervised objectives are commonly used, including DINO encoders~\citep{dinov2, dinov3} and V-JEPA~2 encoder~\citep{v-jepa-2}. These encoders are pretrained on large collections of unlabeled images and videos, providing visual representations for diverse environments~\citep{dino-wm}. 
In V-JEPA 2-AC, the encoder remains frozen during world model training, which prevents encoder collapse and preserves its pretrained visual representations. 
Following JEPA-WMs~\citep{jepa-wms}, we use DINOv2-S/14 or DINOv3-L/16 as the frozen observation encoder.
More details and ablations on vision encoders are provided in \cref{sec:ablation-vision-encoder}.

\paragraph{\hly{Predictor.}}
The predictor $P_\phi$ takes the visual context $e_{t-w+1:t}$ and the language condition $z$ as input and predicts the next latent representation $\hat{e}_{t+1}$.
Existing world models typically receive conditions in two ways: LWMs incorporate numerical actions through mechanisms such as AdaLN~\citep{jepa-wms}, feature conditioning~\citep{dino-wm}, and sequence conditioning~\citep{causal-jepa}, all of which offer no way to read text. Video generation models support language conditioning through a separate text encoder whose outputs condition the predictor~\citep{wan}, which adds parameters and text-encoding cost.

\ours\ addresses both limitations by using a pretrained language model as the predictor itself: it reads the language condition as text and predicts future latent representations in the same network.
Specifically, we use Qwen3.5-0.8B~\citep{qwen35} as our default predictor backbone.
The language condition $z$ is tokenized using the model's tokenizer and mapped to token embeddings, which are passed to the predictor together with the visual latents.
To map the visual latents into the same embedding space, we use a two-layer MLP and add 3D positional embeddings indexed by time and 2D spatial location~\citep{qwen2vl}.
The output $\hat{e}_{t+1}$ is then read out from learned query tokens~\citep{metaqueries}.
We train the predictor by minimizing the squared error between the predicted and target future representations: 
 $\mathcal{L} = \| \hat{e}_{t+1} - e_{t+1} \|^2. \label{eq:training-objective}$

Notably, we use proprioceptive state neither as an input nor as a prediction target.
While proprioception can benefit LWMs tailored to a specific domain, it is defined in a domain-specific manner, which limits its applicability to a domain-agnostic world model.
We further examine the predictor's backbone and architectural choices through ablations in \cref{sec:ablation-architecture} and \ref{sec:details-architecture}, respectively, including BERT~\citep{bert} and Gemma~\citep{gemma3} as alternative backbones.

\input{table/planning-benchmarks}

\section{MPC Planning}
\label{sec:mpc-planning}

Suppose we want to move a cup from our hand onto a table while avoiding nearby objects. We imagine the outcomes of candidate action sequences and select the one that brings the cup closest to the desired position (\ie, planning with imagination). In robot planning, the world model provides these imagined outcomes: given an initial observation $x_{\mathrm{init}}$ and a goal observation $x_{\mathrm{goal}}$, a planner selects the action sequence whose imagined outcome is closest to the goal. Following prior LWM studies~\citep{dino-wm,jepa-wms}, we evaluate planning with model predictive control (MPC), using the Cross-Entropy Method (CEM) to optimize action sequences~\citep{cem-planning}.

\paragraph{Evaluation Setup.}
\label{sec:mpc-setup}
At each planning round, we sample candidate action sequences and, for each sequence, apply \cref{eq:lwm} autoregressively for $H$ steps and score it by the squared distance between the final predicted latent state and the encoded goal:
\begin{equation}
        \mathcal{C}= \left\| \hat{e}_{t+H} - e_{\mathrm{goal}} \right\|^2,
        \quad e_{\mathrm{goal}} = E_\theta(x_{\mathrm{goal}}).
    \label{eq:planning-cost}
\end{equation}
CEM repeatedly samples sequences and updates the sampling distribution using those with the lowest costs.
For \ours, we verbalize candidate actions using the training templates.
In simulation, we execute the corresponding numerical actions from an initial segment of the selected sequence.
For domains requiring re-planning, we obtain a new observation and repeat this process.
Benchmark-specific settings and evaluation criteria are provided in \cref{sec:details-mpc-planning}.

We use seven control datasets from DINO-WM and JEPA-WMs~\citep{jepa-wms}: six simulation datasets--Wall~\citep{dino-wm}, PointMaze~\citep{point-maze}, Push-T~\citep{pusht}, MetaWorld~\citep{meta-world}, Rope and Granular~\citep{rope-granular}--and the real-robot dataset DROID~\citep{droid}.
We also evaluate on RoboCasa~\citep{robocasa}, a domain excluded from training, to assess zero-shot transfer. 
In addition to existing LWM baselines including DINO-WM,
we also train the world models of IRIS~\citep{IRIS}, DreamerV3~\citep{DreamerV3}, and TD-MPC2~\citep{TD-MPC-2} on the same control datasets without reward or task information, which is the same setup for these baselines in DINO-WM.
For a fair comparison, we train each baseline using its public implementation on the same data and evaluate all under the same planning settings, including horizon $H$ and none uses proprioceptive states.

\paragraph{\hly{Decoder.}}
To visualize and interpret predicted futures, we build a 12-layer transformer decoder $D$ that maps predicted latents to RGB images: $\hat{x}_{t+1} = D(\hat{e}_{t+1})$. We train the decoder with a combination of $L_1$ reconstruction and LPIPS perceptual losses~\citep{iplps-loss}, following the decoder training recipe of RAEv2~\citep{raev2}. It should be noted that the decoder is not used during world model training (\cref{sec:model-arch-objective}).

\subsection{Results and Analysis}
\label{sec:mpc-results}

\paragraph{\hlp{A jointly trained LaLa-World predictor performs well across diverse domains.}}
\cref{tab:planning-benchmarks} reports MPC planning results across simulation and robot domains.
While existing LWMs are typically trained within individual domains, \ours\ trains its predictor jointly across all domains with different action spaces and physical dynamics.
Despite these differences, \ours{} achieves the highest average score in both benchmark groups in \cref{tab:planning-benchmarks}.
On the five simulation tasks in the upper table, \ours{} reaches an average success rate of 72.1\%, exceeding JEPA-WMs by 7.5 percentage points.
The largest gain over the best baseline in this group is on Push-T, where success increases from 68.8\% to 81.2\%.
\ours{} also achieves the highest zero-shot success rates on both RoboCasa tasks, reaching 40.6\% on Reach and 46.9\% on Place.

\input{figure/phrasing-robustness}
\paragraph{\hlp{Training on image and video data improves performance and robustness to phrasing.}}
We evaluate the effect of training \ours\ on image and video transitions and find two benefits.
The results are shown in \cref{fig:phrasing-robustness}.
First, adding image and video transitions improves planning performance.
Specifically, the average score increases from 59.3 with control trajectories alone to 61.6 with all data.
Second, \ours\ becomes more robust when the same actions are expressed in different words.
To evaluate this, we paraphrase the action templates at evaluation, for example, replacing \prompt{move the gripper} with \prompt{shift the robot hand} and \prompt{move the agent} with \prompt{push the dot}.
\ours\ remains robust when trained on all data, including image and video transitions, whereas paraphrasing lowers the average score by 6.8 points with control trajectories alone.
Further details and domain-level results are provided in \cref{sec:additional-phrasing-lang}.

\paragraph{\hlp{LaLa-World can predict the future from language without numbers.}}
The language actions in our control trajectories retain numerical magnitudes.
We next ask whether \ours\ can predict outcomes from qualitative descriptions such as \prompt{move slightly down}.
Using the action ranges observed in the training data, we define a conversion function that replaces each numerical magnitude with one of three qualitative levels: \emph{slightly}, no modifier, or \emph{far}.
CEM samples numerical action sequences as before and passes them through this function to obtain the language conditions for \ours.
The bottom two rows of \cref{fig:lang-planning} show successful planning examples in Wall and RoboCasa under these conditions.
These examples indicate that the predictor can use qualitative descriptions to guide planning even when exact action magnitudes are unspecified.
We provide more details on the evaluation setup in \cref{sec:details-quali-lang} and additional results in \cref{sec:additional-quali-lang}.

\subsection{Latent vs. Generative World Models}
\label{sec:latent-vs-pixel}
With language as the action interface, \ours\ can be compared with generative world models trained on the same kinds of data. LWMs learn to match future visual latents, whereas generative world models are trained for pixel-level generation. We ask how these approaches compare in planning performance and computational cost. 
To the best of our knowledge, this is the first comparison of latent and generative world models in planning under matched conditions, including language conditions and training data, as discussed in \cref{sec:rel-work-video-gen}.

\paragraph{Setup for a matched comparison.}
We use RAEv2~\citep{raev2} as our generative world model baseline because it uses DINOv3-L/16~\citep{dinov3} rather than a VAE as its vision encoder.
\ours\ uses the same encoder in this comparison.
Both models use language as the action interface and receive the same language conditions.
We train both models on the same control trajectories and image and video transitions used to train \ours.
The full-depth predictors are comparable in size: \ours's predictor and RAEv2's DiT~\citep{dit} each contain approximately 0.8B parameters.
We initialize RAEv2 from its publicly released pretrained DiT weights and follow its training implementation, including flow matching~\citep{flowmatching} and REPA~\citep{repa}.
During planning, we use one or five denoising steps per prediction and compute the cost in \cref{eq:planning-cost} between the resulting future representations and the encoded goal.
Further details of RAEv2 are provided in \cref{sec:details-raev2}.

We compare the two world models in terms of training cost and planning time.
For training cost, we track planning performance while training the full-depth predictors on eight H100 GPUs with a 12-hour budget.
For planning time, we vary the number of predictor layers to obtain four model sizes for each approach, train each variant on 16M transitions in total, and measure average planning time per trajectory on a single H100 GPU.
Both experiments follow the planning setup in \cref{sec:mpc-setup}.

\input{figure/latent-vs-pixel}
\input{figure/imagination-visualization}

\paragraph{\hlp{LaLa-World outperforms the generative baseline with lower training and planning costs.}}
As shown in \cref{fig:latent-vs-pixel}(a), \ours\ reaches five-step RAEv2's 12-hour performance in two hours and later achieves a planning performance of 59.4\%, exceeding RAEv2's best performance of 48.0\%.
For RAEv2, using five denoising steps per prediction improves performance but increases planning time.
In \cref{fig:latent-vs-pixel}(b), full-depth \ours\ outperforms every RAEv2 variant and plans $2.9\times$ and $13.9\times$ faster than full-depth RAEv2 with one and five denoising steps, respectively.
\footnote{Evaluating RAEv2 with 50 denoising steps would require approximately 25 days on 8 H100 GPUs.\vspace{-3.em}}
Together, these results show that a world model trained to match future visual latents can support effective and efficient planning without targeting the reconstruction of visual details.
Domain-level results are provided in \cref{sec:additional-latent-vs-generative}.
\cref{fig:imagination-visualization} shows imagined DROID rollouts under the selected language actions: RAEv2 generates plausible future observations even with one denoising step, while decoded \ours\ predictions also depict plausible future states.

\section{\ourpolicy: From World Model To Action Predictor}
\label{sec:lala-policy}
We now ask whether this pretrained \ours\ provides a useful initialization for policy fine-tuning. Cosmos Policy~\citep{cosmos-policy} takes this approach by fine-tuning the pretrained video generation model Cosmos-Predict2~\citep{cosmos-predict2} to jointly predict actions and future frames.

\paragraph{\hly{LaLa-Policy.}}
Following Cosmos Policy's fine-tuning procedure, we fine-tune \ours\ into \ourpolicy\ using policy data to predict an action chunk together with a future latent representation.
We add learned action queries to the predictor $P_\phi$ alongside its queries for future prediction.
Given an observation $x_t$ and a task instruction $\ell$, the outputs at the action queries provide a sequence of conditioning features $h_{t:t+N-1}$ for an action generator $A_\psi$ built on a diffusion transformer.
We use the action-head architecture of Isaac-GR00T~\citep{issac-gr00t}, which conditions action generation on vision-language features.
Starting from Gaussian noise, $A_\psi$ generates the action chunk through iterative denoising:
\begin{equation}
    \begin{aligned}
        \left(\hat{e}_{t+N}, h_{t:t+N-1}\right) &= P_\phi\left(E_\theta(x_t), \ell\right), \\
        \hat{a}_{t:t+N-1} &= A_{\psi}\left(\epsilon; h_{t:t+N-1}\right), \quad \epsilon \sim \mathcal{N}(0,I).
    \end{aligned}
    \label{eq:lala-policy}
\end{equation}
Here, $N$ is the number of actions in the chunk, and the future prediction targets $E_\theta(x_{t+N})$.
Modeling an action distribution allows the policy to represent multiple action sequences for reaching the goal.

\paragraph{Policy fine-tuning setup.}
We compare \ourpolicy\ with policies that predict actions alone~\citep{intelligence2025pi, li2025cogvla} and models that jointly predict actions and future observations~\citep{li2025unified, liang2025video}.
For a fair comparison with the joint prediction baselines, we fine-tune Cosmos Policy~\citep{cosmos-policy} to predict only actions and future observations, without auxiliary future proprioception or value prediction.
\ourpolicy\ follows the same prediction setup in latent space.
We fine-tune on LIBERO~\citep{liu2024libero} and RoboCasa~\citep{robocasa}, using demonstration trajectories composed of transitions $(x_t,a_t,x_{t+1})$ and paired with task instructions.
For RoboCasa, we consider two training sets: 50 human demonstrations or 300 demonstrations generated with MimicGen per task.
For \ourpolicy, we keep the observation encoder frozen and jointly train the pretrained predictor and newly added policy modules.
The action generator is trained from scratch.
Dataset and training details are provided in \cref{sec:details-policy-training-data,sec:details-policy-fine-tuning}.

\paragraph{Evaluation Setup.}
The policy models are evaluated in LIBERO and RoboCasa simulators.
At each step, we execute 16 predicted actions on LIBERO and the first 16 of the 32 predicted actions on RoboCasa.
We report success rates, counting a trial as successful only if it meets the simulator's task-specific success criterion within the allowed number of actions.
More details are in \cref{sec:details-policy-evaluation}

\input{table/policy-benchmarks}
\input{figure/policy-visualization}

\input{table/policy-inference-time}
\paragraph{\hlp{LaLa-Policy achieves competitive and efficient action prediction via latent world modeling.}}
\cref{tab:policy-libero,tab:policy-robocasa} show that \ours\ can be fine-tuned into an effective policy, performing competitively with Cosmos Policy on both benchmarks.
Notably, this performance comes with up to $5.5\times$ faster inference, measured by average inference time per trajectory (\cref{tab:policy-inference-time}).
Together with our planning results, these findings suggest that effective world models and action predictors can be learned without a pixel reconstruction objective.
World modeling pretraining also benefits policy fine-tuning.
Initializing the predictor from \ours\ outperforms both random and Qwen3.5 initialization on both LIBERO and RoboCasa.
This agrees with the finding of Cosmos Policy~\citep{cosmos-policy} that Cosmos-Predict2, pretrained on diverse videos, provides a useful prior for policy learning, and extends this observation to latent world models.

\paragraph{\hlp{LaLa-Policy can anticipate future latent states resulting from its predicted actions.}} 
Given the current observation and a task instruction, 
\ourpolicy\ jointly predicts an action chunk and a future latent representation.
During policy execution in LIBERO and RoboCasa, we decode the predicted representations into images and visualize them alongside simulator observations in \cref{fig:policy-visualization}.
The decoded futures capture both robot motion and changes in object position during lifting and transport, showing that latent prediction can represent the dynamics of manipulation.

\section{Conclusion}
\label{sec:conclusion}

We presented \ours, a latent world model that learns from heterogeneous control trajectories and broader image and video data through a shared language interface.
This enables comparisons with existing LWMs in MPC planning and generative models in planning and policy fine-tuning.
Despite sharing one predictor across domains, \ours\ outperforms existing LWMs in average MPC planning performance.
Under matched settings, including training data, it also surpasses the generative baseline in planning performance and efficiency.
Following Cosmos Policy's use of world models pretrained on diverse data for policy learning, we fine-tuned \ours\ into \ourpolicy.
It outperforms evaluated VLA baselines and achieves substantially faster inference than Cosmos Policy.
These results demonstrate the value of latent world models for planning and policy learning without training to reconstruct visual details.
We hope this work draws greater attention to latent world models, their scalability, and downstream uses.

For broader practical impact, future work could explore scaling training data and model size, extending prediction horizons, and evaluating policies on real robots, as discussed in \cref{sec:future-directions}.

\clearpage
\section*{\normalsize Acknowledgments}
\addcontentsline{toc}{section}{Acknowledgments}
{\small
We gratefully acknowledge NAVER AI Lab for providing the resources and support that facilitated this research.
\par}

{\small
  \phantomsection
  \addcontentsline{toc}{section}{References}
  \bibliographystyle{lalaworld}
  \bibliography{main}
}

\clearpage
\appendix
\addtocontents{toc}{\protect\normalfont\protect\fontsize{10}{12}\protect\selectfont}
\ExplSyntaxOn
\renewcommand\thefigure{\int_to_Alph:n{\value{figure}}}
\ExplSyntaxOff
\setcounter{figure}{0}
\renewcommand\thetable{\Alph{table}}
\setcounter{table}{0}

\section{Related Work}
\label{sec:related-work}

We summarize in \cref{fig:relate-work-table} how \ours\ and \ourpolicy\ compare with representative world models and policy models.

\input{figure/relate-work-table}

\subsection{Latent World Models}
Latent world models predict future states in a learned representation space rather than in pixels. Recent work follows the joint-embedding predictive architecture~\citep{ijepa,v-jepa-2}.
Existing studies focus on improving latent prediction for planning~\citep{v-jepa-2} or stabilizing training~\citep{leworldmodel}.
In the former direction, PLDM~\citep{pldm} demonstrates reward-free planning with learned latent dynamics, while DINO-WM~\citep{dino-wm} shows the effectiveness of using a frozen DINOv2 vision encoder. V-JEPA~2-AC~\citep{v-jepa-2} extends latent planning to real-robot manipulation, and JEPA-WMs~\citep{jepa-wms} study design choices including multistep rollouts. An orthogonal direction addresses training stability: LeWorldModel~\citep{leworldmodel} trains the vision encoder and predictor end to end, using SIGReg to prevent representation collapse. AdaJEPA~\citep{adajepa} further improves robustness by updating the predictor at test time. These methods typically train action-conditioned predictors within a single domain, limiting their development into General World Models. Although Newt~\citep{newt} learns across multiple domains, its training remains focused on control and game environments. 
Joint training of action-conditioned predictors on control trajectories and broader image and video transitions remains underexplored. We therefore use language to unify actions across control datasets and co-train them with diverse image and video transitions. 
Our use of language as the action interface differs from Dynalang~\citep{dynalang}, which incorporates language as observations or task descriptions.

\subsection{Video Generation Models}
\label{sec:rel-work-video-gen}
Video generation models (\ie, generative world models) predict future observations conditioned on language~\citep{ho2022video,blattmann2023align,villegas2022phenaki,kondratyuk2023videopoet}.
Language conditioning lets them learn from web-scale video~\citep{sora,wan}, and the resulting models serve as interactive simulators for games~\citep{genie}, driving~\citep{gaia1}, and robotics~\citep{unisim,cosmos-predict,cosmos-predict2}.
While video generation models can produce visually compelling content, using them for MPC planning is costly because evaluating candidate action sequences requires repeatedly synthesizing the future with visual details.
V-JEPA~2-AC and V-JEPA~2.1~\citep{v-jepa-2,v-jepa-2-1} report faster planning than video generation baselines~\citep{cosmos-predict,nwm}, but these comparisons differ in action interface, encoder, predictor size, and training data.
RAE~\citep{raev1,raev2} replaces the VAE encoder~\citep{vae} with a vision encoder pretrained with self-supervision, allowing a generative world model to share its encoder with an LWM.
Using RAEv2 as a generative world model baseline, we compare it with \ours\ under matched settings, including language conditioning.

\subsection{Planning and Policy Learning with World Models}
Planning with learned dynamics models has a long history~\citep{dyna,ha2018world,e2c,visual-foresight,visual-foresight-2}.
PlaNet~\citep{planet} plans with the Cross-Entropy Method (CEM) in a latent space trained with reconstruction and reward prediction. Dreamer~\citep{dreamer,dreamerv2,DreamerV3,dreamerv4} and IRIS~\citep{IRIS} instead train actor-critic policies on imagined rollouts. MuZero~\citep{Muzero} and TD-MPC~\citep{TD-MPC,TD-MPC-2} learn latent dynamics with reward and value prediction to support tree search and model predictive control, respectively.
More recent work defines a world model more narrowly as a predictor of future states~\citep{pldm,v-jepa-2,leworldmodel}.
Planning with them is performed at test time by sampling-based optimization toward a goal image (\ie, MPC planning).
Our experiments follow this reward-free and goal-conditioned protocol.

\subsection{Action Prediction Models}
Vision-language-action (VLA) models such as RT-2~\citep{brohan2023rt}, OpenVLA~\citep{openvla}, $\pi_{0.5}$~\citep{intelligence2025pi}, UniVLA~\citep{bu2025univla}, and CogVLA~\citep{li2025cogvla} fine-tune vision-language models on large-scale robot demonstrations to map an observation and an instruction to actions.
These models predict actions alone. When humans want to move a cup, however, they can imagine the action, how the cup will move, and where it will end up.
Video-based policies such as UniPi~\citep{unipi} and AVDC~\citep{avdc} first generate videos of task execution and then translate them into control actions. Unified video-action models~\citep{li2025unified,zhu2025unified} instead predict actions and future frames jointly.
Building on pretrained video models, Cosmos Policy~\citep{cosmos-policy} fine-tunes Cosmos-Predict2~\citep{cosmos-predict2} to jointly predict actions, future states, and values.
\ourpolicy\ follows this recipe but starts from a latent world model, predicting future representations rather than reconstructing visual details.

\subsection{Visual Representation Pretraining}
Self-supervised visual encoders provide the representations on which latent world models learn dynamics~\citep{dino-wm,v-jepa-2}.
Contrastive and self-distillation methods~\citep{moco,simclr,dino,dinov2,dinov3} learn invariances across augmented views, masked image modeling~\citep{mae,ibot} predicts masked patches, and JEPAs~\citep{ijepa,v-jepa} predict the representations of masked regions instead of their pixels.
These encoders are conventionally evaluated by probing or fine-tuning on recognition tasks such as classification, segmentation, and depth estimation.
V-JEPA~2~\citep{v-jepa-2} introduced planning as a further evaluation: the pretrained encoder is frozen, an action-conditioned predictor is trained on top of it from robot trajectories, and the pair is used for MPC planning toward image goals.
We follow the same evaluation protocol as V-JEPA~2, DINO-WM, and JEPA-WMs~\citep{jepa-wms}.

\section{Future Directions}
\label{sec:future-directions}

\ours\ shows that a shared language interface enables joint LWM training on heterogeneous control trajectories and broader image and video data, with benefits for planning and policy fine-tuning. These results suggest several directions for extending the current approach.

\begin{enumerate}[leftmargin=15pt, itemsep=4pt, topsep=0pt, parsep=0pt]
    \item \textbf{Scaling data and model capacity.} With a 0.8B predictor, adding image and video data improves planning performance and robustness to action phrasing (\cref{fig:phrasing-robustness}). However, our training data remain smaller in scale than those of video generation models such as Wan~\citep{wan}, and we have not established whether these gains continue at larger scales. Future work should increase dataset size and predictor capacity separately and together, measuring their effects on planning performance and downstream policy learning.
    \item \textbf{Learning visual representations end to end.} We freeze the vision encoder to prevent representation collapse, but this also prevents its representations from adapting to future prediction. Better world models may require visual representations that capture information useful for anticipating future states. Jointly training the vision encoder and predictor while preventing collapse is therefore an important direction for future work.
    \item \textbf{Extending prediction beyond one step.} Video generation models produce sequences of future frames, whereas our predictor is trained with supervision only on the next latent state. Whether latent world models can learn to generate similarly long sequences through autoregressive prediction remains an open question. Addressing this would require training on autoregressive rollouts with losses over multiple future states, rather than only the next state. We leave this investigation to future work because of the additional training cost, with particular interest in how prediction quality changes as the rollout length increases.
    \item \textbf{Extending prediction horizons.} Following the evaluation protocols of JEPA-WMs and DINO-WM~\citep{jepa-wms,dino-wm}, our MPC experiments use relatively short planning horizons of at most six steps (\cref{tab:planning-setup}). Evaluating longer horizons would help assess whether the model can anticipate distant consequences and support planning in real-world tasks that require such foresight. The multi-step supervision discussed above could support this extension.
    \item \textbf{Evaluating on real robots.} Our experiments on LIBERO and RoboCasa show that initializing \ourpolicy\ from \ours\ outperforms existing VLAs, demonstrating the value of a pretrained latent world model for policy fine-tuning. While these simulation experiments establish this benefit, extending the evaluation to real robots is an important next step.
    \item \textbf{Best-of-N planning with \ourpolicy.} \ourpolicy\ achieves competitive action prediction without value supervision. Cosmos Policy~\citep{cosmos-policy} further learns from policy rollouts to predict future states and values for candidate action chunks, enabling best-of-N planning. We consider extending \ourpolicy\ to such planning a valuable avenue for future work.
\end{enumerate}

We hope these directions broaden the use of LWMs while retaining their efficiency in planning and policy learning.

\section{Experiments and Analyses}
\label{sec:experiments-analyses}

\subsection{ImageNet Training and Evaluation of \ours}
\label{sec:imagenet-training-evaluation}

Inspired by Image World Models (IWM)~\citep{iwm}, we use ImageNet~\citep{imagenet} transformations to examine whether \ours\ can predict visual changes specified in language.

\paragraph{Training.}
We construct transitions from a source image, a transformed target view, and a language description of the transformation (\cref{fig:imagenet-train-eval}, left).
Crops, shifts of the view, rotations, and horizontal flips become language actions, allowing static images to enter the same training format as control data.
For example, a crop followed by a horizontal flip is paired with a sentence describing both operations.
The predictor receives the source representation and this description, and learns to match the encoded target view.
We train the predictor jointly on these pairs and other transition data.

\paragraph{Evaluation.}
For each of 200 ImageNet validation images, we construct a bank of 256 transformed views of that image, including photometric distractors with changes in color or brightness.
Given the source and a geometric language action, the predictor must retrieve the corresponding target by the mean cosine similarity of aligned patch representations.
We report mean reciprocal rank (MRR), averaging the reciprocal of the target's rank over queries.
A target ranked first contributes $1$, second $1/2$, and third $1/3$, so higher MRR indicates more accurate retrieval.
Because all candidates show the same underlying image, retrieval requires distinguishing the requested transformation.

\paragraph{Controls.}
The source latent is the encoded input image, used directly for retrieval.
The predicted latents are obtained from \ours\ with either empty language (\texttt{""}) or the language action.

\paragraph{Results.}
\cref{tab:imagenet-mrr} shows that \ours\ reaches 0.856 MRR with language actions, while using the source latent directly or predicting with empty language gives 0.024 and 0.026, respectively.
This gap indicates that the predictor uses language to model the specified visual change, extending the action interface to transformations of static images.

\input{figure/imagenet-train-eval}
\input{table/imagenet-mrr}

\subsection{Communicable World Modeling}
\label{sec:communicable-world-modeling}

CEM-based MPC requires thousands of imagined action sequences, which limits its practicality even when each world-model rollout is inexpensive.
A natural direction is to let a capable LLM act as the decision-maker, using the world model to evaluate its proposed actions.
Since \ours\ reads actions as language, it provides an interface for this interaction.
We therefore ask whether an LLM can replace CEM with fewer imaginations, and whether this reduced search budget still yields successful plans.

\paragraph{Planning procedure.}
We call this setting \emph{communicable world modeling}, in which an LLM actor and \ours\ exchange language actions and their evaluation costs (\cref{fig:communicable-setup}).
The LLM receives the initial and goal observations and proposes $N$ candidate action sequences using the numerical language of the training templates, such as \prompt{move 0.6 left and 1.7 down}.
\ours\ imagines each sequence and returns the cost in \cref{eq:planning-cost}; a lower cost indicates a predicted outcome closer to the goal.
The LLM uses these costs to revise its proposals.
After $K$ rounds, we select the lowest-cost sequence for execution and replan when required by the benchmark.
During this search, the costs are the only feedback: the LLM sees neither the imagined states nor privileged simulator states.

\paragraph{Evaluation.}
We use Claude Opus 5 and GPT-6 Astra as actors and compare their results with the CEM baseline from \cref{tab:planning-benchmarks} in \cref{tab:communicable-world-modeling}.
We count one imagined candidate sequence as one rollout and report the total per trajectory, including replanning.
Success rates measure task completion in simulation, while DROID uses the score defined in \cref{eq:droid-action-score}.

\input{figure/communicable-setup}
\input{table/communicable-world-modeling}

\paragraph{Results.}
\cref{tab:communicable-world-modeling} shows that the LLM actors use 8–96 rollouts per trajectory, compared with 4,500--54,000 for CEM.
Claude Opus 5 reaches 100.0\% success on Wall with 90 rollouts and 53.1\% on RoboCasa-R with 80 rollouts.
\cref{fig:communicable-qualitative} shows successful simulation episodes.
These results demonstrate that an LLM can plan through the language interface of \ours\ with a small imagination budget, while leaving room to improve performance on tasks requiring precise control.

\subsection{\ours\ Architecture Ablations}
\label{sec:ablation-architecture}

We examine how training data and predictor design affect the use of language conditions, beyond what validation loss alone reveals.
A predictor may fit transitions using visual regularities or numerical patterns without responding to the meaning of action words.

\paragraph{Language sensitivity.}
For a held-out transition, we measure prediction accuracy relative to retaining the current state:
\begin{equation}
    S(z) = 1 - \frac{\|\hat{e}_{t+1}(z) - e_{t+1}\|_2}{\|e_t - e_{t+1}\|_2},
    \label{eq:language-sensitivity}
\end{equation}
where $\hat{e}_{t+1}(z)$ is the prediction conditioned on $z$.
The score is $1$ for an exact prediction, $0$ for returning $e_t$, and negative when the prediction is farther from the target than $e_t$.

We keep the target fixed and vary the condition: numerical language, qualitative language without numbers, reversed direction words in numerical language, and swapped magnitude words in qualitative language (\cref{fig:llm-ablation-1}(a)).
Qualitative descriptions preserve direction but coarsen magnitude, so some loss of accuracy is expected.
By contrast, following reversed direction words should move the prediction away from the original target and can make the score negative.
Magnitude swaps preserve direction but change the amount of motion, so we expect a smaller drop.
Panel (b) illustrates this expected pattern.

\paragraph{Training data.}
With control data alone, the score drops when numerical magnitudes are replaced by qualitative descriptions (\cref{fig:llm-ablation-1}(c)).
Adding image and video transitions narrows this gap and makes the direction-swap score more negative.
These results suggest that broader transition data help the predictor interpret qualitative magnitudes and respond to direction words.

\input{figure/communicable-qualitative}

\paragraph{Language backbone.}
We train BERT-0.3B~\citep{bert}, Gemma-0.3B~\citep{gemma3}, Nemotron-1B~\citep{blakeman2025nvidia}, and Qwen3.5-0.8B~\citep{qwen35} on the same data.
Qwen3.5 most closely follows the expected sensitivity pattern, while Gemma shows a similar pattern with lower scores under the original actions (\cref{fig:llm-ablation-1}(d)).
Nemotron reaches a validation loss close to Qwen3.5 (\cref{fig:llm-ablation-2}(a)), yet its score barely changes when direction words are reversed.
Thus, low prediction loss alone does not establish that a predictor follows action semantics.
Pretrained Qwen3.5 also reaches a lower loss earlier in training than its randomly initialized counterpart, although their final losses are similar.

\paragraph{Positional encoding and readout.}
We compare the positional encoding and readout designs shown in \cref{fig:arch-examples}.
We measure language reliance as $\Delta S = S(z) - S(\varnothing)$, the score drop when the condition is replaced by empty text.
A larger gap indicates that language contributes more to prediction accuracy.
The 3D input positional embeddings index time and spatial location, preserving the structure of the visual tokens.
In \cref{fig:llm-ablation-2}(b), 3D input positional embeddings with learned query tokens yield the highest reliance among the tested designs.
Reading predictions from the last frame's state tokens instead of learned queries reduces reliance under both 3D embeddings and MRoPE~\citep{qwen2vl}.
These results support our default predictor, which combines Qwen3.5 with 3D input positional embeddings and learned queries.

\input{figure/llm-ablation-1}
\input{figure/llm-ablation-2}

\input{figure/arch-examples}

\subsection{Predictor Initialization and Joint Training}
\label{sec:ablation-initialization-joint-training}

We examine how predictor initialization and training scope affect MPC planning (\cref{tab:predictor-initialization-joint-training}).
All variants use the same predictor architecture and planning settings (\cref{sec:mpc-setup}).
The jointly trained predictors use the full data mixture, including control trajectories and image and video transitions.
For comparison, we train separate predictors from pretrained weights on each domain's control transitions, using the DROID-trained predictor for zero-shot evaluation on RoboCasa.

With pretrained initialization, joint training achieves an average score of 60.2, compared with 54.5 for single-domain training.
Random initialization lowers the jointly trained predictor's average to 55.6.
The two initializations reach similar final validation losses (\cref{fig:llm-ablation-2}(a)), suggesting that prediction loss alone does not fully capture planning performance.

\input{table/predictor-initialization-joint-training}

\subsection{\ours\ Vision Encoder Ablations}
\label{sec:ablation-vision-encoder}

\paragraph{Setup.}
JEPA-WMs~\citep{jepa-wms} shows that the preferred vision encoder for planning varies across domains.
We examine this dependence in \ours\ by comparing frozen DINOv2-S/14~\citep{dinov2}, DINOv3-L/16~\citep{dinov3}, SigLIP2-So400m/14~\citep{siglip2}, and DINOv2-S/14 + DINOv3-L/16.
For each encoder configuration, we train a separate predictor on all training data, including control trajectories and image and video transitions.
For DINOv2-S/14 + DINOv3-L/16, one predictor receives both representations and predicts both as future targets.
We also report JEPA-WMs with each DINO encoder.
\cref{tab:vision-encoder-ablation} presents results on all ten benchmarks and their average score.
The comparison in \cref{tab:planning-benchmarks} allows the vision encoder to vary across benchmarks, as does JEPA-WMs~\citep{jepa-wms}; we include those reported scores as reference rows. DINO-WM in \cref{tab:planning-benchmarks} likewise uses the better-performing vision encoder for each benchmark, while V-JEPA 2-AC retains the V-JEPA 2 encoder used in the original method.

\paragraph{Results.}
Our encoder ablations suggest that DINOv2-S/14 and DINOv3-L/16 provide complementary information for planning, motivating further work on combining encoders~\citep{cambrian} and learning representations tailored to latent world modeling.
The preferred DINO encoder varies across benchmarks for both methods.
For \ours, DINOv2-S/14 performs better on Push-T and Granular, while DINOv3-L/16 performs better on PointMaze and both MetaWorld tasks.
Using DINOv2-S/14 + DINOv3-L/16 with one predictor increases the average to 63.0, although improvements are not uniform across benchmarks.

\input{table/vision-encoder-ablation}

\subsection{Additional Results on Planning with Qualitative Language Actions}
\label{sec:additional-quali-lang}

\input{table/qualitative-language-actions}

We compare numerical and qualitative language actions to evaluate whether \ours\ can plan from descriptions that omit exact action magnitudes.
The action conversion and planning procedure are detailed in \cref{sec:details-quali-lang}.

\cref{tab:qualitative-language-actions} shows that qualitative language actions support planning in several domains.
\ours\ achieves success rates of 83.3\% on Wall, 70.8\% on MetaWorld-R, and 60.4\% on MetaWorld-RW.
Success rates decrease on some tasks, as expected when many distinct numerical magnitudes are mapped to only three qualitative levels.
Even with this reduced information, it is still surprising that \ours\ can use descriptions of direction and relative magnitude to guide planning without receiving exact numerical action values.

\subsection{Additional Results on Action Phrasing Robustness}
\label{sec:additional-phrasing-lang}

\input{table/phrasing-robustness}

We provide the domain-level results underlying \cref{fig:phrasing-robustness} and the corresponding action template paraphrases.
We compare predictors trained on control trajectories alone and on all data, including image and video transitions.
At evaluation, we apply the substitutions in \cref{tab:phrasing-robustness} (top) to the language conditions supplied to the predictor, preserving the numerical action values and directions.
For each predictor, we keep the model weights and planning settings unchanged.
\cref{tab:phrasing-robustness} shows that paraphrasing lowers scores in all five domains when training uses control trajectories alone, reducing the average from 59.3 to 52.5.
These results suggest that image and video transitions reduce the predictor's dependence on the wording of the action templates.

\subsection{Additional Results on Latent vs. Generative World Models}
\label{sec:additional-latent-vs-generative}

\input{figure/latent-vs-generative-per-domain}

\cref{fig:latent-vs-generative-per-domain} breaks down the planning results in \cref{fig:latent-vs-pixel} by domain.
We use the full-depth LLM predictor for \ours\ and the full-depth DiT for both RAEv2 variants.
\ours\ outperforms both RAEv2 variants on six of the seven benchmarks and matches five-step RAEv2 on MetaWorld-R.
Its average score is 59.4, compared with 40.4 and 48.0 for one- and five-step RAEv2, respectively.
These results follow the trend in \cref{sec:latent-vs-pixel}, showing that the planning advantage of \ours\ extends across domains.

\subsection{Additional Qualitative Planning Results}
\label{sec:additional-qualitative-planning}

\subsubsection{Planning with Numerical Language Actions}
\label{sec:qualitative-planning-numerical}

We provide additional visualizations of \ours\ with numerical language actions on Wall, PointMaze, Push-T, MetaWorld, DROID, and RoboCasa Reach and Place in \cref{fig:wm-num-wall}, \cref{fig:wm-num-point}, \cref{fig:wm-num-pusht}, \cref{fig:wm-num-meta}, \cref{fig:wm-num-droid}, \cref{fig:wm-num-robo}, and \cref{fig:wm-num-robo-p}.
The figures show initial and goal observations alongside decoded future predictions, with the language actions below the frames; Wall also includes PCA visualizations of the predicted latents.

\subsubsection{Planning with Qualitative Language Actions}
\label{sec:qualitative-planning-qualitative}

We provide additional planning examples with qualitative language actions on Wall, PointMaze, Push-T, MetaWorld, and RoboCasa Reach and Place in \cref{fig:wm-qual-wall}, \cref{fig:wm-qual-point}, \cref{fig:wm-qual-pusht}, \cref{fig:wm-qual-meta}, \cref{fig:wm-qual-robo}, and \cref{fig:wm-qual-robo-p}.
Each example shows simulator observations during MPC execution, with qualitative action descriptions below the frames; action conversion and execution follow \cref{sec:details-quali-lang}.

\subsection{Qualitative Policy Results}
\label{sec:qualitative-policy}

We provide additional \ourpolicy\ rollouts on LIBERO Spatial (\cref{fig:policy-libero-spatial-1}, \cref{fig:policy-libero-spatial-2}), Object (\cref{fig:policy-libero-object-1}, \cref{fig:policy-libero-object-2}), Goal (\cref{fig:policy-libero-goal-1}, \cref{fig:policy-libero-goal-2}), and Long (\cref{fig:policy-libero-long-1}, \cref{fig:policy-libero-long-2}).
RoboCasa examples cover coffee tasks (\cref{fig:policy-robocasa-coffee-1}, \cref{fig:policy-robocasa-coffee-2}), opening doors and drawers (\cref{fig:policy-robocasa-open-1}, \cref{fig:policy-robocasa-open-2}), object transfer (\cref{fig:policy-robocasa-pick-1}, \cref{fig:policy-robocasa-pick-2}), and stove control (\cref{fig:policy-robocasa-turn-1}).
For each camera view, simulator observations (upper row) are paired with decoded future predictions (lower row) during policy execution.

\section{Implementation and Experimental Details}
\label{sec:implementation-details}

\subsection{\ours\ Training Data Details}
\label{sec:details-training-data}

\input{table/world-model-training-data}

\cref{tab:world-model-training-data} reports the training data size for each dataset.
We combine these datasets and sample in proportion to their sizes.
We reserve approximately 2\% of episodes or editing pairs for validation and keep sequences from the same source video in one split.

\paragraph{Simulation data.}
We use the six simulation datasets from DINO-WM and JEPA-WMs~\citep{dino-wm,jepa-wms} listed in \cref{sec:mpc-setup}.
For Wall, PointMaze, Push-T, and MetaWorld, each transition spans five simulator steps; for Rope and Granular, it spans one push.
We enumerate overlapping transitions within each episode and pair each with both numerical and qualitative action descriptions.
Action aggregation is described in \cref{sec:details-ours-mpc-planning}.

\paragraph{DROID.}
We use the LeRobot release of DROID~\citep{droid}, retaining one external camera view and every fourth frame of the 15-fps recordings.\footnote{\url{https://huggingface.co/datasets/lerobot/droid_1.0.1}}
We derive action descriptions from changes in end-effector position, orientation, and gripper state between retained frames.
Our DROID subset contains 1.5 million target transitions, approximately 20\% of the original dataset's transitions.
This keeps DROID from dominating the mixture under a fixed training budget.
The same subset is used for both action description formats.

\paragraph{Image and video data.}
From OpenVid-1M~\citep{Openvid}, we select clips by duration, motion, and aesthetic scores and extract ten-frame sequences at 0.5-second intervals.
Qwen3-VL-30B-A3B-Instruct describes the visible change between each adjacent frame pair to provide its language condition.
We also use source--edited image pairs and their instructions from OmniEdit~\citep{Omniedit} and ShareGPT-4o-Image~\citep{sharegpt}.

Further DROID-focused post-training, as in V-JEPA 2-AC~\citep{v-jepa-2}, may improve robot planning; we leave this to future work.

\subsection{\ours\ Architecture Details}
\label{sec:details-architecture}

Each example contains observed context frames, language conditions for the corresponding transitions, and one target frame.
In multi-frame training, we sample the context length uniformly from one to three frames, subject to the available episode history; image pairs provide one context frame.
A two-layer MLP projects each frame's patch representations into the language model's embedding space.
As illustrated in \cref{fig:arch-examples}(a), we append one query token per target patch after the context and language tokens.
The queries are copies of a shared learned vector, with their target locations specified by 3D positional embeddings.
These embeddings encode a temporal index and the row and column of each spatial patch.
Context patches share a temporal index within each frame, while queries receive the target frame's temporal index and spatial coordinates.
We concatenate sinusoidal encodings for the three axes, multiply them by a learnable scale, and add them to the input embeddings while retaining the backbone's native positional encoding.
The appended queries receive the preceding visual and language context through the LLM backbone.
A two-layer MLP prediction head maps their output hidden states back to the vision encoder's feature dimension, producing the future patch grid in one forward pass.
Both projection MLPs use GELU between their linear layers.
The target frame is used only to provide supervision for these predictions.

\subsection{\ours\ Training Details}
\label{sec:details-training}

We keep the observation encoder frozen and jointly train the language-model backbone, visual projection layers, and learned query tokens.
The same encoder provides the context and target representations.
We optimize the prediction objective in \cref{eq:training-objective}.
We use AdamW with an effective global batch size of 512, $(\beta_1,\beta_2)=(0.9,0.98)$, and weight decay of $0.05$.
The learning rate increases linearly from $1\times10^{-5}$ to $2\times10^{-4}$ over the first 1,000 optimizer steps, followed by a cosine decay schedule with a minimum learning rate of $1\times10^{-6}$.
We clip the gradient norm at $1.0$ and use gradient accumulation to maintain the global batch size.
We train for 20 epochs using bfloat16 mixed precision on eight NVIDIA H100 80GB GPUs.
We define each epoch as approximately one million training examples sampled from the training pool.

\subsection{MPC Planning and Evaluation Details}
\label{sec:details-mpc-planning}

\subsubsection{Benchmark Settings and Evaluation Metrics}
\label{sec:details-planning-benchmarks}

\input{table/planning-setup}

We follow the planning and evaluation protocols of JEPA-WMs~\citep{jepa-wms} and DINO-WM~\citep{dino-wm}, with our own success criteria for Rope and Granular. As in \cref{sec:mpc-planning}, we evaluate planning with model predictive control (MPC), using the cross-entropy method (CEM) to optimize action sequences. \Cref{tab:planning-setup} summarizes the MPC settings, CEM parameters, and evaluation criteria for each benchmark.
One planning step corresponds to one transition predicted by the world model. In simulation, each action update within a step is called a substep. The planning horizon $H$ specifies how many steps the model predicts ahead during each candidate rollout, while trajectory length gives the total step budget.

In simulation, we execute an initial segment of the sequence optimized by CEM, whose length is specified by executed steps. For benchmarks requiring replanning, we follow the MPC loop: after executing this segment, we obtain a new observation and re-optimize the action sequence with CEM. \cref{fig:planning-step-vis} illustrates these settings. Wall executes all six planned steps in one round, whereas MetaWorld Reach executes three of six and Rope and Granular execute one of five before replanning.

For DROID, we evaluate planned actions offline without simulator execution. We compute the endpoint position error $\mathrm{error}_{\mathrm{xyz}}$ in meters from the cumulative planned displacement and report the Action Score:
\begin{equation}
    \mathrm{Action\ Score}
    = \max\left(0,\,800\left(0.1-\mathrm{error}_{\mathrm{xyz}}\right)\right).
    \label{eq:droid-action-score}
\end{equation}
For zero-shot planning on RoboCasa~\citep{robocasa}, we reuse DROID's action sampling statistics because the two domains share the same action definitions and dimensionality. 

\input{figure/planning-setup-vis}

\subsubsection{Action Aggregation and Execution}
\label{sec:details-ours-mpc-planning}

The baselines represent each planning step by concatenating the actions across its substeps. For example, five substeps with two action components, $x$ and $y$, yield a 10-dimensional vector. Verbalizing every component would produce a long language condition. We therefore sum each component across substeps when constructing the training data and verbalize only these sums. In this example, the language condition describes two values,
$\bar{x}=\sum_{i=1}^{5}x_i$ and
$\bar{y}=\sum_{i=1}^{5}y_i$,
as shown in \cref{fig:substep-aggregation}(a).

Within MPC, CEM optimizes action sequences in this aggregated action space, with its initial sampling statistics computed from the aggregated training data. Each sampled action is verbalized using the training templates to condition the predictor. To execute a selected action, we divide each optimized sum by the number of substeps and repeat the resulting action at every substep. For the five-substep example, we apply $(\bar{x}/5,\bar{y}/5)$ five times, as illustrated in \cref{fig:substep-aggregation}(b). This preserves the component sums but discards variation within a step, which can lead to differences between predicted and executed transitions. It is notable that \ours\ achieves accurate future prediction and high planning success rates despite this loss of information.

\input{figure/substep-aggregation}

\subsubsection{Qualitative Language Action Conversion}
\label{sec:details-quali-lang}

\cref{fig:qualitative-language-actions} illustrates the conversion rule and the planning procedure.

\paragraph{Action conversion.}
Numerical language actions preserve each action component's direction and magnitude.
To construct qualitative descriptions, we define a conversion rule for each component and direction using the action range observed in the corresponding training data.
We divide the interval from zero to the maximum observed magnitude into three equal-width bins, represented by \emph{slightly}, no modifier, and \emph{far}, respectively.
The direction remains unchanged.
For example, when the maximum leftward magnitude is 6, \prompt{move 0.4 left} becomes \prompt{move slightly left}, and \prompt{move 4.5 left} becomes \prompt{move far left}.

\paragraph{MPC planning procedure.}
At each MPC planning round, CEM samples numerical action sequences as in the main experiments.
Before each sequence is evaluated by \ours, the conversion rule maps its actions to qualitative descriptions.
These descriptions condition the latent rollout, which CEM scores using the same goal-distance cost in \cref{eq:planning-cost}.
The simulator executes the selected actions using their original numerical values, and re-planning follows the same MPC procedure.
We keep the trained predictor and all other planning settings unchanged.

\input{figure/qualitative-language-actions}

\subsection{RAEv2 Baseline Details}
\label{sec:details-raev2}

\paragraph{Model and initialization.}
We initialize the generative world model baseline's DiT from the released RAEv2 checkpoint pretrained for image generation on ImageNet.\footnote{\url{https://huggingface.co/nyu-visionx/RAEv2-models/tree/main/stage2/imagenet/dinov3l-k7}}
Its frozen DINOv3-L/16 encoder aggregates features from seven layers to form the visual representation $e_t = E_\theta(x_t)$.
A frozen Qwen3-0.6B encodes the language condition $z$ into text embeddings $T(z)$.

\paragraph{Training.}
For each transition $(x_t, z, x_{t+1})$, we encode the future observation as $e_{t+1} = E_\theta(x_{t+1})$ and interpolate it with Gaussian noise for flow matching~\citep{raev2}:
\begin{equation}
    e_{t+1}^{(\tau)} = (1-\tau)e_{t+1} + \tau\epsilon,
    \qquad \epsilon \sim \mathcal{N}(0,I), \quad \tau \in [0,1].
    \label{eq:raev2-noising}
\end{equation}
Here, $t$ indexes environment time and $\tau$ denotes the noise level.
The DiT receives $e_{t+1}^{(\tau)}$, $\tau$, the current observation representation $e_t$, and the text embeddings $T(z)$.
We train it with flow matching and a representation alignment (REPA) loss~\citep{repa} that aligns intermediate DiT features with DINOv2-B/14 features of the future observation.

\paragraph{MPC planning.}
We use the same MPC procedure as \ours, with CEM optimizing candidate action sequences.
Starting from Gaussian noise, the DiT produces a future representation $\hat{e}_{t+1}$ through denoising, conditioned on the current observation and language action.
We evaluate RAEv2 with one and five denoising steps per prediction.
CEM scores candidate action sequences using the mean squared error between the final predicted representation $\hat{e}_{t+H}$ and the encoded goal $E_\theta(x_{\mathrm{goal}})$, following \cref{eq:planning-cost}.
The goal is encoded with the same DINOv3-L/16 feature aggregation as the training targets.
Pixel decoding is not used to compute the planning cost.

\paragraph{Visualization.}
For \cref{fig:imagination-visualization}, RAEv2's predicted representations are mapped to pixels with its released DINOv3-L-K7 decoder,\footnote{\url{https://huggingface.co/nyu-visionx/RAEv2-models/tree/main/stage1/general/dinov3l-k7}} while \ours\ uses the decoder described in \cref{sec:mpc-setup}.

\subsection{\ourpolicy\ Architecture Details}
\label{sec:details-policy-arch}

As illustrated in \cref{fig:policy-architecture}, \ourpolicy\ extends the predictor of \ours\ with proprioceptive input and action queries.
Following previous VLAs that use SigLIP-based visual encoders~\citep{black2024pi_0, kim2025fine, li2025cogvla}, we use SigLIP2-So400M/14 at $224\times224$ resolution and the 24-layer Qwen3.5-0.8B as the predictor backbone.
Each camera view is encoded independently, and a shared two-layer MLP projects its patch representations into the backbone's embedding space.
We add learned view embeddings and map the current proprioceptive state to one token through a separate MLP.

The input sequence contains visual tokens, the proprioceptive token, task instruction tokens, action queries, and future queries, in this order.
We use $N=16$ action queries for LIBERO and $N=32$ for RoboCasa, with one future query per spatial patch in each camera view.
A two-layer MLP maps the future-query outputs to the encoder's feature dimension to predict the representations at $t+N$.
The action generator follows Isaac-GR00T~\citep{issac-gr00t}, using 16 DiT blocks of width 1,536 with alternating cross-attention and self-attention.
It generates the action chunk from Gaussian noise, conditioned on the predictor's full sequence of output hidden states.

\subsection{Policy Model Training Data}
\label{sec:details-policy-training-data}

We describe the datasets used to fine-tune \ourpolicy\ in \cref{sec:lala-policy}.
The episode counts below refer to the data before the training/validation split.

\paragraph{LIBERO.}
We use the Spatial, Object, Goal, and Long suites of LIBERO~\citep{liu2024libero}, comprising 40 language-conditioned manipulation tasks.
The original data contain 50 human demonstrations per task, for a total of 2,000 episodes.
We adopt the replayed datasets released by Cosmos Policy~\citep{cosmos-policy}.\footnote{\url{https://huggingface.co/datasets/nvidia/LIBERO-Cosmos-Policy}}
Its preprocessing removes no-op transitions and filters unsuccessful replays from the policy demonstration set.
The resulting set contains 1,700 successful episodes: 430, 456, 438, and 376 from Spatial, Object, Goal, and Long, respectively.
We use these episodes for policy training and the full set of 2,000 replays, including failures, for action-conditioned future prediction.

\input{figure/policy-architecture.tex}

\paragraph{RoboCasa.}
We use 24 kitchen manipulation tasks from RoboCasa~\citep{robocasa} under two data settings.
For the setting with approximately 50 human demonstrations per task, we use the Cosmos Policy release.\footnote{\url{https://huggingface.co/datasets/nvidia/RoboCasa-Cosmos-Policy}}
It provides 1,199 successful episodes for policy training and 1,291 replay episodes, including 92 failures, for action-conditioned future prediction.
For the 300-demonstration setting, we use RoboCasa's released data generated with MimicGen~\citep{mandlekar2023mimicgen}.

\subsection{\ourpolicy\ Fine-tuning Details}
\label{sec:details-policy-fine-tuning}

We initialize the predictor from pretrained \ours\ and jointly train it with the newly added policy modules, while keeping the observation encoder frozen.
The action generator is trained from scratch.
We minimize a flow-matching velocity loss on policy examples and a mean squared error loss on future representations for both types of examples.
Both loss coefficients are set to $1$.

We use AdamW with an effective global batch size of 256, $(\beta_1,\beta_2)=(0.9,0.98)$, and weight decay of $0.05$.
The learning rate increases linearly from $1\times10^{-6}$ to $1\times10^{-4}$ over the first 1,000 optimizer steps, then follows a cosine decay to $1\times10^{-6}$.
We clip the gradient norm at $1.0$ and use bfloat16 mixed precision on eight NVIDIA H100 80GB GPUs, with gradient accumulation to maintain the global batch size.
We train for 8 epochs on LIBERO, 12 epochs on RoboCasa with approximately 50 human demonstrations per task, and 3 epochs on RoboCasa with 300 MimicGen demonstrations per task.

\subsection{Policy Model Evaluation Details}
\label{sec:details-policy-evaluation}

We evaluate \ourpolicy\ through direct interaction with the LIBERO~\citep{liu2024libero} and RoboCasa~\citep{robocasa} simulators, following the evaluation procedure of Cosmos Policy~\citep{cosmos-policy}.
The policy receives the task instruction, current proprioceptive state, and camera observations.
LIBERO provides an external camera view and a wrist view, while RoboCasa provides left and right external views and a wrist view.

After 10 initial no-op steps, the policy predicts an action chunk.
On LIBERO, the chunk contains 16 actions, all of which are executed before the next query.
On RoboCasa, the chunk contains 32 actions, of which we execute the first 16.
Each subsequent query uses the latest simulator observations.
The predicted future representations are not used for action selection during evaluation.

We run 50 episodes for each of the 40 tasks across LIBERO's Spatial, Object, Goal, and Long suites.
For RoboCasa, we run 50 episodes for each of 24 tasks, distributed equally across five test scenes with held-out object instances from split B.
The step budgets follow the Cosmos Policy evaluation code: 220, 280, 300, and 520 for the four LIBERO suites, and 300--1,000 depending on the RoboCasa task.
An episode ends upon task success or exhaustion of its step budget.
We report success rates averaged across suites for LIBERO and across tasks for RoboCasa.

\clearpage
\input{figure-qualitative/world-model/all}

\clearpage
\input{figure-qualitative/policy/all}

\end{document}

%% file: sec/config.tex
\usepackage[nopatch=footnote]{microtype}
\usepackage{url}
\usepackage{booktabs}

\usepackage{xcolor}
\usepackage{hyperref}
\definecolor{darkblue}{rgb}{0, 0, 0.5}
\hypersetup{colorlinks=true, citecolor=darkblue, linkcolor=darkblue, urlcolor=darkblue}

\usepackage[utf8]{inputenc}
\usepackage[T1]{fontenc}
\usepackage{carlito}      
\usepackage{nicefrac}

\usepackage{wrapfig}
\usepackage{graphicx}
\usepackage{float}
\usepackage{caption}
\usepackage{amsmath}
\usepackage{amssymb}
\usepackage{pifont}
\usepackage{array}
\usepackage{multirow, overpic, textpos}
\usepackage{colortbl}
\usepackage[linesnumbered,ruled,vlined]{algorithm2e}
\SetKwInOut{Parameter}{Parameters}
\usepackage{listings}
\usepackage{tikz}
\usepackage{bm}
\usepackage{arydshln}
\usepackage{hhline}
\usepackage{enumitem}
\usepackage{soul}
\usepackage{setspace}
\usepackage{tcolorbox}
\tcbuselibrary{most}
\usetikzlibrary{arrows.meta, positioning, calc, decorations.pathreplacing}

\usepackage{adjustbox}

\newlength\savewidth

\newcolumntype{x}[1]{>{\centering\arraybackslash}p{#1pt}}
\newcolumntype{y}[1]{>{\raggedright\arraybackslash}p{#1pt}}
\newcolumntype{z}[1]{>{\raggedleft\arraybackslash}p{#1pt}}
\newcommand{\app}{\raise.17ex\hbox{$\scriptstyle\sim$}}

\newcommand{\x}{{\times}}
\definecolor{deemph}{gray}{0.58}

\definecolor{baselinecolor}{gray}{.9}

\newcommand{\gr}{\rowcolor[gray]{.95}}

\newcommand{\RNum}[1]{\uppercase\expandafter{\romannumeral #1\relax}}

\definecolor{userborder}{rgb}{0.2608, 0.7071, 0.0218}
\definecolor{brightgreen}{rgb}{0.001, 0.921, 0.38}
\definecolor{darkgreen}{rgb}{0.3608, 0.6471, 0.2118}
\definecolor{userfont}{RGB}{0, 0, 0}
\definecolor{darkred}{rgb}{0.8,0.02,0.02}
\definecolor{ddarkred}{rgb}{0.5,0.02,0.02}
\definecolor{highlightred}{HTML}{ffcbc0}

\definecolor{cadmiumgreen}{rgb}{0.0, 0.42, 0.24}
\definecolor{aliceblue}{rgb}{0.91, 0.94, 0.97}
\definecolor{lightblue}{rgb}{0.83, 0.89, 0.97}
\definecolor{Blue9}{rgb}{0.098,0.3,0.9}

\definecolor{paviolet}{HTML}{DEDBE9}
\definecolor{pasky}{HTML}{D5EBEE}
\definecolor{payellow}{HTML}{FFF6EA}
\definecolor{pared}{HTML}{F9E5ED}

\RequirePackage[capitalize, noabbrev, nameinlink]{cleveref}

\newcommand{\cmark}{\ding{51}}%
\newcommand{\xmark}{\ding{55}}%

\newcommand{\ie}{\emph{i.e.}}

\usepackage{xspace}
\newcommand{\dy}[1]{\textcolor{red}{#1\xspace}}

\newtcbox{\prompt}{on line, colback=gray!10, colframe=gray!40,
  boxrule=0.3pt, arc=1.5pt, boxsep=0pt,
  left=2pt, right=2pt, top=0.8pt, bottom=0.8pt,
  fontupper=\footnotesize
}

\definecolor{hlviolet}{HTML}{EFEEFE}  
\definecolor{hlsky}   {HTML}{E0F5F8}  
\definecolor{hlpink}  {HTML}{FDEAF2}  
\definecolor{hlred}   {HTML}{FFF1EE}  
\definecolor{hlyellow}{HTML}{FFFFE4}  

\newif\ifshowhl \showhltrue            %

\DeclareRobustCommand{\hlc}[2]{%
  \ifshowhl{\sethlcolor{#1}\hl{#2}}\else#2\fi}

\DeclareRobustCommand{\hlv}[1]{\hlc{hlviolet}{#1}}

\DeclareRobustCommand{\hly}[1]{\hlc{hlyellow}{#1}}
\DeclareRobustCommand{\hlp}[1]{\hlc{hlpink}{#1}}
\DeclareRobustCommand{\hlr}[1]{\hlc{hlred}{#1}}

\soulregister{\cite}{7}  \soulregister{\citep}{7} \soulregister{\citet}{7}
\soulregister{\ref}{7}   \soulregister{\cref}{7}  \soulregister{\Cref}{7}
\soulregister{\texttt}{7}

\usepackage{etoolbox}

%% file: sec/author.tex
\author{%
  Junha Song\textsuperscript{1*}\quad
  Geonmo Gu\textsuperscript{1}\quad
  Jeongeun Park\textsuperscript{2}\quad
  Heejeong Nam\textsuperscript{3}\quad
  Lucas Maes\textsuperscript{4}\\
  Byeongho Heo\textsuperscript{1}\quad
  Dongyoon Han\textsuperscript{1}\quad
  Sangdoo Yun\textsuperscript{1\ensuremath{\dagger}}
}
\affiliations{%
  \textsuperscript{1}NAVER AI Lab\quad
  \textsuperscript{2}Georgia Institute of Technology\\
  \textsuperscript{3}Brown University\quad
  \textsuperscript{4}Mila \& Universit\'e de Montr\'eal
}
\contribution[*]{Project Lead}
\contribution[\dagger]{Corresponding Author}

%% file: figure/teaser.tex
\begin{figure}[t]
    \centering
    \includegraphics[width=1.\linewidth]{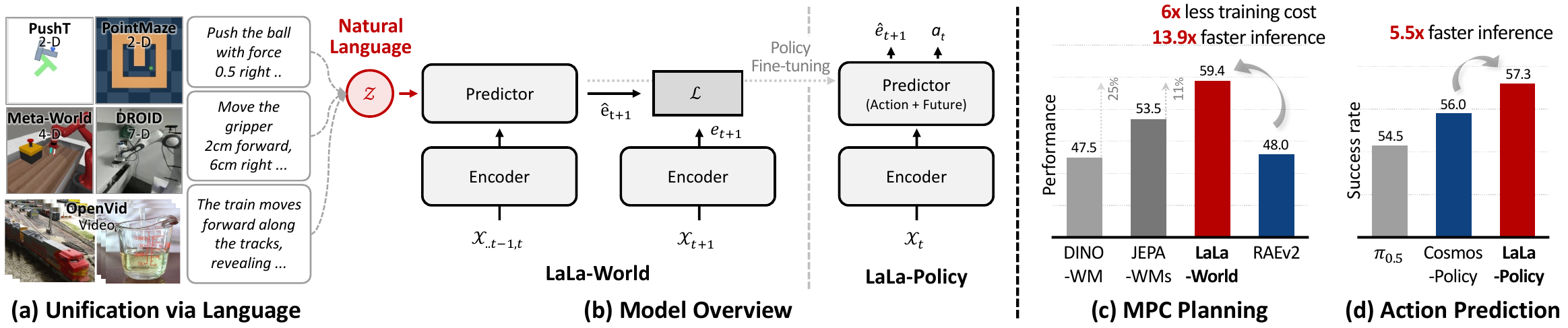}
    \vspace{-1.6em}
    \caption{
    \textbf{Overview of \ours.}
    \textbf{(a)} Language provides a shared interface for numerical actions and visual change, enabling joint training on control and video data.
    \textbf{(b)} A single predictor built on a language model uses encoded observations and language conditions to predict future visual representations. Fine-tuning the pretrained \ours\ yields \ourpolicy, which predicts actions as well as future visual representations.
    \textbf{(c)} The jointly trained predictor achieves higher average MPC planning performance than both separately trained LWM baselines and the generative world model baseline, RAEv2.
    \textbf{(d)} \ourpolicy{} performs competitively with our adapted Cosmos Policy baseline on RoboCasa, while reducing average inference time per trajectory by a factor of $5.5\times$.
    }
    \label{fig:teaser}
    \vspace{-.2em}
\end{figure}

%% file: figure/lang-planning.tex
\begin{figure}[t]
    \centering
    \includegraphics[width=1.\linewidth]{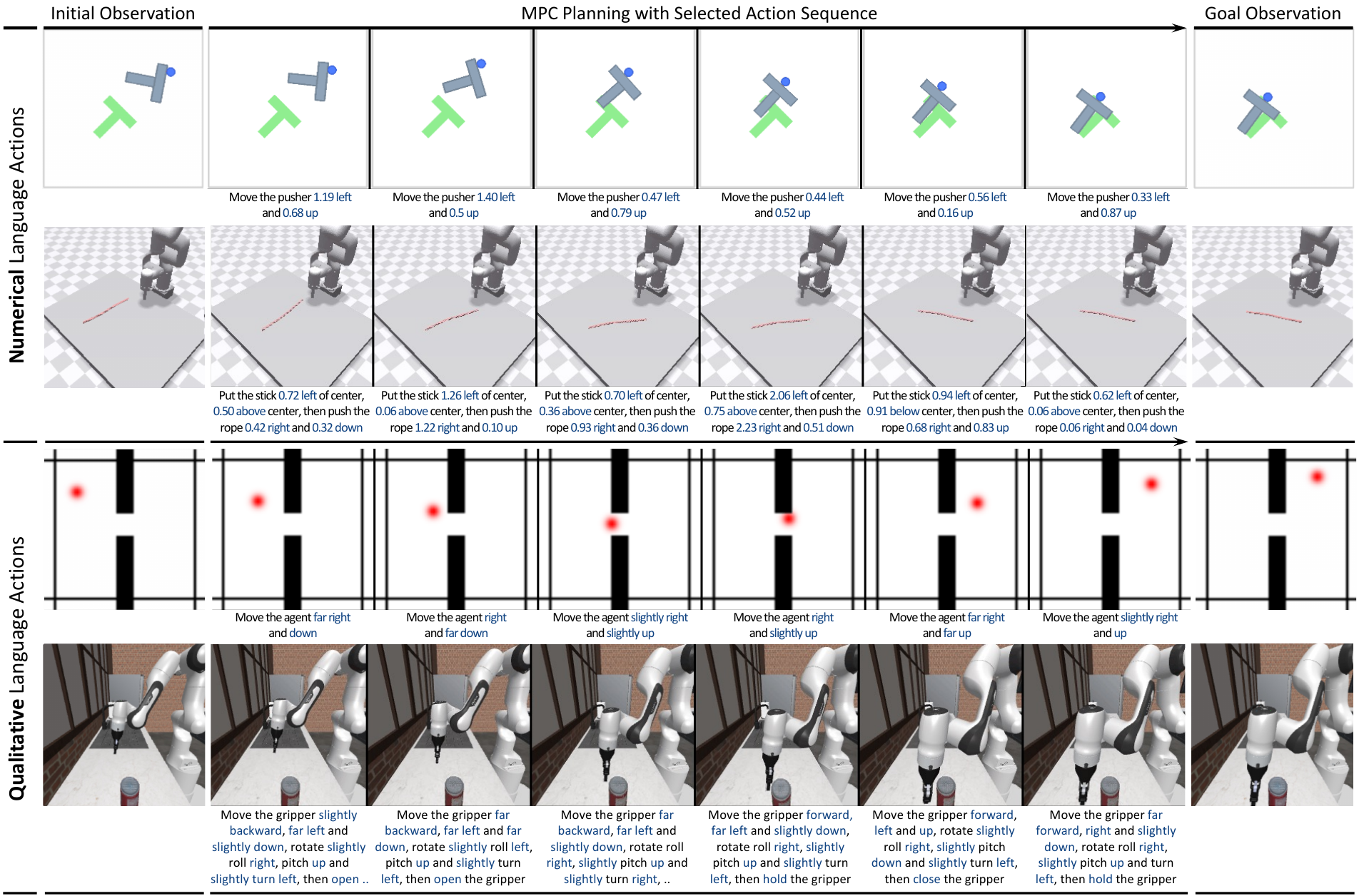}
    \vspace{-1.2em}
\caption{
\textbf{MPC planning with language actions.}
\ours\ imagines future representations under candidate actions with MPC and selects the most promising action toward the goal.
Because all actions share one language interface, a single \ours\ predictor can plan across multiple domains, despite differences in action dimensionality and physical meaning.
The predictor also supports language-based actions without numerical values (bottom two rows), as detailed in \cref{sec:details-quali-lang}.
The visualized observations result from executing the selected actions in the simulator.}
    \label{fig:lang-planning}
    \vspace{-1.em}
\end{figure}

%% file: figure/data-examples.tex
\begin{figure}[t]
    \centering
    \includegraphics[width=1.\linewidth]{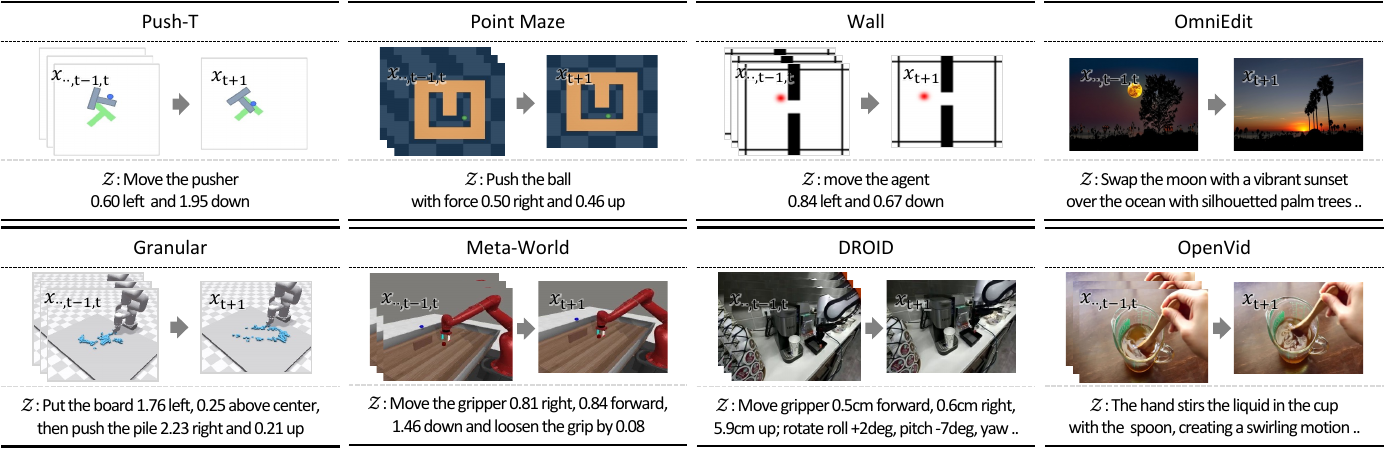}
    \vspace{-1.2em}
    \caption{
    \textbf{Examples of language-conditioned transition data.}
    Each example consists of past observations, a language condition, and the next observation. 
    Language conditions describe either numerical control actions from simulation and robotics or visual changes in images and videos.
    This shared interface enables training a single LWM across different action spaces and data sources.
    }
    \label{fig:data-examples}
    \vspace{-.5em}
\end{figure}

%% file: table/planning-benchmarks.tex
\begin{table}[t]
    \centering
    {\footnotesize
    \renewcommand{\arraystretch}{1.3}
    \setlength{\aboverulesep}{0.5pt}
    \setlength{\belowrulesep}{0.5pt}
    \setlength{\arrayrulewidth}{0.8pt}
    \setlength{\dashlinedash}{0.6pt}
    \setlength{\dashlinegap}{1.5pt}
    \begin{minipage}{\linewidth}
        \centering
    \resizebox{\linewidth}{!}{%
    \begin{tabular}{>{\arrayrulecolor{black!25}}w{l}{9.3em}:w{c}{5.7em}:*{5}{w{c}{5.7em}}:w{c}{4.5em}}
    \toprule
        Method & Joint-training & Wall & PointMaze & Push-T & MetaWorld-R & MetaWorld-RW & Average \\
        \midrule
        DreamerV3~(\citeyear{DreamerV3})       & \xmark & $39.6$\,{\tiny $(5.0)$}          & $43.8$\,{\tiny $(5.1)$}          & $32.3$\,{\tiny $(4.8)$}          & $25.0$\,{\tiny $(6.2)$}          & $10.4$\,{\tiny $(4.4)$}          & $30.2$\,{\tiny $(2.3)$}          \\
        TD-MPC2~(\citeyear{TD-MPC-2})          & \xmark & $28.1$\,{\tiny $(4.6)$}          & $8.3$\,{\tiny $(2.8)$}           & $16.7$\,{\tiny $(3.8)$}          & $12.5$\,{\tiny $(4.8)$}          & $22.9$\,{\tiny $(6.1)$}          & $17.7$\,{\tiny $(2.0)$}          \\
        IRIS~(\citeyear{IRIS})                 & \xmark & $30.2$\,{\tiny $(4.7)$}          & $41.7$\,{\tiny $(5.0)$}          & $31.2$\,{\tiny $(4.7)$}          & $20.8$\,{\tiny $(5.9)$}          & $6.2$\,{\tiny $(3.5)$}           & $26.0$\,{\tiny $(2.2)$}          \\
        LeWorldModel~(\citeyear{leworldmodel}) & \xmark & $24.0$\,{\tiny $(4.4)$}          & $24.0$\,{\tiny $(4.4)$}          & $59.4$\,{\tiny $(5.0)$}          & $16.7$\,{\tiny $(5.4)$}          & $16.7$\,{\tiny $(5.4)$}          & $28.1$\,{\tiny $(2.2)$}          \\
        DINO-WM~(\citeyear{dino-wm})           & \xmark & $90.6$\,{\tiny $(3.0)$}          & $59.4$\,{\tiny $(5.0)$}          & $65.6$\,{\tiny $(4.8)$}          & $52.1$\,{\tiny $(7.2)$}          & $43.8$\,{\tiny $(7.2)$}          & $62.3$\,{\tiny $(2.5)$}          \\
        JEPA-WMs~(\citeyear{jepa-wms})         & \xmark & $79.2$\,{\tiny $(4.1)$}          & $\mathbf{60.4}$\,{\tiny $(5.0)$} & $68.8$\,{\tiny $(4.7)$}          & $66.7$\,{\tiny $(6.8)$}          & $47.9$\,{\tiny $(7.2)$}          & $64.6$\,{\tiny $(2.6)$}          \\
        \gr
        \textbf{\ours}                    & \cmark & $\mathbf{95.8}$\,{\tiny $(2.0)$} & $56.2$\,{\tiny $(5.1)$}          & $\mathbf{81.2}$\,{\tiny $(4.0)$} & $\mathbf{75.0}$\,{\tiny $(6.2)$} & $\mathbf{52.1}$\,{\tiny $(7.2)$} & $\mathbf{72.1}$\,{\tiny $(2.3)$} \\
        \bottomrule
    \end{tabular}%
    }
    \end{minipage}
    \par\vspace{0.3em}
    \begin{minipage}{\linewidth}
        \centering
    \resizebox{\linewidth}{!}{%
    \begin{tabular}{>{\arrayrulecolor{black!25}}w{l}{9.3em}:w{c}{5.7em}:*{5}{w{c}{5.7em}}:w{c}{4.5em}}
    \toprule
        Method & Joint-training & Rope & Granular & RoboCasa-R & RoboCasa-P & DROID & Average \\
        \midrule
        DINO-WM~(\citeyear{dino-wm})       & \xmark & $54.2$\,{\tiny $(10.2)$}          & $50.0$\,{\tiny $(14.4)$}          & $18.8$\,{\tiny $(6.9)$} & $21.9$\,{\tiny $(7.3)$}  & $39.4$\,{\tiny $(3.8)$}          & $36.8$\,{\tiny $(4.1)$}          \\
        JEPA-WMs~(\citeyear{jepa-wms})     & \xmark & $62.5$\,{\tiny $(9.9)$}           & $50.0$\,{\tiny $(14.4)$}          & $37.5$\,{\tiny $(8.6)$} & $31.2$\,{\tiny $(8.2)$}  & $\mathbf{43.0}$\,{\tiny $(3.8)$} & $44.8$\,{\tiny $(4.3)$}          \\
        V-JEPA 2-AC~(\citeyear{v-jepa-2})   & \xmark & --                       & --                       & $15.6$\,{\tiny $(6.4)$} & $34.4$\,{\tiny $(8.4)$}  & $42.9$\,{\tiny $(2.5)$} & --                       \\
        \gr
        \textbf{\ours}                & \cmark & $\mathbf{66.7}$\,{\tiny $(9.6)$} & $\mathbf{58.3}$\,{\tiny $(14.2)$} & $\mathbf{40.6}$\,{\tiny $(8.7)$} & $\mathbf{46.9}$\,{\tiny $(8.8)$} & $40.2$\,{\tiny $(5.2)$} & $\mathbf{50.5}$\,{\tiny $(4.4)$} \\
        \bottomrule
    \end{tabular}%
    }
    \end{minipage}
    \par}
    \caption{
    \textbf{MPC planning results.}
    We compare \ours\ with existing world-model baselines across simulation and robot domains.
    \ours\ achieves the highest average score in both tables.
    We report success rates (\%), except for the DROID action score~\citep{jepa-wms}, with standard errors in parentheses.
    RoboCasa is evaluated zero-shot; -R, -RW, and -P denote Reach, Reach-Wall, and Place, respectively. Joint-training denotes a single predictor trained across domains.
    Additionally, scores in \cref{fig:teaser}(c) are averaged over the benchmarks used to compare \ours\ with the generative world model baseline (\cref{sec:latent-vs-pixel}).
    }
    \label{tab:planning-benchmarks}
    \vspace{-.5em}
\end{table}

%% file: figure/phrasing-robustness.tex
\begin{wrapfigure}[10]{r}{0.23\textwidth}
    \vspace{-1.3em}
    \centering
    \includegraphics[width=1.0\linewidth]{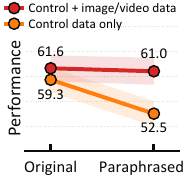}
    \vspace{-1.8em}
    \captionsetup{name=Fig., font=small, justification=raggedright, singlelinecheck=false}
    \caption{Benefits of image /video data inclusion.}
    \vspace{+.2em}
    \label{fig:phrasing-robustness}
\end{wrapfigure}

%% file: figure/latent-vs-pixel.tex
\begin{figure}[t]
    \centering
    \includegraphics[width=\linewidth]{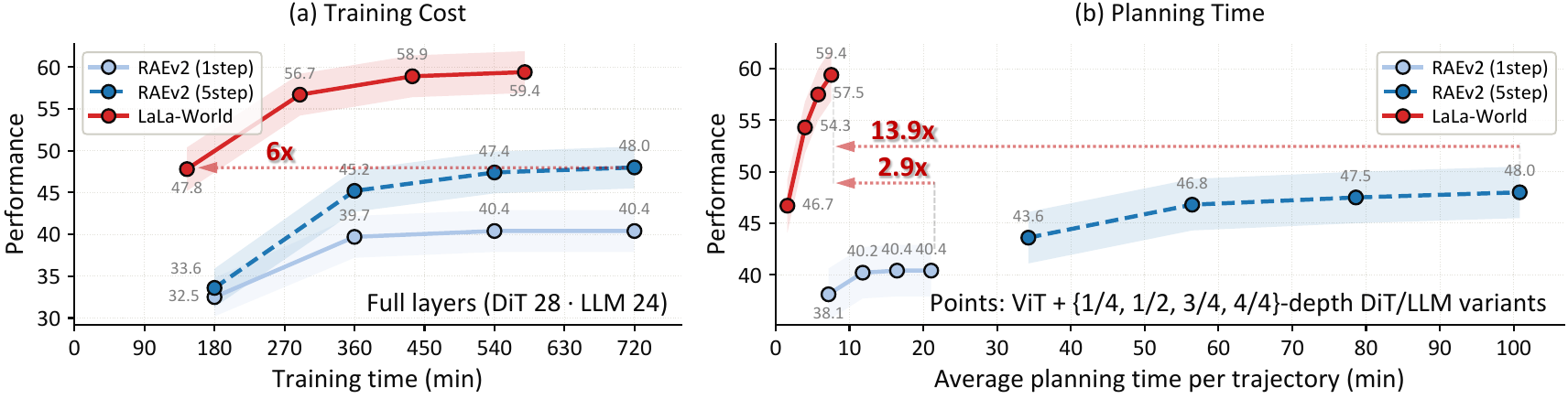}
    \vspace{-1.5em}
    \caption{
    \textbf{Latent vs. generative world models for MPC planning under matched setups.}
    We match the vision encoder, language conditions, and training data to compare the two world models.
    (a) We track planning performance over training time for full-depth predictors under a 12-hour training budget.
    (b) We compare planning performance and average planning time per trajectory across four predictor depths, with all models trained on the same number of transition data.
    These results show that \ours\ achieves higher planning performance than RAEv2 with lower training cost and planning time.
    }
    \vspace{-.2em}
    \label{fig:latent-vs-pixel}
\end{figure}

%% file: figure/imagination-visualization.tex
\begin{figure}[t]
    \centering
    \includegraphics[width=\linewidth]{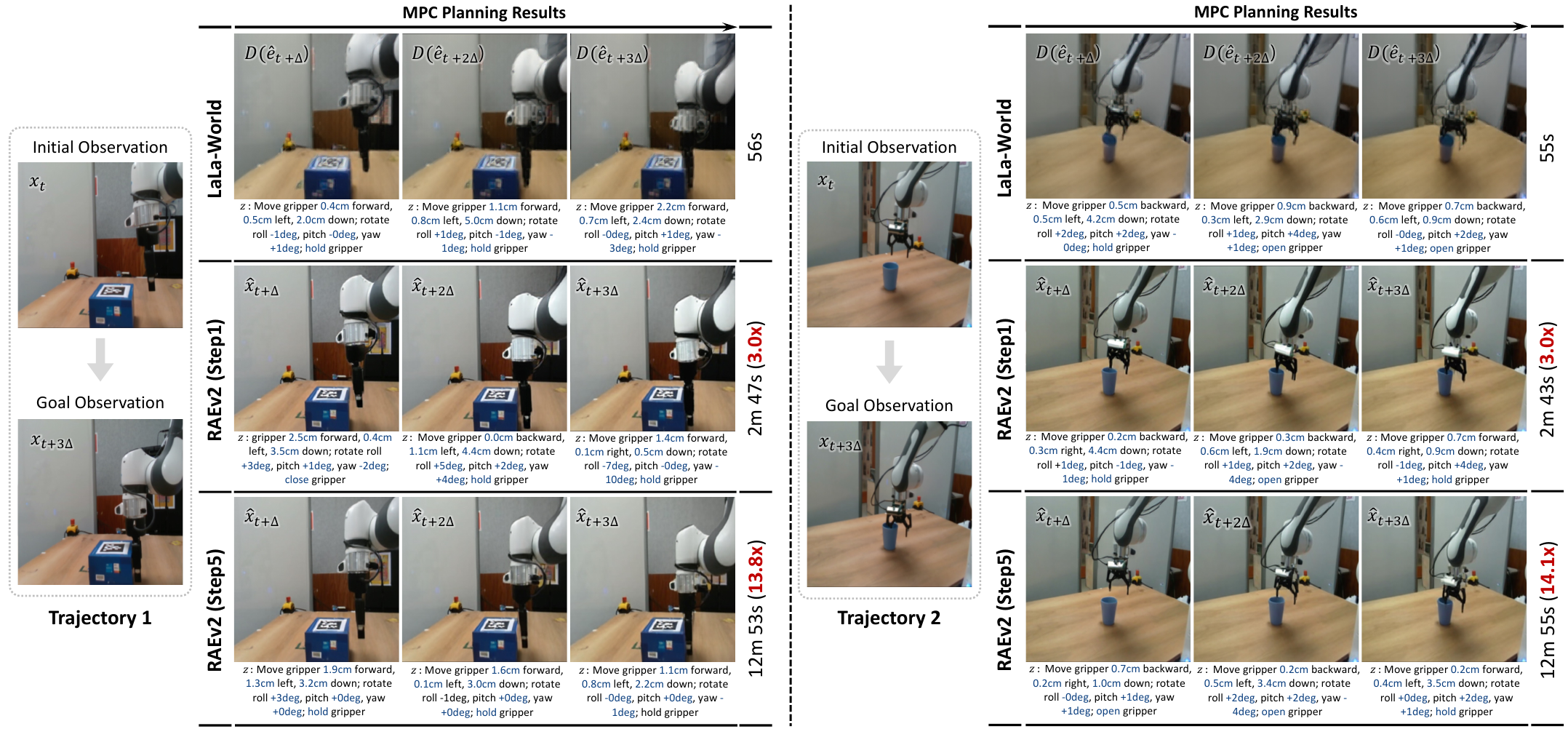}
    \vspace{-1.2em}
    \caption{
    \textbf{Visualization of predicted future states during MPC planning on DROID.}
    We visualize decoded predictions from \ours\ and RAEv2, using one or five denoising steps for RAEv2. Each example shows the initial and goal observations, the predicted rollout under the selected language actions, and planning time. Decoded images from \ours\ may have lower visual quality because it is not trained for pixel reconstruction.
    Both models predict plausible future states, while \ours\ requires less planning time.
    }
    \label{fig:imagination-visualization}
    \vspace{-1.em}
\end{figure}

%% file: table/policy-benchmarks.tex
\begin{table}[t]
    \centering
    \makeatletter
    \newcommand{\weightcite}[1]{%
        \nocite{#1}%
        \@ifundefined{b@#1\@extra@b@citeb}{\textbf{?}}{%
            \NAT@parse{#1}%
            \textsuperscript{\hyper@natlinkstart{#1\@extra@b@citeb}'\expandafter\@gobbletwo\NAT@date\hyper@natlinkend}}}
    \makeatother
    \begin{minipage}[t]{0.54\linewidth}
        \vspace{0pt}
        \centering
        {\scriptsize
        \renewcommand{\arraystretch}{1.25}
        \setlength{\tabcolsep}{1.5pt}
        \setlength{\aboverulesep}{0.5pt}
        \setlength{\belowrulesep}{0.5pt}
        \resizebox{\linewidth}{!}{%
        \begin{tabular}{@{}>{\arrayrulecolor{black!25}}w{l}{9.4em}:w{c}{8.em}:*{4}{w{c}{\dimexpr(\linewidth-14.8em-12\tabcolsep-3\arrayrulewidth)/5\relax}}:w{c}{\dimexpr(\linewidth-14.8em-12\tabcolsep-3\arrayrulewidth)/5\relax}@{}}
            \toprule
            Method & Pretrained weights & Spatial & Object & Goal & Long & Avg. \\
            \midrule
            Dita~(\citeyear{hou2025dita})                & DINOv2\weightcite{dinov2}+CLIP\weightcite{clip} & $97.4$          & $94.8$           & $93.2$          & $83.6$          & $92.3$ \\
            $\pi_0$~(\citeyear{black2024pi_0})          & PaliGemma\weightcite{paligemma} & $96.8$          & $98.8$           & $95.8$          & $85.2$          & $94.2$ \\
            UniVLA~(\citeyear{bu2025univla})            & Prismatic\weightcite{prismatic} & $96.5$          & $96.8$           & $95.6$          & $92.0$          & $95.2$ \\
            $\pi_{0.5}$~(\citeyear{intelligence2025pi}) & PaliGemma\weightcite{paligemma} & $\mathbf{98.8}$ & $98.2$           & $\mathbf{98.0}$ & $92.4$          & $96.9$ \\
            OpenVLA-OFT~(\citeyear{kim2025fine})        & Prismatic\weightcite{prismatic} & $97.6$          & $98.4$           & $97.9$          & $94.5$          & $97.1$ \\
            CogVLA~(\citeyear{li2025cogvla})            & Prismatic\weightcite{prismatic} & $98.6$          & $98.8$           & $96.6$          & $\mathbf{95.4}$ & $97.4$ \\
            \hdashline
            Cosmos Policy~(\citeyear{cosmos-policy})   & Random init. & $94.7$          & $98.9$           & $96.3$          & $88.6$          & $94.6$ \\
            Cosmos Policy~(\citeyear{cosmos-policy})   & Cosmos-Predict2\weightcite{cosmos-predict2} & $97.6$          & $99.8$           & $97.9$          & $94.0$          & $97.3$ \\
            \hdashline
            \ourpolicy                                & Random init. & $94.4$          & $99.6$           & $95.2$          & $87.2$          & $94.1$ \\
            \ourpolicy                                & Qwen3.5\weightcite{qwen35} & $96.8$          & $99.8$           & $96.0$          & $90.2$          & $95.7$ \\
            \gr
            \textbf{\ourpolicy}                       & \ours   & $98.6$          & $\mathbf{100.0}$ & $97.2$          & $95.0$          & $\mathbf{97.7}$ \\
            \bottomrule
        \end{tabular}%
        }
        }
        \vspace{-.3em}
        \caption{
        \textbf{LIBERO policy results.} We compare LaLa-Policy with existing policies on LIBERO. Pretrained weights indicates the source model for policy fine-tuning, while Random init. denotes random initialization. This table reports success rates (SR, \%).}
        \label{tab:policy-libero}
    \end{minipage}\hfill
    \begin{minipage}[t]{0.44\linewidth}
        \vspace{0pt}
        \centering
        {\scriptsize
        \renewcommand{\arraystretch}{1.25}
        \setlength{\tabcolsep}{1.5pt}
        \setlength{\aboverulesep}{0.5pt}
        \setlength{\belowrulesep}{0.5pt}
        \resizebox{\linewidth}{!}{%
        \begin{tabular}{@{}>{\arrayrulecolor{black!25}}w{l}{9.4em}:w{c}{8.em}:*{2}{w{c}{\dimexpr(\linewidth-14.8em-6\tabcolsep-3\arrayrulewidth)/2\relax}}@{}}
            \toprule
            Method & Pretrained weights & Demos & SR (\%) \\
            \midrule
            $\pi_0$~(\citeyear{black2024pi_0})          & PaliGemma\weightcite{paligemma} & 300 & $62.5$ \\
            GR00T-N1.5~(\citeyear{issac-gr00t})     & Eagle 2.5\weightcite{eagle25} & 300 & $64.1$ \\
            Video Policy~(\citeyear{liang2025video})    & Stable Video Diff.\weightcite{stablevideodiffusion} & 300 & $66.0$ \\
            \hdashline
            \ourpolicy                                & Random init. & 300 & $65.1$ \\
            \gr
            \textbf{\ourpolicy}                       & \ours   & 300 & $\mathbf{66.4}$ \\
            \midrule
            UVA~(\citeyear{li2025unified})              & MAR\weightcite{mar} & 50  & $50.0$ \\
            $\pi_{0.5}$~(\citeyear{intelligence2025pi}) & PaliGemma\weightcite{paligemma} & 50  & $54.5$ \\
            MolmoAct2~(\citeyear{molmoact2})            & Molmo2-ER\weightcite{molmoact2} & 50  & $54.3$ \\
            \hdashline
            Cosmos Policy~(\citeyear{cosmos-policy})   & Random init. & 50  & $53.8$ \\
            Cosmos Policy~(\citeyear{cosmos-policy})   & Cosmos-Predict2\weightcite{cosmos-predict2} & 50  & $56.0$ \\
            \hdashline
            \ourpolicy                                & Random init. & 50  & $55.5$ \\
            \ourpolicy                                & Qwen3.5\weightcite{qwen35} & 50  & $56.8$ \\
            \gr
            \textbf{\ourpolicy}                       & \ours   & 50  & $\mathbf{57.3}$ \\
            \bottomrule
        \end{tabular}%
        }
        }
        \vspace{-.3em}
        \caption{
        \textbf{RoboCasa policy results.} We evaluate policy models on RoboCasa. Demos denotes the number of training trajectories.}
        \label{tab:policy-robocasa}
    \end{minipage}
    \vspace{-1.em}
\end{table}

%% file: figure/policy-visualization.tex
\begin{figure}[t]
    \centering
    \includegraphics[width=\linewidth]{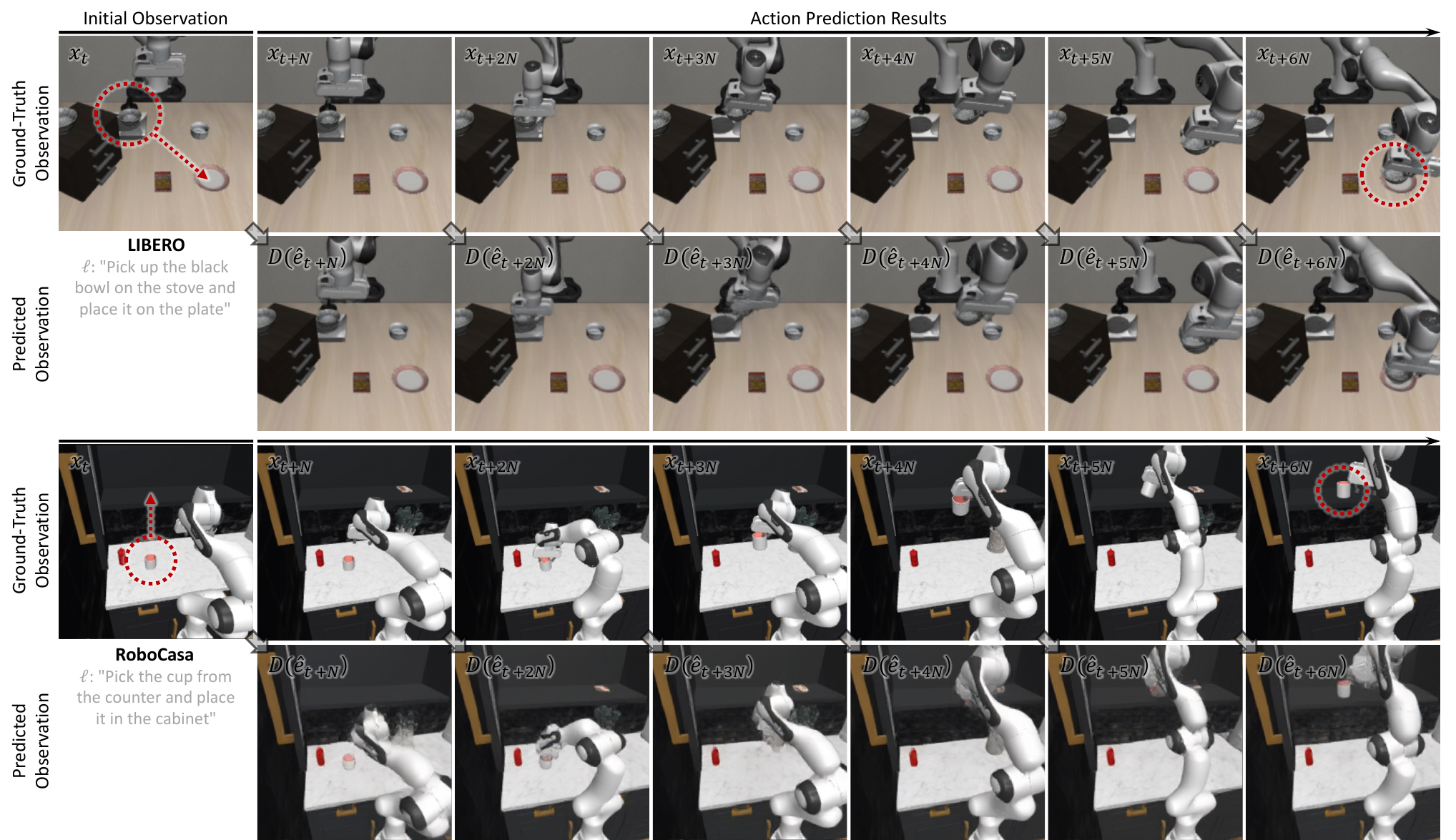}
    \vspace{-1.em}
    \caption{
    \textbf{Predicted future observations on LIBERO and RoboCasa.}
    Each example shows the initial observation and task instruction, followed by simulator observations during policy execution (upper row) and decoded future predictions (lower row). 
    \ourpolicy{} predicts $(x_t, \ell) \xrightarrow{} (\hat{e}_{t+N}, \hat{a}_{t:t+N-1})$, and the simulator executes $\hat{a}_{t:t+N-1}$ to yield $x_{t+N}$; see Equation~\ref{eq:lala-policy}.
    Since \ourpolicy\ is not trained for pixel reconstruction, decoded images may contain visual inaccuracies.
    }
    \label{fig:policy-visualization}
    \vspace{-.5em}
\end{figure}

%% file: table/policy-inference-time.tex
\begin{wraptable}[6]{r}{0.28\textwidth}
    \vspace{-1.2em}
    \centering
    {\scriptsize
    \renewcommand{\arraystretch}{1.1}
    \setlength{\tabcolsep}{2pt}
    \setlength{\aboverulesep}{0.5pt}
    \setlength{\belowrulesep}{0.5pt}
    \begin{tabular}{@{}lcc@{}}
        \toprule
                      & LIBERO                & RoboCasa \\
        \midrule
        Cosmos Policy & $4.45$\,s             & $11.71$\,s \\
        \hdashline
        \multirow{2}{*}{\ourpolicy}
                      & $\mathbf{0.86}$\,s    & $\mathbf{2.12}$\,s \\
                      & {\tiny ($\times 5.2$)}        
                      & {\tiny ($\times 5.5$)} \\
        \bottomrule
    \end{tabular}
    }
    \captionsetup{name=Tab., font=small, justification=raggedright, singlelinecheck=false, skip=3pt}
    \caption{Avg. inference time per trajectory and speedup.}
    \label{tab:policy-inference-time}
    \vspace{-.8em}
\end{wraptable}

%% file: figure/relate-work-table.tex
\begin{figure}[h]
    \centering
    \includegraphics[width=\linewidth]{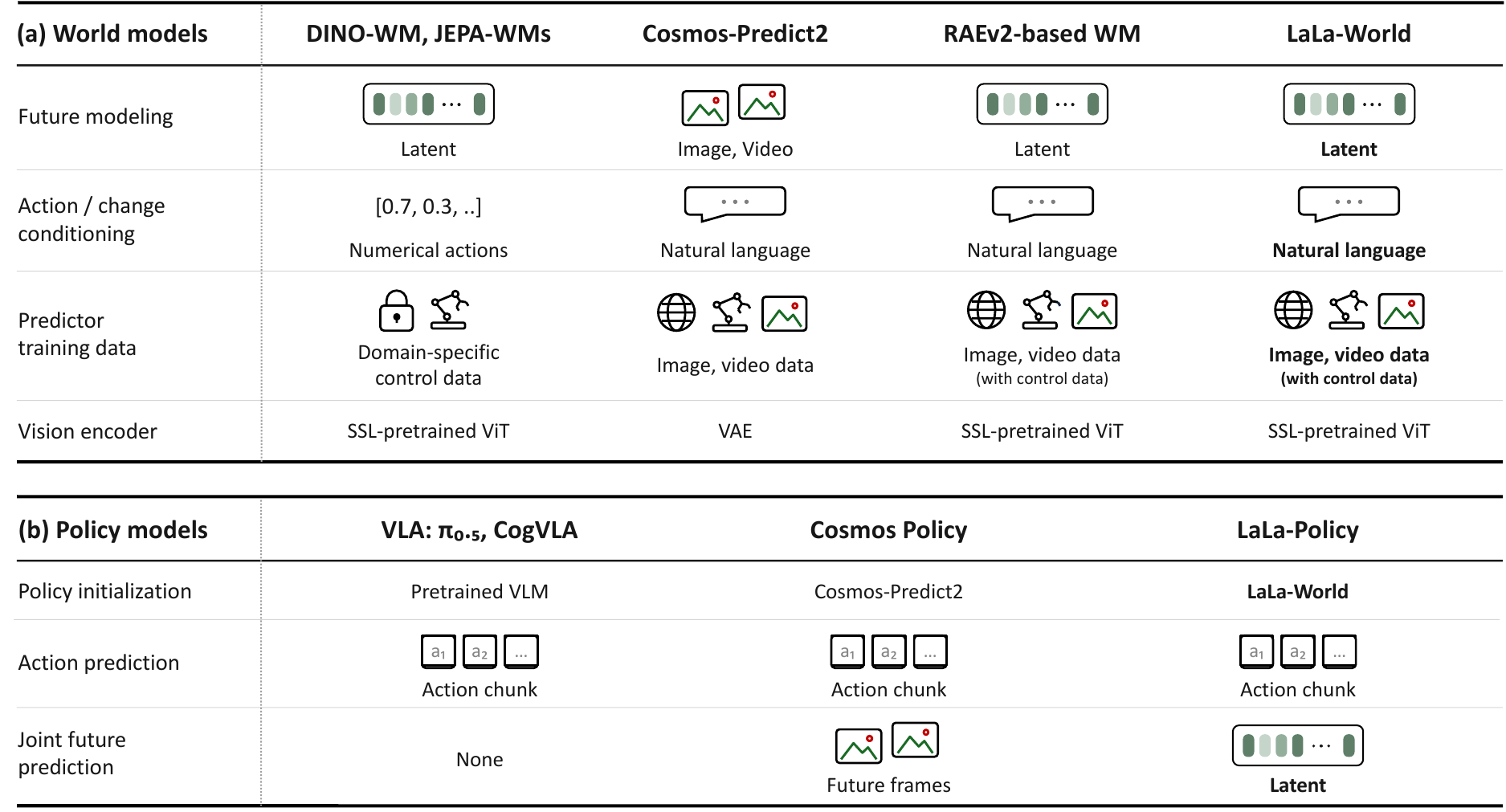}
    \vspace{-1.5em}
    \caption{
    \textbf{Comparison with representative world models and policy models.}
    (a) \ours\ uses natural language conditions, enabling training on control trajectories together with image and video transitions.
    The RAEv2-based WM is our generative world model baseline under matched settings in \cref{sec:latent-vs-pixel}.
    (b) Fine-tuned from \ours, \ourpolicy\ jointly predicts actions and future latents, whereas VLAs predict actions alone and Cosmos Policy predicts actions and the future with visual details.
    }
    \label{fig:relate-work-table}
    \vspace{-.5em}
\end{figure}

%% file: figure/imagenet-train-eval.tex
\begin{figure}[t]
    \centering
    \includegraphics[width=\linewidth]{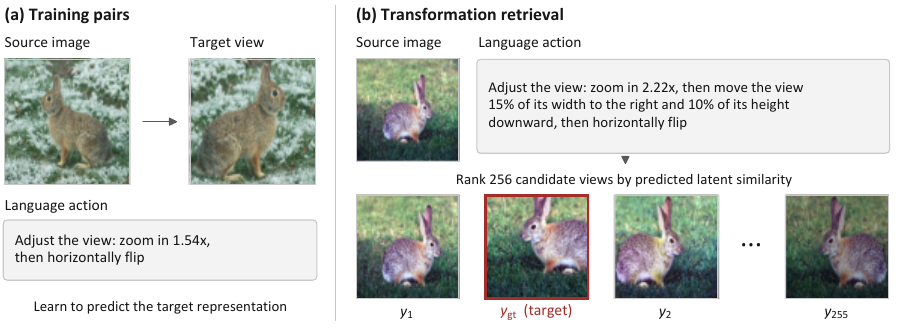}
    \vspace{-2em}
    \caption{
    \textbf{ImageNet training pairs and transformation retrieval.}
    Left: a source image and a transformed view form a training pair, with the transformation described in language.
    Right: the predicted representation is used to rank 256 transformed views of the same validation image; a subset is illustrated.
    The outlined view is the target specified by the language action.
    }
    \label{fig:imagenet-train-eval}
\end{figure}

%% file: table/imagenet-mrr.tex
\begin{table}[t]
    \centering
    {\scriptsize
    \renewcommand{\arraystretch}{1.2}
    \setlength{\tabcolsep}{6pt}
    \setlength{\aboverulesep}{0.5pt}
    \setlength{\belowrulesep}{0.5pt}
    \begin{tabular}{@{}lc@{}}
        \toprule
        Method & MRR $\uparrow$ \\
        \midrule
        IWM (default)$^{\dagger}$ & $0.79$ \\
        IWM (strong jitter)$^{\dagger}$ & $0.85$ \\
        \hdashline
        \multicolumn{2}{@{}l}{\textbf{\ours}} \\
        Source latent & $0.024$ \\
        Predicted latents, empty language & $0.026$ \\
        Predicted latents, language action & $\mathbf{0.856}$ \\
        \bottomrule
    \end{tabular}
    }
    \vspace{-.5em}
    \caption{
        \textbf{ImageNet transformation retrieval.}
        We report mean reciprocal rank (MRR) over 256 transformed views of each source image.
        The \ours\ rows compare retrieval using the source latent with predicted latents conditioned on empty language or a language action.
        $^{\dagger}$Published IWM results~\citep{iwm} use different training and transformation settings.
    }
    \label{tab:imagenet-mrr}
    \vspace{-.5em}
\end{table}

%% file: figure/communicable-setup.tex
\begin{figure}[t]
    \centering
    \begingroup
    \definecolor{cwmred}{HTML}{C00000}
    \definecolor{cwmpink}{HTML}{FAE3E3}
    \definecolor{cwmblue}{HTML}{0F4281}
    \definecolor{cwmsky}{HTML}{E4ECF6}
    \definecolor{cwmgraytext}{HTML}{7F7F7F}
    \definecolor{cwmgrayfill}{HTML}{F2F2F2}
    \definecolor{cwmgrayline}{HTML}{BFBFBF}
    \resizebox{\linewidth}{!}{%
    \begin{tikzpicture}[
        x=1cm, y=1cm,
        font=\sffamily\fontsize{8}{10}\selectfont, text=black,
        cardtitle/.style={font=\sffamily\bfseries\fontsize{8.5}{10}\selectfont},
        bodytext/.style={font=\sffamily\fontsize{6.8}{8.6}\selectfont},
        note/.style={font=\sffamily\fontsize{6.2}{7.6}\selectfont, text=cwmgraytext},
        chat/.style={font=\ttfamily\fontsize{5.2}{6.5}\selectfont,
            anchor=north west, inner sep=0pt},
        messageheading/.style={bodytext, anchor=north west, inner sep=0pt},
        card/.style={rounded corners=5pt, line width=1pt},
        bubble/.style={rounded corners=4pt, line width=0.7pt, fill=white},
        pill/.style={rounded corners=8pt, line width=0.7pt},
        flow/.style={-{Stealth[length=1.8mm]}, line width=0.9pt}
    ]
        \path[use as bounding box] (0,0.15) rectangle (14.8,6.45);

        \node[note,anchor=south] at (1.05,6.13) {initial};
        \node[note,anchor=south] at (1.05,4.33) {goal};
        \foreach \yy/\dx/\dy in {4.7/1.42/5.7, 2.9/0.62/3.6} {
            \path[draw=cwmgrayline, fill=white, line width=0.6pt] (0.35,\yy) rectangle (1.75,\yy+1.4);
            \path[fill=black] (0.98,\yy) rectangle (1.12,\yy+0.5);
            \path[fill=black] (0.98,\yy+0.9) rectangle (1.12,\yy+1.4);
            \path[fill=cwmred] (\dx,\dy) circle (0.07);
        }
        \draw[flow, cwmgraytext] (1.9,4.5) -- (2.4,4.5);

        \path[card, draw=cwmred, fill=cwmpink] (2.5,2.5) rectangle (6.1,6.1);
        \node[cardtitle, text=cwmred, anchor=north] at (4.3,5.95) {LLM actor};
        \node[note, anchor=north] at (4.3,5.58) {in place of CEM};
        \node[bodytext, text width=3.15cm, align=left, anchor=north] at (4.3,5.2)
            {Reads the initial and goal images.\\[3pt]
             Writes $N$ candidate plans as language actions.\\[3pt]
             Revises them from the returned costs.};
        \node[note, anchor=south] at (4.3,2.62) {Claude Opus / GPT};

        \path[card, draw=cwmblue, fill=cwmsky] (10.9,2.5) rectangle (14.5,6.1);
        \node[cardtitle, text=cwmblue, anchor=north] at (12.7,5.95) {\ours};
        \node[note, anchor=north] at (12.7,5.58) {critic};
        \node[bodytext, text width=3.15cm, align=left, anchor=north] at (12.7,5.2)
            {Imagines each plan in latent space.\\[6pt]
             Returns the goal-distance cost:\\[5pt]
             $\|\hat{e}_{t+H}-e_{\mathrm{goal}}\|^2$.};

        \path[bubble, draw=cwmred] (6.3,4.45) rectangle (10.7,5.95);
        \path[fill=white] (6.3,5.45) -- (6.02,5.25) -- (6.3,5.05) -- cycle;
        \draw[cwmred, line width=0.7pt] (6.3,5.45) -- (6.02,5.25) -- (6.3,5.05);
        \draw[white, line width=1.2pt] (6.3,5.42) -- (6.3,5.08);
        \node[messageheading, text=cwmred] at (6.52,5.79) {\textbf{$N$ plans}};
        \node[chat] at (6.52,5.44) {%
            \renewcommand{\arraystretch}{1.15}%
            \begin{tabular}{@{}r@{:\ }l@{\ }r@{\ }l@{\ }l@{\ }r@{\ }l@{}}
                1 & move & 0.6 & left  & and & 1.7 & down \ldots\\
                2 & move & 1.2 & left  & and & 0.9 & down \ldots\\
                3 & move & 0.0 & right & and & 1.7 & down \ldots
            \end{tabular}};
        \node[note, anchor=base west, inner sep=0pt] at (6.52,4.57) {\ldots};

        \path[bubble, draw=cwmblue] (6.3,2.5) rectangle (10.7,4.0);
        \path[fill=white] (10.7,3.45) -- (10.98,3.25) -- (10.7,3.05) -- cycle;
        \draw[cwmblue, line width=0.7pt] (10.7,3.45) -- (10.98,3.25) -- (10.7,3.05);
        \draw[white, line width=1.2pt] (10.7,3.42) -- (10.7,3.08);
        \node[messageheading, text=cwmblue] at (6.52,3.84) {\textbf{Cost of each plan}};
        \node[chat] at (6.52,3.49) {%
            \renewcommand{\arraystretch}{1.15}%
            \begin{tabular}{@{}r@{:\ }r@{}}
                1 & 0.046\\
                2 & 0.052\\
                3 & 0.061
            \end{tabular}};
        \node[note, anchor=base west, inner sep=0pt] at (6.52,2.62) {\ldots};
        \node[note, anchor=base east, inner sep=0pt] at (10.48,2.72) {(lower is better)};

        \node[note, inner sep=0pt] at (8.5,4.225) {$\circlearrowright$ repeated for $K$ rounds};

        \draw[flow, cwmblue] (12.7,2.5) -- (12.7,1.8);
        \path[card, draw=cwmgrayline, fill=cwmgrayfill, line width=0.7pt] (10.9,0.95) rectangle (14.5,1.75);
        \node[bodytext, align=center] at (12.7,1.35) {Simulator executes\\the lowest-cost plan};
        \draw[flow, cwmgraytext, dashed] (12.7,0.95) -- (12.7,0.5) -- (1.05,0.5) -- (1.05,2.75);
        \node[note, fill=white, inner sep=2pt] at (5.4,0.5) {replan (if \#horizon > executed steps)};

        \path[pill, draw=cwmgrayline, fill=cwmgrayfill] (2.5,0.95) rectangle (6.1,1.75);
        \node[bodytext, align=center] at (4.3,1.35)
            {\textbf{CEM}: $300\times 30$ samples\\$=\mathbf{9{,}000}$ \textbf{rollouts} per plan};
        \path[pill, draw=cwmred, fill=white] (6.3,0.95) rectangle (10.7,1.75);
        \node[bodytext, align=center] at (8.5,1.35)
            {\textbf{LLM actor}: $30\times 3$ plans\\$=\mathbf{90}$ \textbf{rollouts} per plan};
        \node[note, text=cwmred, anchor=south] at (8.5,1.8) {$100\times$ fewer world-model calls};
    \end{tikzpicture}%
    }
    \endgroup
    \vspace{-2em}
    \caption{
    \textbf{Communicable world modeling.}
    An LLM takes the place of CEM and talks to \ours\ in language.
    It reads the initial and goal observations, writes $N$ candidate plans as language actions (top message), and receives only the goal-distance cost of each plan (bottom message), which is the same cost that CEM minimizes.
    After $K$ rounds, the lowest-cost plan is executed.
    The LLM needs two orders of magnitude fewer world-model rollouts than CEM.
    }
    \label{fig:communicable-setup}
\end{figure}
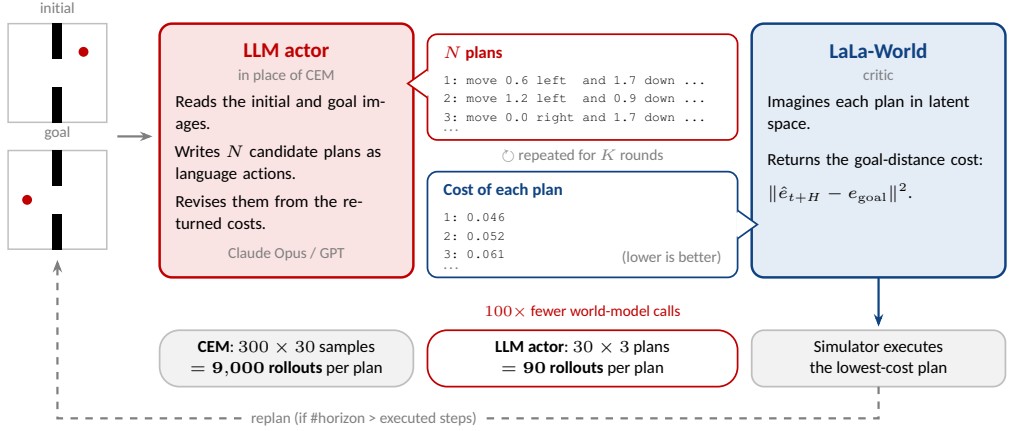

%% file: table/communicable-world-modeling.tex
\begin{table}[t]
    \centering
    {\fontsize{8}{9.5}\selectfont
    \renewcommand{\arraystretch}{1.3}
    \setlength{\aboverulesep}{0.5pt}
    \setlength{\belowrulesep}{0.5pt}
    \setlength{\tabcolsep}{5pt}
    \begin{tabular}{@{}w{l}{0.18\linewidth}*{4}{w{c}{\dimexpr(0.82\linewidth-8\tabcolsep)/4\relax}}@{}}
        \toprule
        Planner & Wall & PointMaze & RoboCasa-R & DROID \\
        \midrule
        CEM           & $95.8\,(9{,}000)$          & \textbf{56.2} $\,(9{,}000)$          & $40.6\,(54{,}000)$          & $40.2\,(4{,}500)$          \\
        Claude Opus 5 & $\mathbf{100.0}\,(30{\times}3{=}90)$ & $31.2\,(30{\times}3{=}90)$ & $\mathbf{53.1}\,(10{\times}2{\times}4{=}80)$ & $36.2\,(20{\times}2{=}40)$ \\
        GPT-6 Astra   & $91.3\,(8{\times}1{=}8)$            & $26.1\,(8{\times}3{=}24)$  & $17.4\,(8{\times}1{\times}12{=}96)$          & $\mathbf{44.7}\,(16{\times}3{=}48)$ \\
        \bottomrule
    \end{tabular}
    }
    \vspace{-0.5em}
    \caption{
        \textbf{MPC planning with an LLM actor in place of CEM.}
    We report success rates (\%), except for the DROID action score, with the number of world-model rollouts per trajectory in parentheses.
    For the LLM actors, the rollout count is written as $N \times K$, e.g., $30 \times 3 = 90$, where $N = 30$ is the number of candidate action sequences proposed per round and $K = 3$ is the number of rounds; for RoboCasa-R, the third factor is the number of replanning steps per trajectory.
    The LLM actors use 8--96 rollouts per trajectory, compared with 4,500--54,000 for CEM.
}
\label{tab:communicable-world-modeling}
    \vspace{-1em}
\end{table}

%% file: figure/communicable-qualitative.tex
\begin{figure}[t]
    \centering
    \includegraphics[width=\linewidth]{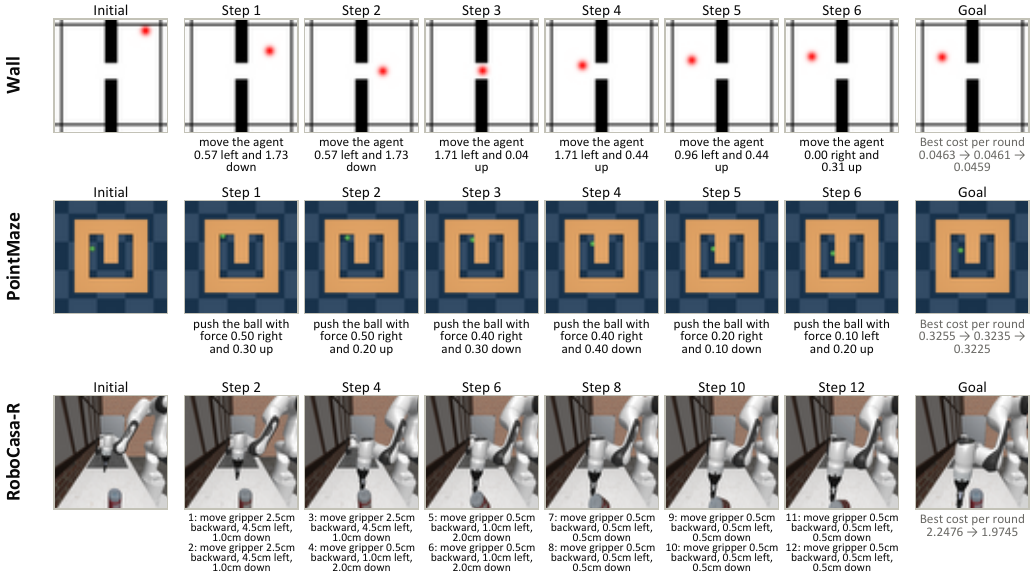}
    \vspace{-1.3em}
    \caption{
    \textbf{Qualitative Results of Communicable World Modeling with an LLM Actor.}
    Each row shows the initial and goal observations and the simulator observations after executing the lowest-cost sequence written by the LLM, with its language action below each frame.
    Below the goal observation, we list the cost of the best sequence at each round, which decreases as the LLM revises its plans from the returned costs.
    }
    \label{fig:communicable-qualitative}
    \vspace{-.5em}
\end{figure}

%% file: figure/llm-ablation-1.tex
\begin{figure}[t]
    \centering
    \includegraphics[width=\linewidth]{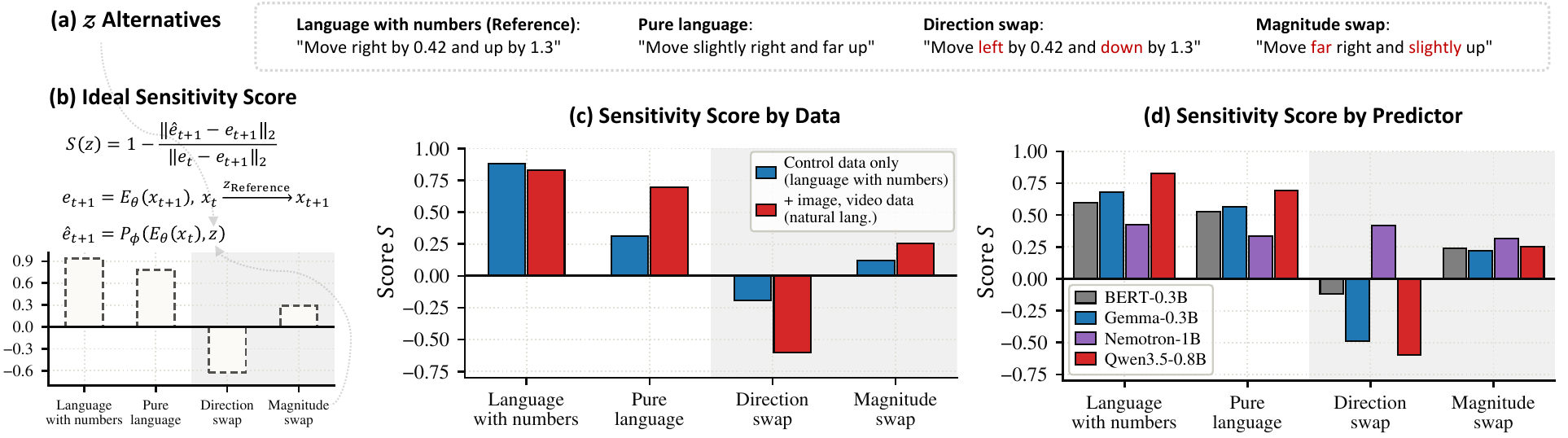}
    \vspace{-1.5em}
    \caption{
    \textbf{Language sensitivity across training data and predictor backbones.}
    (a) Language conditions used to evaluate the same target transition; red words mark semantic swaps.
    (b) Expected score pattern for a predictor that follows language.
    (c) Effect of adding image and video transitions to control data.
    (d) Comparison of language model backbones.
    }
    \label{fig:llm-ablation-1}
    \vspace{-.5em}
\end{figure}

%% file: figure/llm-ablation-2.tex
\begin{figure}[t]
    \centering
    \includegraphics[width=\linewidth]{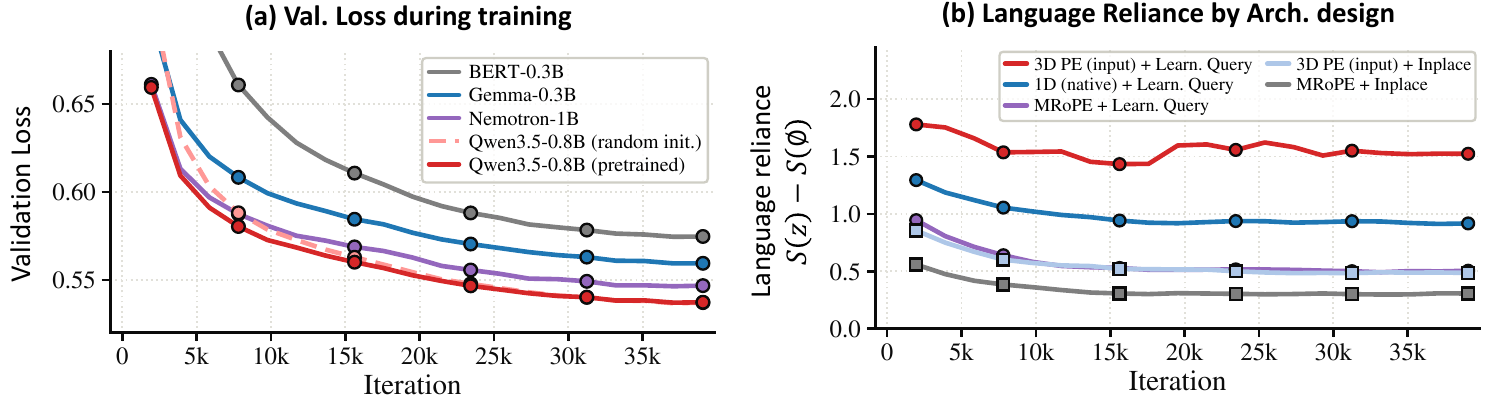}
    \vspace{-1.5em}
    \caption{
    \textbf{Predictor backbone, initialization, and architecture ablations.}
    (a) Validation loss for language model backbones, including pretrained and randomly initialized Qwen3.5.
    (b) Language reliance $S(z)-S(\varnothing)$ for positional encoding and readout variants.
    3D input positional embeddings with learned queries yield the highest reliance.
    }
    \label{fig:llm-ablation-2}
    \vspace{-1.em}
\end{figure}

%% file: figure/arch-examples.tex
\begin{figure}[t]
\centering
\begingroup
\newcommand{\acfigurefont}{\sffamily}
\ifdefined\IfFontExistsTF
  \IfFontExistsTF{Calibri}{\renewcommand{\acfigurefont}{\fontspec{Calibri}}}{}
\fi
\definecolor{acblue}{HTML}{0F4281}
\definecolor{acsky}{HTML}{E4ECF6}
\definecolor{acred}{HTML}{C00000}
\definecolor{acpink}{HTML}{FAE3E3}
\definecolor{acorange}{HTML}{FF7F0E}
\definecolor{acpeach}{HTML}{FFE8D1}
\definecolor{acink}{HTML}{202020}
\definecolor{acmuted}{HTML}{6A6A6A}
\definecolor{acline}{HTML}{BFBFBF}
\definecolor{acgray}{HTML}{F2F2F2}
\definecolor{acpale}{HTML}{F4F4F4}
\resizebox{\linewidth}{!}{%
\begin{tikzpicture}[
  x=1cm,y=1cm,
  every node/.style={inner sep=0pt,outer sep=0pt,align=center,text=acink,
    font={\acfigurefont\fontsize{6.7}{8.4}\selectfont}},
  actitle/.style={anchor=west,align=left,
    font={\acfigurefont\bfseries\fontsize{7.44}{9.3}\selectfont}},
  acheading/.style={anchor=west,align=left,
    font={\acfigurefont\bfseries\fontsize{8.2}{10.25}\selectfont}},
  acnote/.style={text=acmuted,font={\acfigurefont\fontsize{6.4}{8}\selectfont}},
  accoord/.style={fill=acpale,draw=none,rounded corners=2.5pt},
  actoken/.style={draw=acblue,fill=acsky,line width=.65pt,
    rounded corners=1.1pt,minimum width=.50cm,minimum height=.44cm},
  aclanguage/.style={actoken,draw=acred,fill=acpink,minimum width=.83cm},
  acquery/.style={actoken,draw=acorange,fill=acpeach},
  acblock/.style={draw=acblue,fill=acsky,text=acblue,line width=.65pt,
    rounded corners=1.3pt,font={\acfigurefont\fontsize{7}{8.75}\selectfont}},
  acbackbone/.style={acblock,draw=black,fill=acgray,text=black},
  acmlp/.style={draw=acline,fill=acgray,rounded corners=1.3pt,line width=.65pt,
    minimum width=1.34cm,minimum height=.60cm},
  acarrow/.style={draw=acblue,line width=.65pt,line cap=round,line join=round,
    -{Stealth[length=1.35mm,width=1.0mm]}}
]
\path[use as bounding box] (0,0) rectangle (15.8,7.3);

\node[actitle] at (.18,7.10) {(a) 3D input PE + learned queries};
\node[acnote,anchor=east] at (15.60,7.10) {one context frame shown};

\foreach \x/\y/\i in {.82/4.02/1,1.26/4.02/2,.82/3.58/3,1.26/3.58/4} {
  \node[actoken,rounded corners=0pt,minimum width=.44cm,
    font={\acfigurefont\fontsize{7.2}{9}\selectfont}] at (\x,\y) {$v_{\i}$};
}
\node[font={\acfigurefont\bfseries\fontsize{7.3}{9.1}\selectfont}]
  at (1.04,4.48) {Current state};
\node[acmlp] (aligner) at (2.40,3.80) {MLP\\Aligner};
\draw[acarrow] (1.52,3.80) -- (aligner.west);
\draw[acarrow] (aligner.east) -- (3.65,3.80);

\foreach \x/\i in {3.90/1,4.61/2,5.32/3,6.03/4} {
  \node[actoken] at (\x,3.80) {$v_{\i}$};
}
\node[aclanguage] at (7.395,3.80) {move};
\node[aclanguage] at (8.435,3.80) {right};
\foreach \x in {9.80,10.51,11.22,11.93} {
  \node[acquery] at (\x,3.80) {$q$};
}
\node[text=acblue,font={\acfigurefont\fontsize{7.2}{9}\selectfont}]
  at (4.965,3.34) {State tokens};
\node[text=acred,font={\acfigurefont\fontsize{7.2}{9}\selectfont}]
  at (7.915,3.34) {Language action};
\node[text=acorange,font={\acfigurefont\fontsize{7.0}{8.75}\selectfont}]
  at (10.865,3.34) {Learned queries};

\foreach \y/\axis in {4.66/t,4.42/h,4.18/w} {
  \path[accoord] (3.60,\y-.11) rectangle (12.23,\y+.11);
  \node at (3.36,\y) {$\axis$};
}
\foreach \x/\t/\h/\w in {
  3.90/0/0/0,4.61/0/0/1,5.32/0/1/0,6.03/0/1/1,
  7.395/1/1/1,8.435/2/2/2,
  9.80/3/0/0,10.51/3/0/1,11.22/3/1/0,11.93/3/1/1} {
  \node[acnote] at (\x,4.66) {\t};
  \node[acnote] at (\x,4.42) {\h};
  \node[acnote] at (\x,4.18) {\w};
  \draw[draw=acline,line width=.5pt] (\x,4.78) -- (\x,4.88);
}
\node[acblock,minimum width=9.12cm,minimum height=.36cm]
  (pe) at (8.00,5.06) {Add 3D PE};
\node[acbackbone,minimum width=9.12cm,minimum height=.56cm]
  (backbone) at (8.00,5.76)
  {Language-model backbone\quad {\color{acmuted}(native positions retained)}};
\draw[acarrow] (pe.north) -- (backbone.south);

\foreach \x/\i in {9.80/1,10.51/2,11.22/3,11.93/4} {
  \node[acquery,minimum height=.43cm] (readout\i) at (\x,6.44) {$h_{\i}^{q}$};
  \draw[acarrow,draw=acorange] (\x,6.04) -- (readout\i.south);
}
\node[acmlp] (head) at (13.45,6.44)
  {Prediction\\head};
\draw[acarrow,draw=acorange] (12.26,6.44) -- (head.west);
\draw[acarrow] (head.east) -- (14.51,6.44);
\foreach \x/\y/\i in {14.815/6.655/1,15.245/6.655/2,14.815/6.225/3,15.245/6.225/4} {
  \node[actoken,rounded corners=0pt,minimum width=.43cm,minimum height=.43cm,
    font={\acfigurefont\fontsize{7.2}{9}\selectfont}] at (\x,\y) {$\hat e_{\i}$};
}
\node[text=acblue,font={\acfigurefont\fontsize{6.2}{7.75}\selectfont}]
  at (15.03,5.80) {future latents};

\node[actitle] at (.18,2.78) {(b) Ablation variants};
\draw[draw=acline,line width=.4pt] (5.285,.06) -- (5.285,2.43);
\draw[draw=acline,line width=.4pt] (10.515,.06) -- (10.515,2.43);
\node[acheading] at (.36,2.39) {1D};
\node[acheading] at (5.59,2.39) {MRoPE};
\node[acheading] at (10.82,2.39) {Inplace};
\node[acnote,anchor=west] at (.36,2.04) {Native 1D positions for all input tokens.};
\node[acnote,anchor=west] at (5.59,2.04) {3D rotary encoding within attention layers.};
\node[acnote,anchor=west] at (10.82,2.04) {Read state outputs without query tokens.};

\path[accoord] (.72,.55) rectangle (4.60,.79);
\node[acbackbone,minimum width=3.88cm,minimum height=.44cm]
  at (2.66,1.14) {Backbone};
\foreach \x/\i/\pos in {1.035/1/0,1.695/2/1} {
  \node[actoken,minimum width=.47cm,minimum height=.40cm] at (\x,.26) {$v_{\i}$};
  \node[acnote] at (\x,.67) {\pos};
  \draw[acarrow] (\x,.81) -- (\x,.92);
}
\foreach \x/\word/\pos in {2.58/move/2,3.55/right/3} {
  \node[aclanguage,minimum width=.82cm,minimum height=.40cm] at (\x,.26) {\word};
  \node[acnote] at (\x,.67) {\pos};
  \draw[acarrow] (\x,.81) -- (\x,.92);
}

\node[actoken,minimum width=.44cm,minimum height=.40cm] at (8.06,.26) {$v_1$};
\node[aclanguage,minimum height=.40cm] at (8.895,.26) {move};
\node[acbackbone,minimum width=2.48cm,minimum height=.90cm]
  at (8.66,1.28) {};
\node[text=black,font={\acfigurefont\fontsize{7}{8.75}\selectfont}]
  at (8.66,1.52) {Backbone};
\node[text=black,font={\acfigurefont\fontsize{6.4}{8}\selectfont}]
  at (8.66,1.07) {Rotate $Q$ and $K$};
\draw[acarrow] (8.06,.49) -- (8.06,.83);
\draw[acarrow] (8.895,.49) -- (8.895,.83);
\node[accoord,minimum width=1.26cm,minimum height=.40cm,inner xsep=.10cm,
  font={\acfigurefont\fontsize{7.7}{9.6}\selectfont}]
  at (6.29,1.07) {$(t,h,w)$};
\draw[acarrow] (6.97,1.07) -- (7.42,1.07);

\foreach \x/\i in {11.295/1,11.765/2,12.235/3,12.705/4} {
  \node[actoken,minimum width=.39cm,minimum height=.40cm] at (\x,.26) {$v_{\i}$};
  \node[acquery,minimum width=.39cm,minimum height=.40cm]
    (inplace\i) at (\x,1.63) {$h_{\i}$};
  \draw[acarrow,draw=acorange] (\x,1.23) -- (inplace\i.south);
}
\node[aclanguage,minimum width=1.52cm,minimum height=.40cm,
  font={\acfigurefont\fontsize{6}{7.5}\selectfont}]
  at (14.04,.26) {move right};
\node[acbackbone,minimum width=3.95cm,minimum height=.44cm]
  (inplacebackbone) at (13.075,1.01) {Backbone};
\foreach \x in {11.295,11.765,12.235,12.705,14.04} {
  \draw[acarrow] (\x,.49) -- (\x,.79);
}
\end{tikzpicture}%
}
\endgroup
\vspace{-1em}
\caption{
    \textbf{Predictor positional encoding and readout.} (a) Projected state patches, language tokens and copies of one shared learned query vector receive additive 3D positional embeddings. The prediction head maps query outputs to future patch representations. (b) Native 1D uses sequence indices for visual and text tokens, handled by the backbone's positional mechanism. MRoPE uses 3D coordinates to rotate attention queries and keys inside the backbone. Inplace readout uses state-token hidden states.}
\label{fig:arch-examples}
\end{figure}
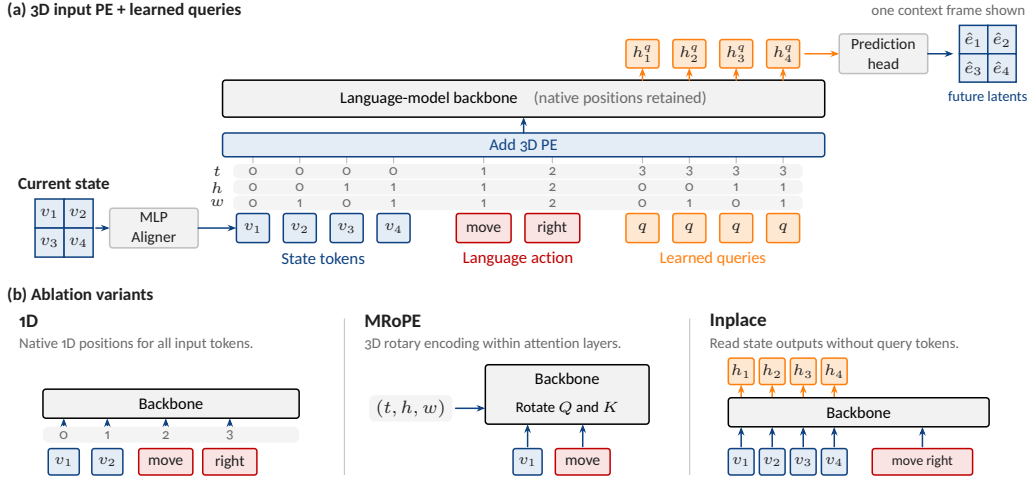

%% file: table/predictor-initialization-joint-training.tex
\begin{table}[t]
    \centering
    {\footnotesize
    \renewcommand{\arraystretch}{1.2}
    \setlength{\tabcolsep}{4pt}
    \setlength{\aboverulesep}{0.5pt}
    \setlength{\belowrulesep}{0.5pt}
    \setlength{\arrayrulewidth}{0.8pt}
    \setlength{\dashlinedash}{0.6pt}
    \setlength{\dashlinegap}{1.5pt}
    \resizebox{\linewidth}{!}{%
    \begin{tabular}{>{\arrayrulecolor{black!25}}ll:*{6}{c}:c}
        \toprule
        Initialization & Training scope & Wall & PointMaze & Push-T & DROID & RoboCasa-R & RoboCasa-P & Average \\
        \midrule
        Pretrained LLM & Single-domain
        & $\mathbf{97.9}$\,{\tiny $(1.5)$} & $\mathbf{66.7}$\,{\tiny $(4.8)$} & $70.8$\,{\tiny $(4.6)$}
        & $38.3$\,{\tiny $(4.5)$} & $21.9$\,{\tiny $(7.3)$} & $31.2$\,{\tiny $(8.2)$}
        & $54.5$\,{\tiny $(2.3)$} \\
        Random init. & Joint
        & $\mathbf{97.9}$\,{\tiny $(1.5)$} & $50.0$\,{\tiny $(5.1)$} & $79.2$\,{\tiny $(4.1)$}
        & $31.6$\,{\tiny $(3.8)$} & $31.2$\,{\tiny $(8.2)$} & $43.8$\,{\tiny $(8.8)$}
        & $55.6$\,{\tiny $(2.4)$} \\
        \midrule
        \gr
        Pretrained & Joint
        & $95.8$\,{\tiny $(2.0)$} & $56.2$\,{\tiny $(5.1)$} & $\mathbf{81.2}$\,{\tiny $(4.0)$}
        & $\mathbf{40.2}$\,{\tiny $(5.2)$} & $\mathbf{40.6}$\,{\tiny $(8.7)$} & $\mathbf{46.9}$\,{\tiny $(8.8)$}
        & $\mathbf{60.2}$\,{\tiny $(2.5)$} \\
        \bottomrule
    \end{tabular}%
    }
    \par}
    \vspace{-0.2em}
    \caption{
    \textbf{\ours\ Predictor initialization and joint training.}
    We compare pretrained and random initialization, and single-domain and joint training.
    We report success rates (\%), except for the DROID action score, with standard errors in parentheses.
    Average is the unweighted mean of the six benchmark scores.
    }
    \label{tab:predictor-initialization-joint-training}
    \vspace{-1em}
\end{table}

%% file: table/vision-encoder-ablation.tex
\begin{table}[t]
    \centering
    {\footnotesize
    \renewcommand{\arraystretch}{1.3}
    \setlength{\tabcolsep}{2pt}
    \setlength{\aboverulesep}{0.5pt}
    \setlength{\belowrulesep}{0.5pt}
    \setlength{\arrayrulewidth}{0.8pt}
    \setlength{\dashlinedash}{0.6pt}
    \setlength{\dashlinegap}{1.5pt}
    \resizebox{\linewidth}{!}{%
    \begin{tabular}{>{\arrayrulecolor{black!25}}l:*{10}{c}:c}
        \toprule
        Vision encoder & Wall & PointMaze & Push-T
        & \shortstack{MetaW.-R} & \shortstack{MetaW.-RW}
        & Rope & Granular & \shortstack{RoboC.-R} & \shortstack{RoboC.-P}
        & DROID & Average \\
        \midrule
        \gr
        \textbf{JEPA-WMs} & & & & & & & & & & & \\
        \midrule
        DINOv2-S/14
        & $\mathbf{79.2}$ & $\mathbf{60.4}$ & $\mathbf{68.8}$ & $20.8$ & $4.2$
        & $54.2$ & $\mathbf{50.0}$ & $\mathbf{37.5}$ & $\mathbf{31.2}$ & $40.1$ & $44.6$ \\
        DINOv3-L/16
        & $76.0$ & $56.2$ & $47.9$ & $\mathbf{66.7}$ & $\mathbf{47.9}$
        & $\mathbf{62.5}$ & $41.7$ & $21.9$ & $28.1$ & $\mathbf{43.0}$ & $\mathbf{49.2}$ \\
        \hdashline
        Table~\ref{tab:planning-benchmarks} (reference)
        & $79.2$ & $60.4$ & $68.8$ & $66.7$ & $47.9$
        & $62.5$ & $50.0$ & $37.5$ & $31.2$ & $43.0$ & $54.7$ \\
        \midrule
        \gr
        \textbf{\ours} & & & & & & & & & & & \\
        \midrule
        DINOv2-S/14
        & $95.8$ & $45.8$ & $\mathbf{81.2}$ & $8.3$ & $4.2$
        & $58.3$ & $58.3$ & $40.6$ & $43.8$ & $40.2$ & $47.7$ \\
        DINOv3-L/16
        & $94.8$ & $56.2$ & $52.1$ & $75.0$ & $52.1$
        & $\mathbf{66.7}$ & $50.0$ & $12.5$ & $\mathbf{46.9}$ & $30.3$ & $53.7$ \\
        SigLIP2-So400m/14
        & $60.4$ & $37.5$ & $66.7$ & $33.3$ & $8.3$
        & $45.8$ & $41.7$ & $9.4$ & $34.4$ & $35.5$ & $37.3$ \\
        DINOv2 + DINOv3
        & $\mathbf{96.9}$ & $\mathbf{59.4}$ & $80.2$ & $\mathbf{77.1}$ & $\mathbf{54.2}$
        & $\mathbf{66.7}$ & $\mathbf{66.7}$ & $\mathbf{43.8}$ & $43.8$ & $\mathbf{41.5}$ & $\mathbf{63.0}$ \\
        \hdashline
        Table~\ref{tab:planning-benchmarks} (reference)
        & $95.8$ & $56.2$ & $81.2$ & $75.0$ & $52.1$
        & $66.7$ & $58.3$ & $40.6$ & $46.9$ & $40.2$ & $61.3$ \\
        \bottomrule
    \end{tabular}%
    }
    \par}
    \vspace{-0.2em}
    \caption{
    \textbf{Vision encoder ablations.}
    We compare frozen vision encoders for JEPA-WMs and \ours.
    DINOv2 + DINOv3 uses both DINOv2-S/14 and DINOv3-L/16.
    We report success rates (\%), except for the DROID action score.
    Average is the unweighted mean of the ten benchmark scores.
    Bold denotes the highest score among encoder configurations within each method; reference rows reproduce the main planning results and are excluded from this comparison.
    }
    \label{tab:vision-encoder-ablation}
    \vspace{-.5em}
\end{table}

%% file: table/qualitative-language-actions.tex
\begin{table}[t]
    \centering
    {\footnotesize
    \renewcommand{\arraystretch}{1.4}
    \setlength{\aboverulesep}{0.5pt}
    \setlength{\belowrulesep}{0.5pt}
    \setlength{\tabcolsep}{6pt}
    \resizebox{\linewidth}{!}{%
    \begin{tabular}{w{l}{9.3em}*{7}{w{c}{5.7em}}}
        \toprule
        Language action & Wall & PointMaze & Push-T & MetaWorld-R & MetaWorld-RW & RoboCasa-R & RoboCasa-P \\
        \midrule
        Qualitative & $83.3$\,{\tiny $(3.8)$} & $47.9$\,{\tiny $(5.1)$} & $4.2$\,{\tiny $(2.0)$} & $70.8$\,{\tiny $(6.6)$} & $60.4$\,{\tiny $(7.1)$} & $28.1$\,{\tiny $(7.9)$} & $25.0$\,{\tiny $(7.7)$} \\
        Numerical & $95.8$\,{\tiny $(2.0)$} & $56.2$\,{\tiny $(5.1)$} & $81.2$\,{\tiny $(4.0)$} & $75.0$\,{\tiny $(6.2)$} & $52.1$\,{\tiny $(7.2)$} & $40.6$\,{\tiny $(8.7)$} & $46.9$\,{\tiny $(8.8)$} \\
        \bottomrule
    \end{tabular}%
    }
    }
    \vspace{-0.2em}
    \caption{
        \textbf{MPC planning with numerical and qualitative language actions.}
    Qualitative language actions replace numerical magnitudes with \emph{slightly}, no modifier, or \emph{far}.
    Lower success rates are expected when exact magnitudes are omitted, yet it is still interesting that \ours\ can plan successfully in several domains from qualitative language actions.
    Only the language conditions supplied to the predictor change; CEM samples numerical actions, and the simulator executes the selected actions in their original numerical form.
}
\label{tab:qualitative-language-actions}
    \vspace{-.5em}
\end{table}

%% file: table/phrasing-robustness.tex
\begin{table}[t]
    \centering
    \begin{minipage}{\linewidth}
        \vspace{0pt}
        \centering
        {\footnotesize
        \renewcommand{\arraystretch}{1.2}
        \setlength{\tabcolsep}{6pt}
        \resizebox{\linewidth}{!}{%
        \begin{tabular}{w{l}{5.7em}*{4}{w{c}{11.6em}}}
            \toprule
            Phrasing & Wall & PointMaze & MetaWorld-R & RoboCasa-R / DROID \\
            \midrule
            Original & move the agent & push the ball with force & move the gripper
            & move gripper, rotate roll \\
            \addlinespace[2pt]
            Rephrased & push the dot & push the sphere & shift the robot hand
            & move robot hand, turn roll \\
            \bottomrule
        \end{tabular}%
        }
        }
    \end{minipage}
    \par\vspace{0.7em}
    \begin{minipage}{\linewidth}
        \vspace{0pt}
        \centering
        {\footnotesize
        \renewcommand{\arraystretch}{1.2}
        \setlength{\tabcolsep}{6pt}
        \setlength{\arrayrulewidth}{0.8pt}
        \setlength{\dashlinedash}{0.6pt}
        \setlength{\dashlinegap}{1.5pt}
        \resizebox{\linewidth}{!}{%
        \begin{tabular}{>{\arrayrulecolor{black!25}}w{l}{9.3em}:*{5}{w{c}{7.1em}}:w{c}{4.5em}}
            \toprule
            Method & Wall & PointMaze & MetaWorld-R & RoboCasa-R & DROID & Average \\
            \midrule
            Control data only
            & $96.9$\,{\tiny $(1.8)$}
            & $64.6$\,{\tiny $(4.9)$}
            & $68.8$\,{\tiny $(6.7)$}
            & $25.0$\,{\tiny $(7.7)$}
            & $41.5$\,{\tiny $(4.6)$}
            & $59.3$\,{\tiny $(2.5)$} \\
            + Paraphrasing
            & $87.5$\,{\tiny $(3.4)$}
            & $56.2$\,{\tiny $(5.1)$}
            & $60.4$\,{\tiny $(7.1)$}
            & $18.8$\,{\tiny $(6.9)$}
            & $39.8$\,{\tiny $(5.0)$}
            & $52.5$\,{\tiny $(2.5)$} \\
            \midrule
            All data
            & $95.8$\,{\tiny $(2.0)$}
            & $56.2$\,{\tiny $(5.1)$}
            & $75.0$\,{\tiny $(6.2)$}
            & $40.6$\,{\tiny $(8.7)$}
            & $40.2$\,{\tiny $(5.2)$}
            & $61.6$\,{\tiny $(2.6)$} \\
            + Paraphrasing
            & $92.7$\,{\tiny $(2.7)$}
            & $57.3$\,{\tiny $(5.0)$}
            & $72.9$\,{\tiny $(6.4)$}
            & $43.8$\,{\tiny $(8.8)$}
            & $38.3$\,{\tiny $(5.3)$}
            & $61.0$\,{\tiny $(2.7)$} \\
            \bottomrule
        \end{tabular}%
        }
        }
    \end{minipage}
    \vspace{-0.5em}
    \caption{\textbf{Robustness to action phrasing.}
    We report planning results with original and paraphrased action templates (bottom) and the substitutions used at evaluation (top).
    All data combines control trajectories with image and video transitions.
    We report success rates (\%), except for the DROID action score, with standard errors in parentheses.}
    \label{tab:phrasing-robustness}
    \vspace{-.5em}
\end{table}

%% file: figure/latent-vs-generative-per-domain.tex
\begin{figure}[t]
    \centering
    \includegraphics[width=\linewidth]{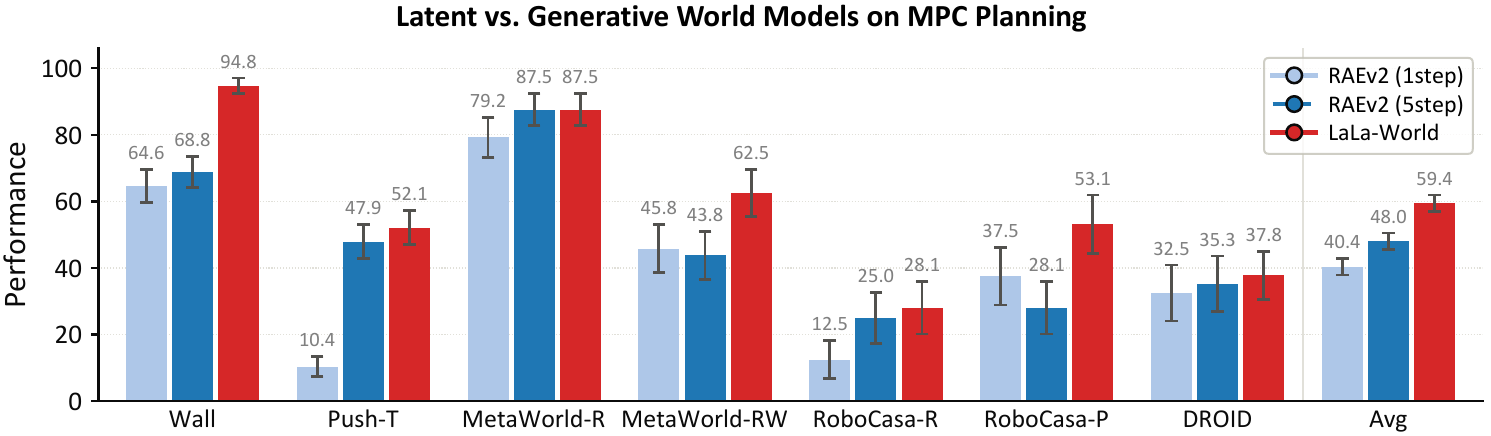}
    \vspace{-1.5em}
    \caption{
    \textbf{Domain-level comparison of latent and generative world models for MPC planning.}
    We compare \ours\ with RAEv2 using one and five denoising steps under the matched setup in \cref{sec:latent-vs-pixel}.
    We report success rates (\%) for simulation tasks and the action score for DROID.
    }
    \vspace{-.5em}
    \label{fig:latent-vs-generative-per-domain}
\end{figure}

%% file: table/world-model-training-data.tex
\begin{table}[t]
    \centering
    \begingroup
    \small
    \setlength{\tabcolsep}{12pt}
    \renewcommand{\arraystretch}{1.1}
    \begin{tabular}{@{}lr@{}}
        \toprule
        Dataset & Training data size \\
        \midrule
        Wall                &    86,296 \\
        PointMaze           &   186,675 \\
        Push-T               & 2,199,488 \\
        MetaWorld          & 1,162,310 \\
        Rope                &    18,525 \\
        Granular            &    18,582 \\
        DROID               & 1,500,000 \\
        \midrule
        OpenVid-1M          & 1,023,363 \\
        OmniEdit            &   295,353 \\
        ShareGPT-4o-Image    &    45,613 \\
        \bottomrule
    \end{tabular}
    \endgroup
    \caption{\textbf{Data used to train \ours.}
    We count target transitions for control and video data and editing pairs for OmniEdit and ShareGPT-4o-Image.
    Counts exclude validation data and count each control transition once across its two action description formats.
    }
    \label{tab:world-model-training-data}
\end{table}

%% file: table/planning-setup.tex
\begin{table}[t]
    \centering
    \begingroup
    \footnotesize
    \setlength{\tabcolsep}{5pt}
    \renewcommand{\arraystretch}{1.2}
    \setlength{\aboverulesep}{0.5pt}
    \setlength{\belowrulesep}{0.5pt}
    \setlength{\dashlinedash}{0.6pt}
    \setlength{\dashlinegap}{1.5pt}
    \newcommand{\mlc}[2][c]{\begin{tabular}[c]{@{}#1@{}}#2\end{tabular}}

    \resizebox{\linewidth}{!}{%
        \begin{tabular}{@{}>{\arrayrulecolor{black!25}}l:cccccc:l@{}}
            \toprule
            Benchmark
            & \mlc{Evaluation\\trajectories}
            & \mlc{Trajectory\\length}
            & \mlc{Planning\\horizon $H$}
            & \mlc{Executed steps\\per plan}
            & \mlc{Substeps\\per step}
            & \mlc{CEM\\settings}
            & \mlc[l]{Evaluation\\criterion} \\
            \midrule
            Wall
            & 96 & 6 & 6 & 6 & 5
            & \(30 \times 300 \times 10\)
            & Position error \(< 4.5\) \\
            PointMaze
            & 96 & 6 & 6 & 6 & 5
            & \(30 \times 300 \times 10\)
            & Position error \(< 0.5\) \\
            Push-T
            & 96 & 6 & 6 & 6 & 5
            & \(30 \times 300 \times 10\)
            & \mlc[l]{Position error \(< 20\)\\and angle error \(< 20^{\circ}\)} \\
            MetaWorld-R
            & 48 & 20 & 6 & 3 & 5
            & \(15 \times 300 \times 10\)
            & Success at any step \\
            MetaWorld-RW
            & 48 & 20 & 6 & 3 & 5
            & \(15 \times 300 \times 10\)
            & Success at any step \\
            Rope
            & 24 & 5 & 5 & 1 & 1
            & \(30 \times 300 \times 30\)
            & Chamfer distance \(< 0.675\) \\
            Granular
            & 12 & 5 & 5 & 1 & 1
            & \(30 \times 300 \times 30\)
            & Chamfer distance \(< 0.323\) \\
            DROID
            & 64 & 3 & 3 & -- & 1
            & \(15 \times 300 \times 10\)
            & Action Score \\
            RoboCasa-R
            & 32 & 12 & 3 & 1 & 5
            & \(15 \times 300 \times 10\)
            & Subtask success at final step \\
            RoboCasa-P
            & 32 & 12 & 3 & 1 & 5
            & \(15 \times 300 \times 10\)
            & Subtask success at final step \\
            \bottomrule
        \end{tabular}%
    }

    \endgroup
    \vspace{-.5em}
    \caption{
    \textbf{MPC planning settings and evaluation criteria.}
    One step corresponds to one world-model transition,
    with the listed number of action updates (substeps).
    CEM settings are iterations \(\times\) candidate sequences
    per iteration \(\times\) retained sequences.
    For benchmarks requiring replanning,
    we execute the listed number of steps, observe the environment,
    and re-optimize the action sequence with CEM.
    Position and angle errors are measured at the final step,
    using \(L_2\) distance for position.
    DROID is evaluated offline without simulator execution.
    }
    \vspace{-1.em}
    \label{tab:planning-setup}
\end{table}

%% file: figure/planning-setup-vis.tex
\begin{figure}[t]
    \centering
    \begingroup
    \definecolor{planblue}{HTML}{0F4281}
    \definecolor{plansky}{HTML}{E4ECF6}
    \definecolor{planink}{HTML}{202020}
    \definecolor{planmuted}{HTML}{6A6A6A}
    \definecolor{planline}{HTML}{BFBFBF}
    \resizebox{\linewidth}{!}{%
    \begin{tikzpicture}[
        x=1cm, y=1cm,
        font=\sffamily\fontsize{8}{10}\selectfont, text=planink,
        predicted/.style={draw=planblue!35!planline, fill=plansky, line width=0.65pt},
        executed/.style={draw=planblue, fill=planblue, line width=0.65pt},
        replan/.style={-{Stealth[length=1.5mm]}, draw=planmuted, line width=0.7pt},
        horizon/.style={decorate, decoration={brace, amplitude=3pt}, draw=planmuted},
        heading/.style={font=\sffamily\bfseries\fontsize{8}{9.6}\selectfont},
        detail/.style={text=planink, font=\sffamily\fontsize{8}{9.5}\selectfont}
    ]
        \path[use as bounding box] (0,-0.15) rectangle (15.3,4.59);

        \path[executed] (3.15,4.22) rectangle +(0.25,0.22);
        \node[anchor=west] at (3.5,4.33) {Executed};
        \path[predicted] (5.65,4.22) rectangle +(0.25,0.22);
        \node[anchor=west] at (6,4.33) {Predicted only};
        \node[anchor=west, detail] at (9.0,4.33) {Each box = one step};

        \draw[planline] (5.1,0.05) -- (5.1,4.05);
        \draw[planline] (10.2,0.05) -- (10.2,4.05);

        \begin{scope}
            \node[heading] at (2.55,3.95) {(a) Wall};
            \node[detail] at (2.55,3.58) {$H=6$; execute 6 steps per plan};
            \draw[horizon] (0.5,3.22) -- (3.03,3.22);
            \foreach \i in {1,...,6} {
                \pgfmathsetmacro{\px}{0.5+(\i-1)*0.43}
                \path[executed] (\px,2.68) rectangle +(0.38,0.4);
                \node[text=white] at (\px+0.19,2.88) {\i};
            }
            \draw[replan] (1.765,2.58) -- (1.765,2.18);
            \node[anchor=north,align=center] at (1.765,2.05)
                {Trajectory complete\\[2pt]No replanning};
            \node at (2.55,0.78) {Trajectory length: 6 steps};
            \draw[planline] (0.5,0.48) -- (4.6,0.48);
            \node[detail] at (2.55,0.15) {1 step = 5 simulator substeps};
        \end{scope}

        \begin{scope}[xshift=5.1cm]
            \node[heading] at (2.55,3.95) {(b) MetaWorld Reach};
            \node[detail] at (2.55,3.58) {$H=6$; execute 3 steps per plan};
            \draw[horizon] (0.5,3.22) -- (3.03,3.22);
            \foreach \i in {1,...,6} {
                \pgfmathsetmacro{\px}{0.5+(\i-1)*0.43}
                \ifnum\i<4
                    \path[executed] (\px,2.68) rectangle +(0.38,0.4);
                    \node[text=white] at (\px+0.19,2.88) {\i};
                \else
                    \path[predicted] (\px,2.68) rectangle +(0.38,0.4);
                    \node[text=planink] at (\px+0.19,2.88) {\i};
                \fi
            }
            \draw[replan] (1.575,2.58) -- (1.575,2.35) -| (1.98,2.03);
            \node[anchor=west,detail] at (2.15,2.35) {Observe; replan};
            \foreach \i in {4,...,9} {
                \pgfmathsetmacro{\px}{0.5+(\i-1)*0.43}
                \ifnum\i<7
                    \path[executed] (\px,1.53) rectangle +(0.38,0.4);
                    \node[text=white] at (\px+0.19,1.73) {\i};
                \else
                    \path[predicted] (\px,1.53) rectangle +(0.38,0.4);
                    \node[text=planink] at (\px+0.19,1.73) {\i};
                \fi
            }
            \node[detail] at (2.55,1.25) {$\cdots$ continue replanning};
            \node at (2.55,0.78) {Trajectory length: 20 steps};
            \draw[planline] (0.5,0.48) -- (4.6,0.48);
            \node[detail] at (2.55,0.15) {1 step = 5 simulator substeps};
        \end{scope}

        \begin{scope}[xshift=10.2cm]
            \node[heading] at (2.55,3.95) {(c) Rope / Granular};
            \node[detail] at (2.55,3.58) {$H=5$; execute 1 step per plan};
            \draw[horizon] (0.5,3.22) -- (2.6,3.22);
            \foreach \i in {1,...,5} {
                \pgfmathsetmacro{\px}{0.5+(\i-1)*0.43}
                \ifnum\i=1
                    \path[executed] (\px,2.68) rectangle +(0.38,0.4);
                    \node[text=white] at (\px+0.19,2.88) {\i};
                \else
                    \path[predicted] (\px,2.68) rectangle +(0.38,0.4);
                    \node[text=planink] at (\px+0.19,2.88) {\i};
                \fi
            }
            \draw[replan] (0.69,2.58) -- (0.69,2.35) -| (1.12,2.03);
            \node[anchor=west,detail] at (1.3,2.35) {Observe; replan};
            \foreach \i in {2,...,6} {
                \pgfmathsetmacro{\px}{0.5+(\i-1)*0.43}
                \ifnum\i=2
                    \path[executed] (\px,1.53) rectangle +(0.38,0.4);
                    \node[text=white] at (\px+0.19,1.73) {\i};
                \else
                    \path[predicted] (\px,1.53) rectangle +(0.38,0.4);
                    \node[text=planink] at (\px+0.19,1.73) {\i};
                \fi
            }
            \node[detail] at (2.55,1.25) {$\cdots$ replan after each push};
            \node at (2.55,0.78) {Trajectory length: 5 steps};
            \draw[planline] (0.5,0.48) -- (4.6,0.48);
            \node[detail] at (2.55,0.15) {1 step = 1 push};
        \end{scope}
    \end{tikzpicture}%
    }
    \endgroup
    \vspace{-1.8em}
    \caption{
    \textbf{MPC planning horizons and action execution.}
    One step corresponds to one world-model transition,
    with the listed number of action updates (substeps).
    For benchmarks requiring replanning, we obtain a new observation
    after executing the prefix and optimize a new sequence with CEM.
    }
    \vspace{-.5em}
    \label{fig:planning-step-vis}
\end{figure}
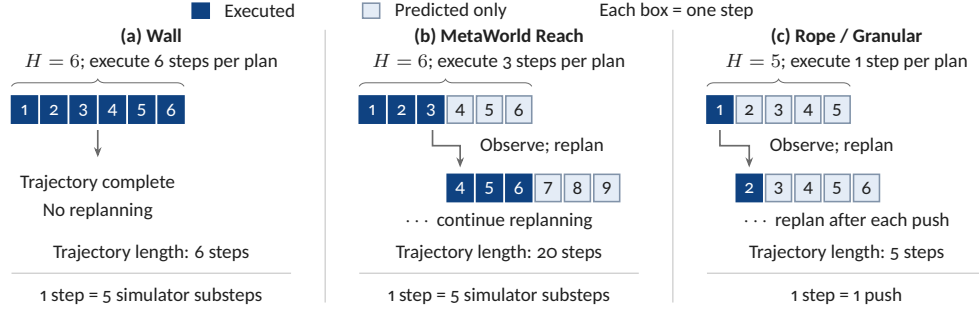

%% file: figure/substep-aggregation.tex
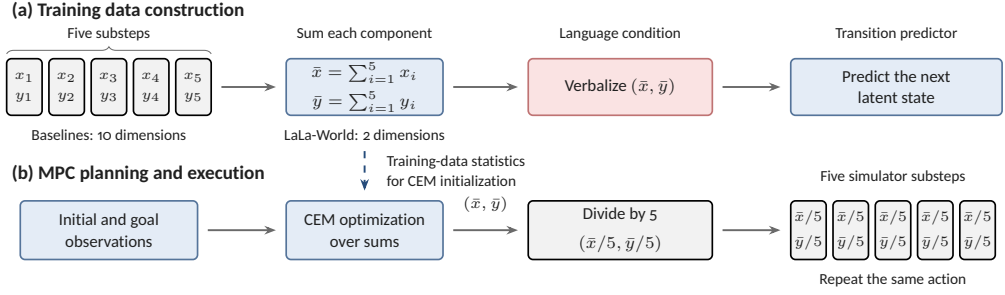
\begin{figure}[t]
    \centering
    \begingroup
    \definecolor{aggregateblue}{HTML}{0F4281}
    \definecolor{aggregatesky}{HTML}{E4ECF6}
    \definecolor{aggregatered}{HTML}{C00000}
    \definecolor{aggregatepink}{HTML}{FAE3E3}
    \definecolor{aggregateink}{HTML}{202020}
    \definecolor{aggregatemuted}{HTML}{6A6A6A}
    \definecolor{aggregateline}{HTML}{BFBFBF}
    \definecolor{aggregategray}{HTML}{F2F2F2}
    \resizebox{\linewidth}{!}{%
    \begin{tikzpicture}[
        x=1cm, y=1cm,
        font=\sffamily\fontsize{8}{10}\selectfont, text=aggregateink,
        heading/.style={font=\sffamily\bfseries\fontsize{8}{9.6}\selectfont},
        boxtext/.style={font=\sffamily\fontsize{7}{9}\selectfont},
        outside/.style={font=\sffamily\fontsize{6.4}{8}\selectfont, text=aggregateink},
        detail/.style={outside},
        data/.style={draw=black, fill=aggregategray, rounded corners=2pt, line width=0.7pt},
        model/.style={data},
        state/.style={data, draw=aggregateblue!60!aggregateline, fill=aggregatesky},
        language/.style={data, draw=aggregatered!40!aggregateline, fill=aggregatepink},
        flow/.style={-{Stealth[length=1.7mm]}, draw=aggregatemuted, line width=0.8pt},
        stats/.style={flow, dashed, draw=aggregateblue!75!aggregatemuted},
        brace/.style={decorate, decoration={brace, amplitude=3pt}, draw=aggregatemuted}
    ]
        \path[use as bounding box] (0,0.7) rectangle (15.3,5.05);

        \begin{scope}[yshift=-0.825cm]
            \node[heading,anchor=west] at (0.1,5.65) {(a) Training data construction};
            \node[outside] at (1.65,5.31) {Five substeps};
            \node[outside] at (5.4,5.31) {Sum each component};
            \node[outside] at (9.15,5.31) {Language condition};
            \node[outside] at (13.2,5.31) {Transition predictor};

            \foreach \i in {1,...,5} {
                \pgfmathsetmacro{\px}{0.15+(\i-1)*0.62}
                \path[data] (\px,4.12) rectangle +(0.52,0.83);
                \node at (\px+0.26,4.535) {$\begin{smallmatrix}x_{\i}\\[3pt]y_{\i}\end{smallmatrix}$};
            }
            \draw[brace] (0.15,5.02) -- (3.15,5.02);
            \node[detail] at (1.65,3.83) {Baselines: 10 dimensions};

            \path[state] (4.25,4.12) rectangle (6.55,4.95);
            \node[align=center,boxtext] at (5.4,4.535)
                {$\bar{x}=\sum_{i=1}^{5}x_i$\\[3pt]$\bar{y}=\sum_{i=1}^{5}y_i$};
            \node[detail] at (5.4,3.83) {\ours: 2 dimensions};

            \path[language] (7.8,4.12) rectangle (10.5,4.95);
            \node[align=center,boxtext] at (9.15,4.535) {Verbalize $(\bar{x},\bar{y})$};
            \path[state] (11.65,4.12) rectangle (14.75,4.95);
            \node[align=center,boxtext] at (13.2,4.535) {Predict the next\\latent state};
            \draw[flow] (3.3,4.535) -- (4.1,4.535);
            \draw[flow] (6.7,4.535) -- (7.65,4.535);
            \draw[flow] (10.65,4.535) -- (11.5,4.535);
        \end{scope}

        \draw[stats] (5.4,2.77) -- (5.4,2.19);
        \node[outside,anchor=west,align=left] at (5.6,2.48)
            {Training-data statistics\\for CEM initialization};

        \node[heading,anchor=west] at (0.1,2.44) {(b) MPC planning and execution};
        \path[state] (0.35,1.18) rectangle (2.95,2.04);
        \node[align=center,boxtext] at (1.65,1.61) {Initial and goal\\observations};
        \path[state] (4.25,1.18) rectangle (6.55,2.04);
        \node[align=center,boxtext] at (5.4,1.61) {CEM optimization\\over sums};

        \path[model] (7.8,1.18) rectangle (10.5,2.04);
        \node[align=center,boxtext] at (9.15,1.61)
            {Divide by 5\\[3pt]$(\bar{x}/5,\bar{y}/5)$};
        \draw[flow] (3.1,1.61) -- (4.1,1.61);
        \draw[flow] (6.7,1.61) -- (7.65,1.61)
            node[outside,midway,above=4pt] {$(\bar{x},\bar{y})$};
        \draw[flow] (10.65,1.61) -- (11.5,1.61);

        \foreach \i in {1,...,5} {
            \pgfmathsetmacro{\px}{11.65+(\i-1)*0.62}
            \path[model] (\px,1.18) rectangle +(0.52,0.86);
            \node at (\px+0.26,1.61)
                {$\begin{smallmatrix}\bar{x}/5\\[3pt]\bar{y}/5\end{smallmatrix}$};
        }
        \node[outside] at (13.15,2.39) {Five simulator substeps};
        \node[detail] at (13.15,0.88) {Repeat the same action};
    \end{tikzpicture}%
    }
    \endgroup
    \vspace{-2em}
    \caption{
    \textbf{Substep aggregation for language conditioning and action execution.}
    (a) Summing across substeps reduces the action dimension per step
    from 10 to 2 for language conditioning.
    (b) Within MPC, CEM optimizes these sums, which are distributed equally
    across five substeps for execution.
    This discards within-step variation.
    Yet, it is notable that \ours\ achieves accurate prediction and high planning success rates.
    }
    \vspace{-1em}
    \label{fig:substep-aggregation}
\end{figure}

%% file: figure/qualitative-language-actions.tex
\begin{figure}[t]
    \centering
    \begingroup
    \definecolor{qualblue}{HTML}{0F4281}
    \definecolor{qualsky}{HTML}{E4ECF6}
    \definecolor{qualred}{HTML}{C00000}
    \definecolor{qualpink}{HTML}{FAE3E3}
    \definecolor{qualink}{HTML}{202020}
    \definecolor{qualmuted}{HTML}{6A6A6A}
    \definecolor{qualline}{HTML}{BFBFBF}
    \definecolor{qualgray}{HTML}{F2F2F2}
    \definecolor{qualbinlow}{HTML}{EDEDED}
    \definecolor{qualbinmid}{HTML}{D0D0D0}
    \definecolor{qualbinhigh}{HTML}{B3B3B3}
    \colorlet{quallanguageline}{qualred!40!qualline}
    \newcommand{\figprompt}[1]{{\ttfamily #1}}
    \resizebox{\linewidth}{!}{%
    \begin{tikzpicture}[
        x=1cm, y=1cm,
        font=\sffamily\fontsize{8}{10}\selectfont, text=qualink,
        heading/.style={font=\sffamily\bfseries\fontsize{8}{9.6}\selectfont},
        boxtext/.style={font=\sffamily\fontsize{7}{9}\selectfont},
        outside/.style={font=\sffamily\fontsize{6.4}{8}\selectfont, text=qualink},
        detail/.style={outside},
        data/.style={draw=black, fill=qualgray, rounded corners=2pt, line width=0.7pt},
        model/.style={data},
        state/.style={data, draw=qualblue!60!qualline, fill=qualsky},
        language/.style={data, draw=quallanguageline, fill=qualpink},
        flow/.style={-{Stealth[length=1.7mm]}, draw=qualmuted, line width=0.8pt},
        stats/.style={flow, dashed, draw=qualmuted}
    ]
        \path[use as bounding box] (0,-0.05) rectangle (15.3,6.4);

        \begin{scope}[text=black, outside/.append style={text=black},
            flow/.append style={draw=black}]
        \node[heading,anchor=west] at (0.1,6.18) {(a) Action conversion};
        \node[outside,anchor=west,align=left] at (0.1,5.53) {Numerical\\language action};
        \node[outside,anchor=west,align=left] at (0.1,4.7) {Leftward magnitude\\in training data};
        \node[outside,anchor=west,align=left] at (0.1,3.8) {Qualitative\\language action};

        \path[draw=black, fill=qualbinlow,  line width=0.7pt] (3,4.45) rectangle (6.333,4.95);
        \path[draw=black, fill=qualbinmid,  line width=0.7pt] (6.333,4.45) rectangle (9.667,4.95);
        \path[draw=black, fill=qualbinhigh, line width=0.7pt] (9.667,4.45) rectangle (13,4.95);
        \node[boxtext] at (4.667,4.7) {\emph{slightly}};
        \node[boxtext] at (8,4.7) {no modifier};
        \node[boxtext] at (11.333,4.7) {\emph{far}};
        \foreach \tx/\tl in {3/0, 6.333/2, 9.667/4, 13/6} {
            \node[outside] at (\tx,4.27) {\tl};
        }
        \node[outside,anchor=west,align=left] at (13.25,4.7)
            {6: maximum\\observed\\magnitude};

        \foreach \vx/\vl in {3.667/0.4, 8/3.0, 10.5/4.5} {
            \path[draw=black, fill=white, rounded corners=2pt, line width=0.7pt]
                (\vx-1.05,5.28) rectangle (\vx+1.05,5.78);
            \node[boxtext] at (\vx,5.53) {\figprompt{move \vl\ left}};
            \draw[flow] (\vx,5.28) -- (\vx,4.98);
        }

        \foreach \bx/\bc/\bl in {4.667/qualbinlow/move slightly left,
                                 8/qualbinmid/move left,
                                 11.333/qualbinhigh/move far left} {
            \path[draw=black, fill=\bc, rounded corners=2pt, line width=0.7pt]
                (\bx-1.4,3.55) rectangle (\bx+1.4,4.05);
            \node[boxtext] at (\bx,3.8) {\figprompt{\bl}};
            \draw[flow] (\bx,4.45) -- (\bx,4.08);
        }
        \end{scope}

        \node[heading,anchor=west] at (0.1,3.08) {(b) MPC planning procedure};

        \path[model] (0.15,1.5) rectangle (2.65,2.3);
        \node[align=center,boxtext] at (1.4,1.9) {CEM samples\\numerical actions};
        \path[language] (4.55,1.5) rectangle (6.85,2.3);
        \node[align=center,boxtext] at (5.7,1.9) {Conversion rule\\from (a)};
        \path[state] (8.75,1.5) rectangle (11.05,2.3);
        \node[align=center,boxtext] at (9.9,1.9) {\ours\\latent rollout};
        \path[model] (12.95,1.5) rectangle (15.15,2.3);
        \node[align=center,boxtext] at (14.05,1.9) {Goal-distance\\cost};

        \draw[flow] (2.8,1.9) -- (4.4,1.9)
            node[outside,midway,above=2pt] {\figprompt{move 4.5 left}};
        \draw[flow] (7.0,1.9) -- (8.6,1.9)
            node[outside,midway,above=2pt] {\figprompt{move far left}};
        \draw[flow] (11.2,1.9) -- (12.8,1.9)
            node[outside,midway,above=2pt] {Predicted latents};

        \draw[stats] (14.05,2.33) -- (14.05,2.68) -- (1.4,2.68) -- (1.4,2.33);
        \node[outside,fill=white,inner sep=2.5pt] at (7.8,2.68)
            {Costs update the sampling distribution};

        \path[state] (0.15,0.05) rectangle (2.65,0.85);
        \node[align=center,boxtext] at (1.4,0.45) {Current and goal\\observations};
        \path[model] (12.95,0.05) rectangle (15.15,0.85);
        \node[align=center,boxtext] at (14.05,0.45) {Simulator\\execution};

        \draw[flow] (0.8,0.88) -- (0.8,1.47);
        \draw[flow] (2.0,1.47) -- (2.0,1.18) -- (14.05,1.18) -- (14.05,0.88);
        \node[outside,fill=white,inner sep=2.5pt] at (7.8,1.18)
            {Selected actions keep their numerical values: \figprompt{move 4.5 left}};
        \draw[flow] (12.8,0.45) -- (2.8,0.45);
        \node[outside,fill=white,inner sep=2.5pt] at (7.8,0.45)
            {Next observation; replan};
    \end{tikzpicture}%
    }
    \endgroup
    \vspace{-1.2em}
    \caption{
    \textbf{MPC planning with qualitative language actions.}
    (a) Each magnitude maps to one of three equal-width bins, and the direction is kept.
    (b) \ours\ evaluates CEM samples through their qualitative descriptions,
    while the simulator executes the selected actions with their numerical values.
    Thus, \ours\ plans without exact action magnitudes.
    }
    \vspace{-0.5em}
    \label{fig:qualitative-language-actions}
\end{figure}
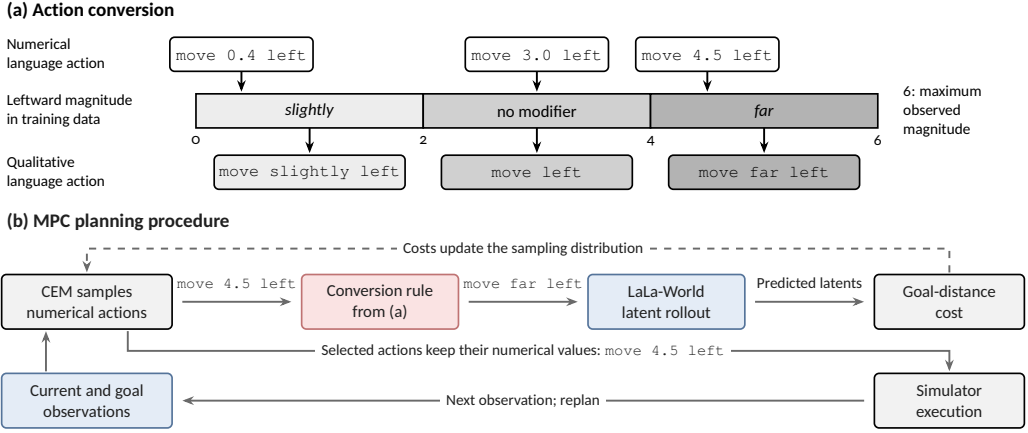

%% file: figure/policy-architecture.tex
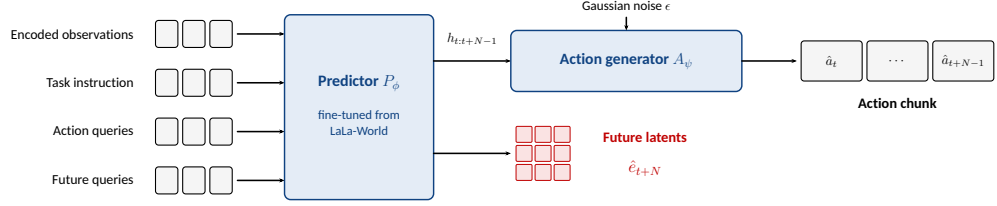
\begin{figure}[t]
\centering
\begingroup
\definecolor{pablue}{HTML}{0F4281}
\definecolor{pabluefill}{HTML}{E4ECF6}
\definecolor{pared}{HTML}{C00000}
\definecolor{paredfill}{HTML}{FAE3E3}
\definecolor{patokenfill}{HTML}{F5F5F5}

\pgfmathsetmacro{\paPredictorY}{.27}
\pgfmathsetmacro{\paPredictorH}{2.12}
\pgfmathsetmacro{\paActionY}{\paPredictorY + .75*\paPredictorH}
\pgfmathsetmacro{\paFutureY}{\paPredictorY + .25*\paPredictorH}
\pgfmathsetmacro{\paReadoutLength}{.77*1.15}
\pgfmathsetmacro{\paHeadX}{5.29 + \paReadoutLength}
\pgfmathsetmacro{\paActionCenterX}{\paHeadX + 1.325}
\pgfmathsetmacro{\paWidth}{11.85 + \paReadoutLength - .77}

\resizebox{\linewidth}{!}{%
\begin{tikzpicture}[
  x=1in,y=1in,
  every node/.style={inner sep=0pt,outer sep=0pt,align=center,text=black,
    font={\sffamily\fontsize{12}{15.6}\selectfont}},
  patoken/.style={draw=black,fill=patokenfill,line width=.9pt,
    rounded corners=2.52pt},
  pablock/.style={draw=pablue,fill=pabluefill,line width=1.2pt,
    rounded corners=5.04pt},
  pafuture/.style={draw=pared,fill=paredfill,line width=.7pt,
    rounded corners=1.08pt},
  paarrow/.style={draw=black,line width=1.3pt,line cap=round,line join=round,
    -{Stealth[length=4.4pt,width=3.3pt]}}
]
\path[use as bounding box] (0,0) rectangle (\paWidth,2.64);

\pgfmathsetmacro{\paTokenW}{(.88-2*.055)/3}
\foreach \paY/\paLabel in {
  2.17/{Encoded observations},
  1.61/{Task instruction},
  1.05/{Action queries},
  .49/{Future queries}} {
  \node[anchor=east,font={\sffamily\fontsize{11.4}{14.82}\selectfont}]
    at (1.85,\paY) {\paLabel};
  \foreach \i in {0,1,2} {
    \pgfmathsetmacro{\paTokenX}{2.10+\i*(\paTokenW+.055)}
    \path[patoken] (\paTokenX,{\paY-.155}) rectangle ++(\paTokenW,.31);
  }
  \draw[paarrow] (3.04,\paY) -- (3.58,\paY);
}

\path[pablock] (3.58,\paPredictorY) rectangle ++(1.71,\paPredictorH);
\node[text=pablue,font={\sffamily\bfseries\fontsize{13.2}{17.16}\selectfont}]
  at (4.435,1.61) {Predictor $P_\phi$};
\node[text=pablue,font={\sffamily\fontsize{10.5}{13.65}\selectfont}]
  at (4.435,1.15) {fine-tuned from\\LaLa-World};

\draw[paarrow] (5.29,\paActionY) -- (\paHeadX,\paActionY);
\node[font={\sffamily\fontsize{11}{14.3}\selectfont}]
  at ({5.29+\paReadoutLength/2},{\paActionY+.25}) {$h_{t:t+N-1}$};
\draw[paarrow] (5.29,\paFutureY) -- (\paHeadX,\paFutureY);

\path[pablock] (\paHeadX,{\paActionY-.35}) rectangle ++(2.65,.70);
\node[text=pablue,font={\sffamily\bfseries\fontsize{13}{16.9}\selectfont}]
  at (\paActionCenterX,\paActionY) {Action generator $A_\psi$};
\node[font={\sffamily\fontsize{11}{14.3}\selectfont}]
  at (\paActionCenterX,2.50) {Gaussian noise $\epsilon$};
\draw[paarrow] (\paActionCenterX,2.36) -- (\paActionCenterX,{\paActionY+.35});
\draw[paarrow] ({\paHeadX+2.65},\paActionY) -- ({\paHeadX+3.28},\paActionY);

\pgfmathsetmacro{\paActionTokenW}{(2.21-2*.055)/3}
\foreach \i/\paActionLabel in {
  0/{\hat a_t},1/{\cdots},2/{\hat a_{t+N-1}}} {
  \pgfmathsetmacro{\paActionTokenX}{\paHeadX+3.33+\i*(\paActionTokenW+.055)}
  \path[patoken] (\paActionTokenX,{\paActionY-.23}) rectangle ++(\paActionTokenW,.46);
  \node[font={\sffamily\fontsize{11}{14.3}\selectfont}]
    at ({\paActionTokenX+\paActionTokenW/2},\paActionY) {$\paActionLabel$};
}
\node[font={\sffamily\bfseries\fontsize{12}{15.6}\selectfont}]
  at ({\paHeadX+4.435},{\paActionY-.48}) {Action chunk};

\pgfmathsetmacro{\paPatchW}{(.62-2*.035)/3}
\foreach \i in {0,1,2} {
  \foreach \j in {0,1,2} {
    \path[pafuture]
      ({\paHeadX+.06+\i*(\paPatchW+.035)},{\paFutureY-.31+\j*(\paPatchW+.035)})
      rectangle ++(\paPatchW,\paPatchW);
  }
}
\node[text=pared,font={\sffamily\bfseries\fontsize{12}{15.6}\selectfont}]
  at ({\paHeadX+1.54},{\paFutureY+.185}) {Future latents};
\node[text=pared,font={\sffamily\fontsize{14}{18.2}\selectfont}]
  at ({\paHeadX+1.54},{\paFutureY-.17}) {$\hat e_{t+N}$};
\end{tikzpicture}%
}
\endgroup
\vspace{-1.3em}
\caption{
    \textbf{\ourpolicy\ architecture.}
The predictor, fine-tuned from \ours, receives encoded observations, a task
instruction, and learned action and future queries. It predicts future visual
latents and provides conditioning features for an action generator, which
produces an action chunk from Gaussian noise. Proprio and intermediate
projections are omitted for clarity.}
\label{fig:policy-architecture}
\vspace{-1em}
\end{figure}

%% file: figure-qualitative/world-model/all.tex
\input{figure-qualitative/world-model/num-wall}
\input{figure-qualitative/world-model/num-point}
\input{figure-qualitative/world-model/num-pusht}
\input{figure-qualitative/world-model/num-meta}
\input{figure-qualitative/world-model/num-droid}
\input{figure-qualitative/world-model/num-robo}
\input{figure-qualitative/world-model/num-robo-p}

\clearpage
\input{figure-qualitative/world-model/qual-wall}
\input{figure-qualitative/world-model/qual-point}
\input{figure-qualitative/world-model/qual-pusht}
\input{figure-qualitative/world-model/qual-meta}
\input{figure-qualitative/world-model/qual-robo}
\input{figure-qualitative/world-model/qual-robo-p}

%% file: figure-qualitative/world-model/num-wall.tex
\begin{figure}[p]
    \centering
    \includegraphics[width=\linewidth]{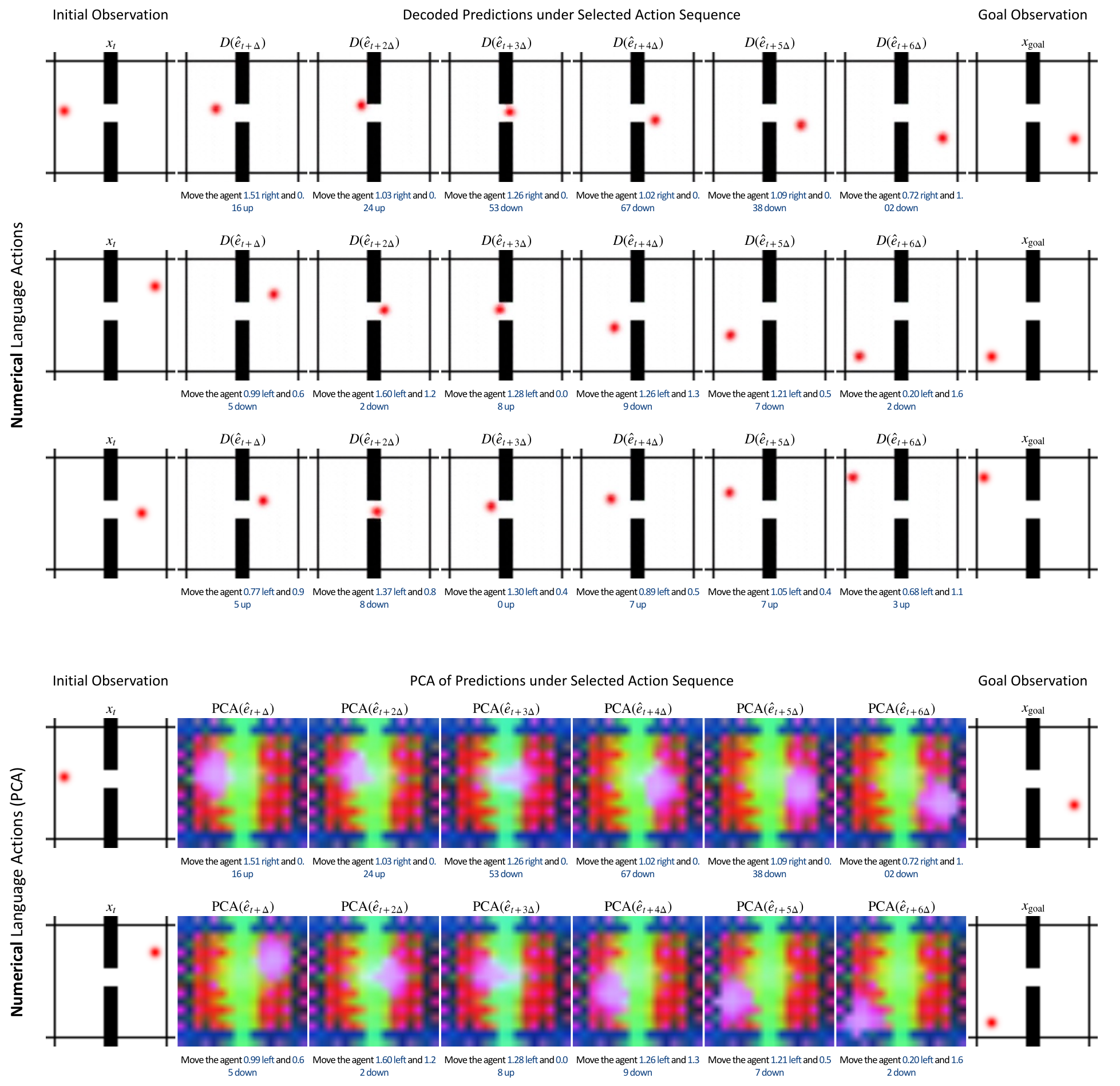}
    \caption{
    \textbf{Numerical language actions on Wall.}
    Initial and goal observations bracket decoded future predictions (top) and PCA visualizations of predicted latents (bottom), with the selected language actions below each frame.
    }
    \label{fig:wm-num-wall}
\end{figure}

%% file: figure-qualitative/world-model/num-point.tex
\begin{figure}[p]
    \centering
    \includegraphics[width=\linewidth]{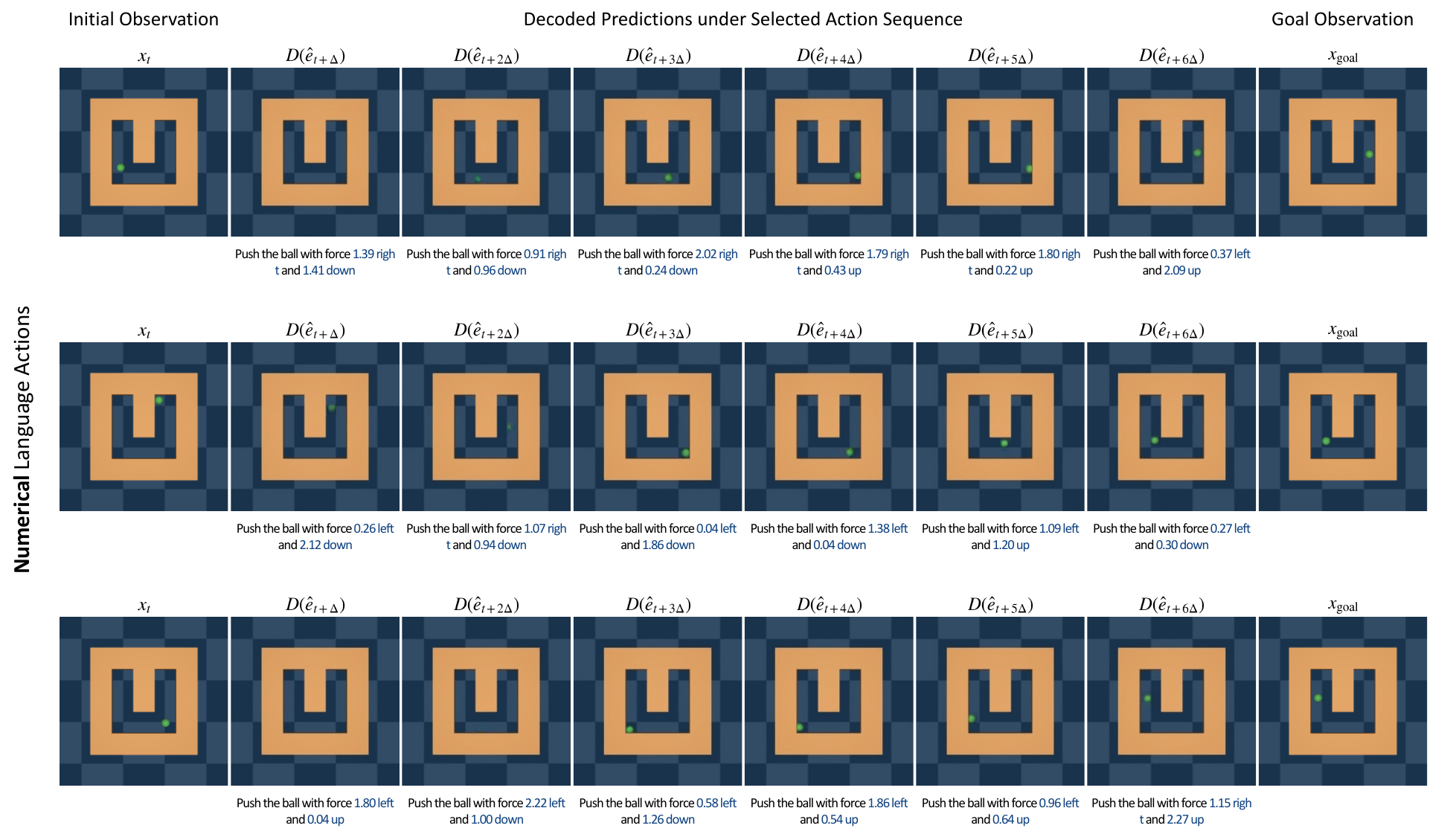}
    \caption{
    \textbf{Numerical language actions on PointMaze.}
    Each row shows initial and goal observations and decoded future predictions under the selected actions, which specify pushing forces and directions.
    }
    \label{fig:wm-num-point}
\end{figure}

%% file: figure-qualitative/world-model/num-pusht.tex
\begin{figure}[p]
    \centering
    \includegraphics[width=\linewidth]{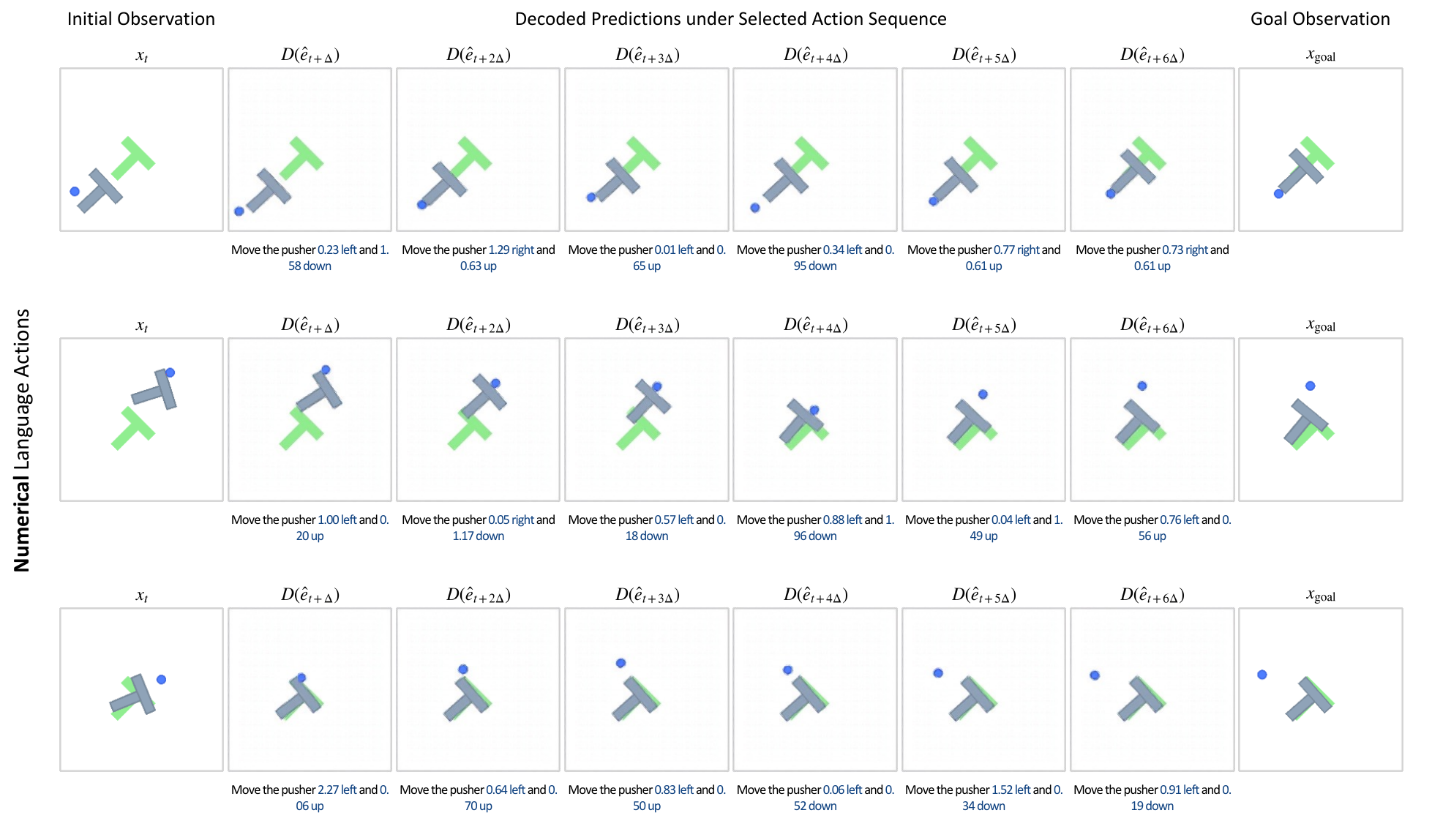}
    \caption{
    \textbf{Numerical language actions on Push-T.}
    Initial and goal observations bracket decoded future predictions under the selected pusher movements, with numerical displacements described below each frame.
    }
    \label{fig:wm-num-pusht}
\end{figure}

%% file: figure-qualitative/world-model/num-meta.tex
\begin{figure}[p]
    \centering
    \includegraphics[width=\linewidth]{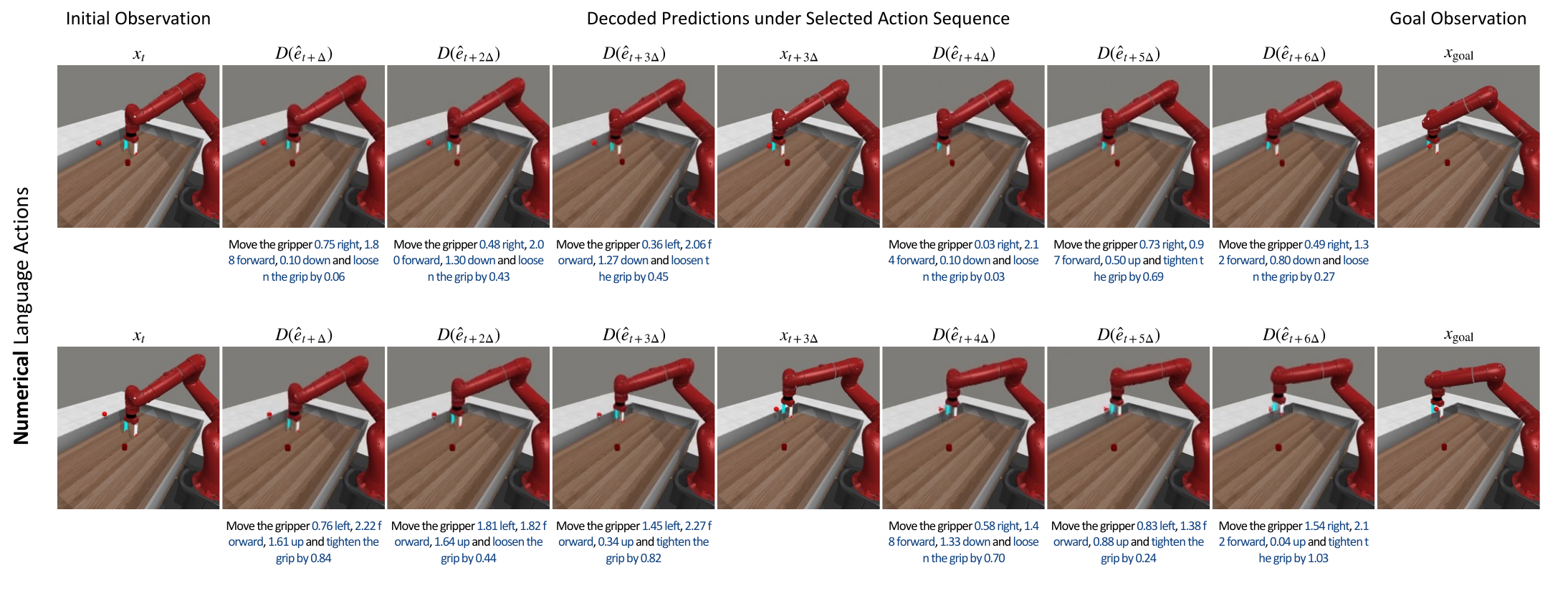}
    \caption{
    \textbf{Numerical language actions on MetaWorld.}
    Decoded predictions are shown before and after replanning from the simulator observation $x_{t+3\Delta}$, alongside initial and goal observations.
    }
    \label{fig:wm-num-meta}
\end{figure}

%% file: figure-qualitative/world-model/num-droid.tex
\begin{figure}[p]
    \centering
    \includegraphics[width=\linewidth]{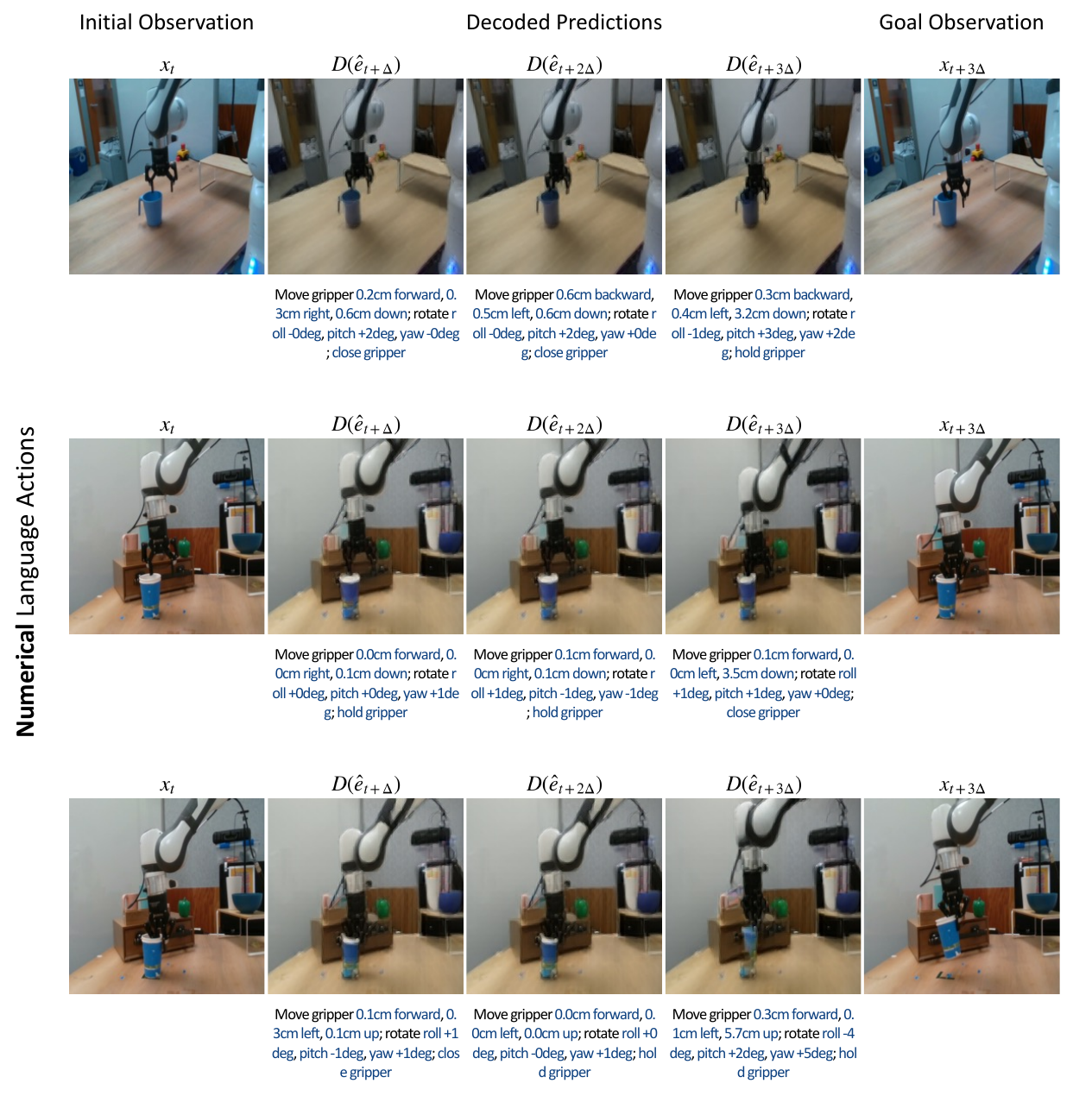}
    \caption{
    \textbf{Numerical language actions on DROID.}
    Each row shows an initial observation, three decoded future predictions, and the goal observation, with language actions below.
    }
    \label{fig:wm-num-droid}
\end{figure}

%% file: figure-qualitative/world-model/num-robo.tex
\begin{figure}[p]
    \centering
    \includegraphics[width=\linewidth]{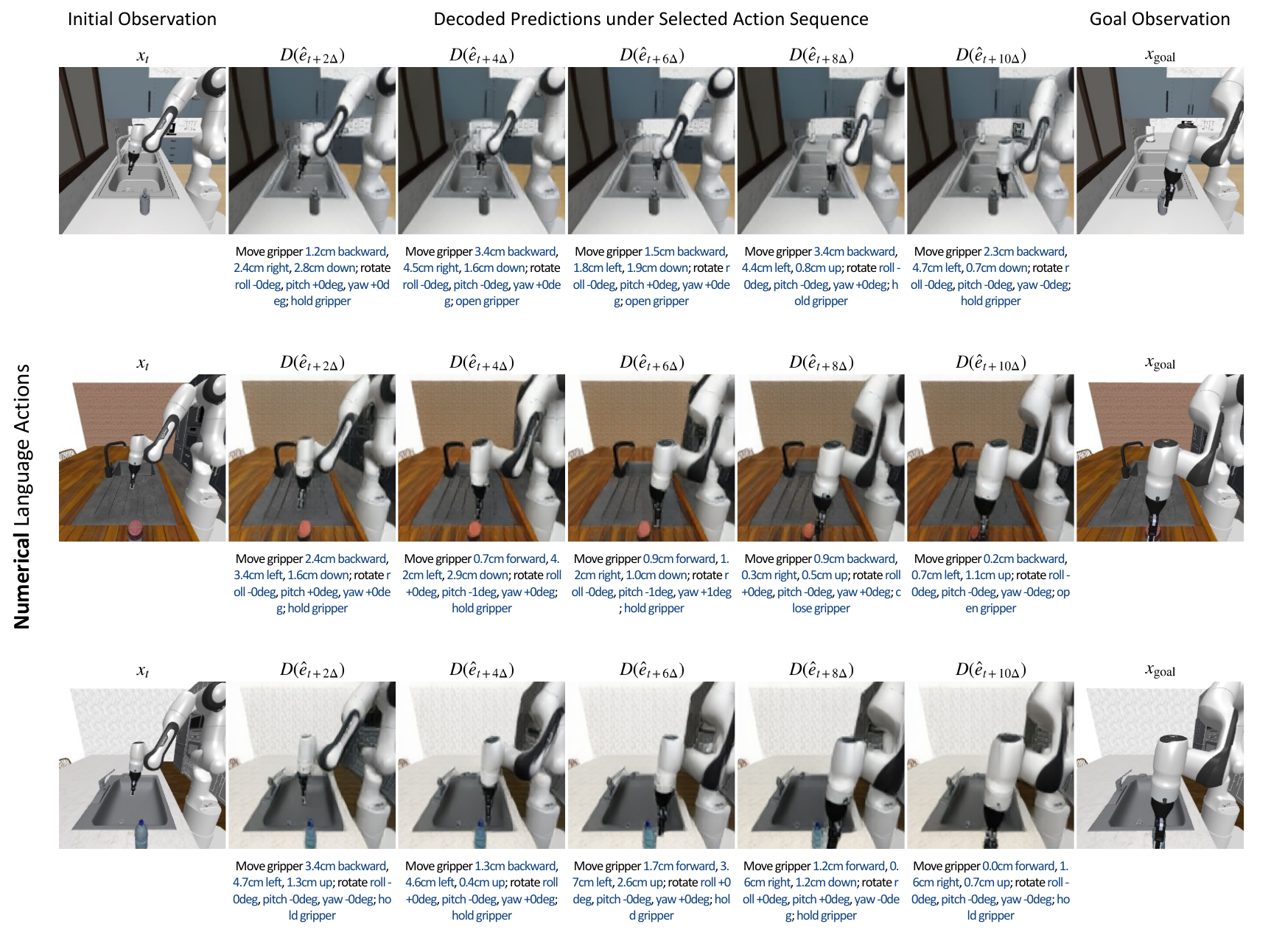}
    \caption{
    \textbf{Numerical language actions on RoboCasa Reach.}
    Decoded future predictions under selected gripper actions are shown between initial and goal observations. RoboCasa is excluded from world-model training.
    }
    \label{fig:wm-num-robo}
\end{figure}

%% file: figure-qualitative/world-model/num-robo-p.tex
\begin{figure}[p]
    \centering
    \includegraphics[width=\linewidth]{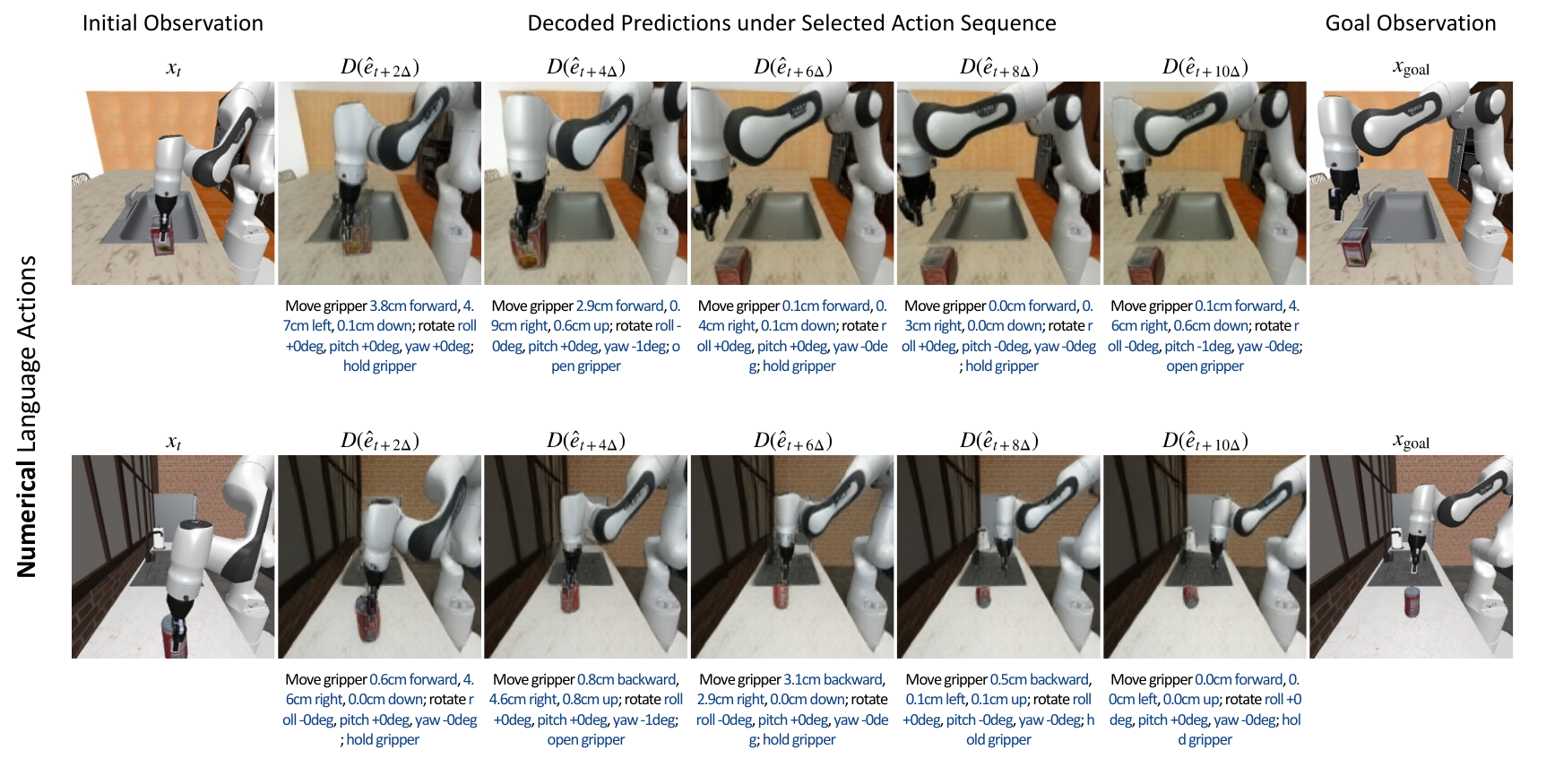}
    \caption{
    \textbf{Numerical language actions on RoboCasa Place.}
    Initial and goal observations bracket decoded future predictions during object placement, with the selected gripper actions below each frame.
    }
    \label{fig:wm-num-robo-p}
\end{figure}

%% file: figure-qualitative/world-model/qual-wall.tex
\begin{figure}[!htbp]
    \centering
    \includegraphics[width=\linewidth]{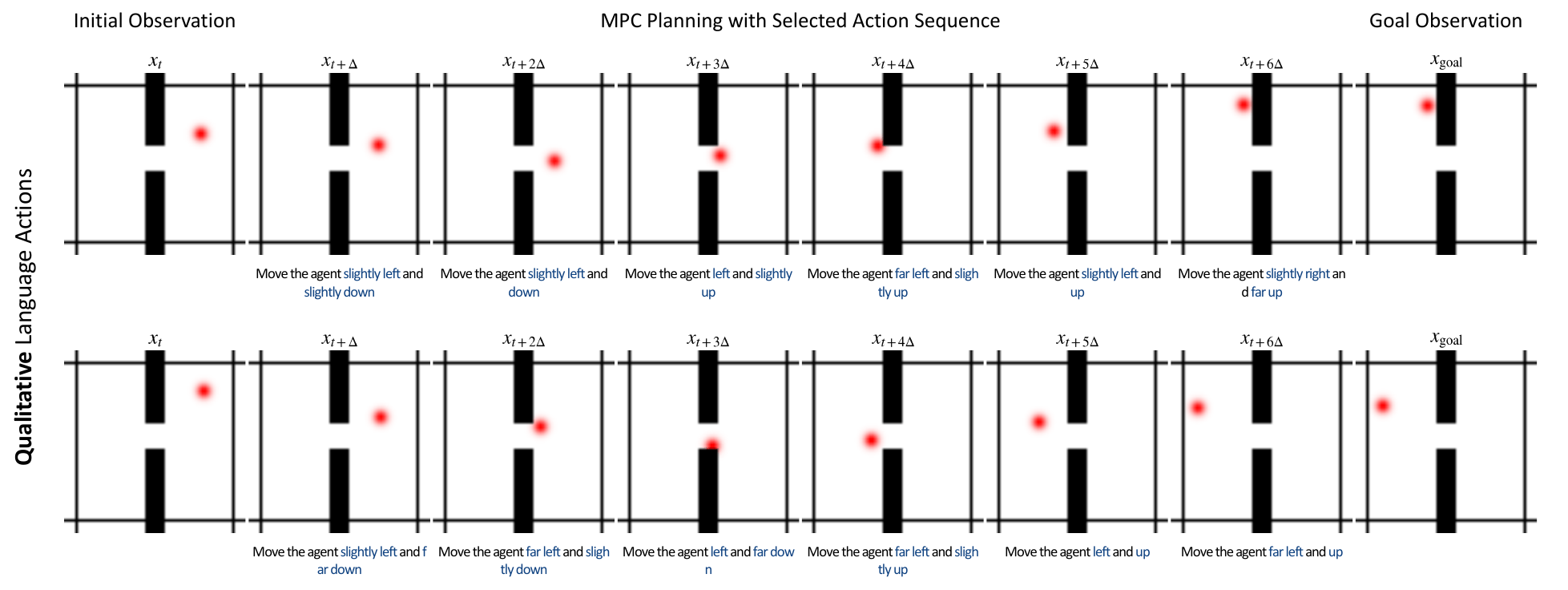}
    \caption{
    \textbf{Qualitative language actions on Wall.}
    Simulator observations show the agent passing through the opening toward the goal. Actions describe directions and relative movement magnitudes.
    }
    \label{fig:wm-qual-wall}
\end{figure}

%% file: figure-qualitative/world-model/qual-point.tex
\begin{figure}[!htbp]
    \centering
    \includegraphics[width=\linewidth]{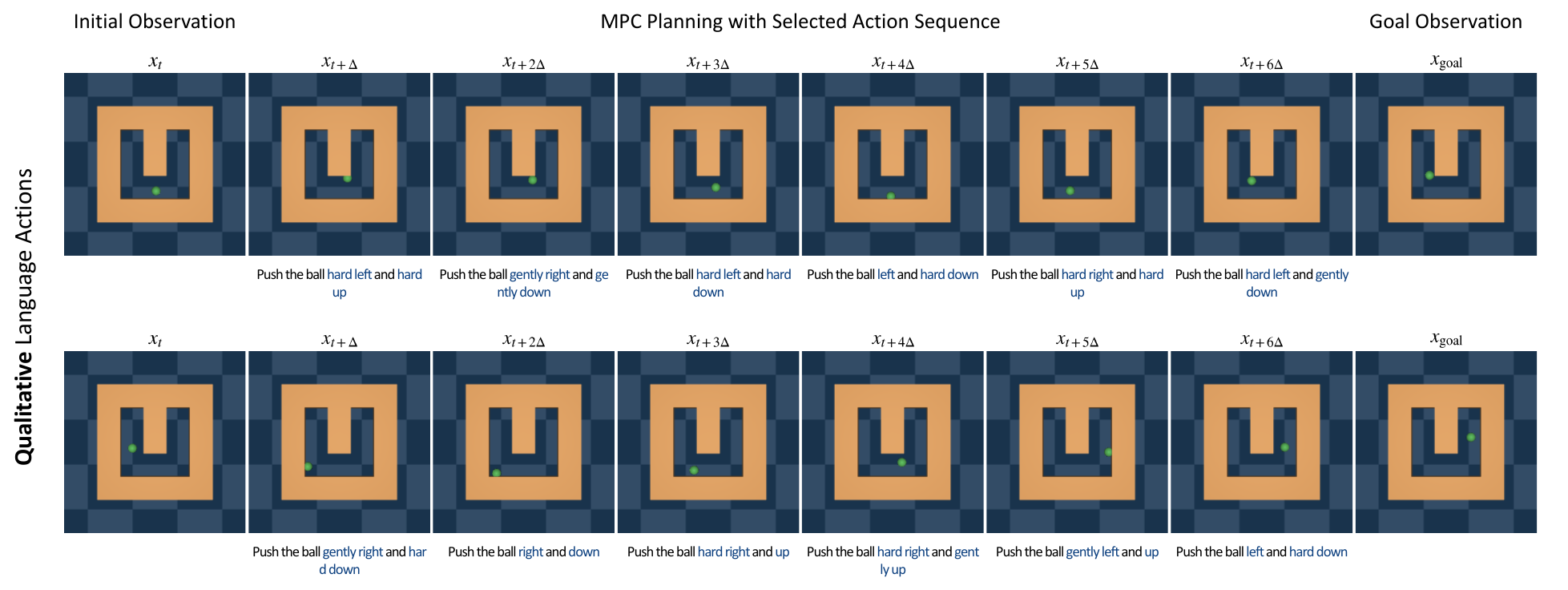}
    \caption{
    \textbf{Qualitative language actions on PointMaze.}
    Simulator observations show the ball moving through the maze under MPC, with pushing directions and qualitative force descriptions below the frames.
    }
    \label{fig:wm-qual-point}
\end{figure}

%% file: figure-qualitative/world-model/qual-pusht.tex
\begin{figure}[!htbp]
    \centering
    \includegraphics[width=\linewidth]{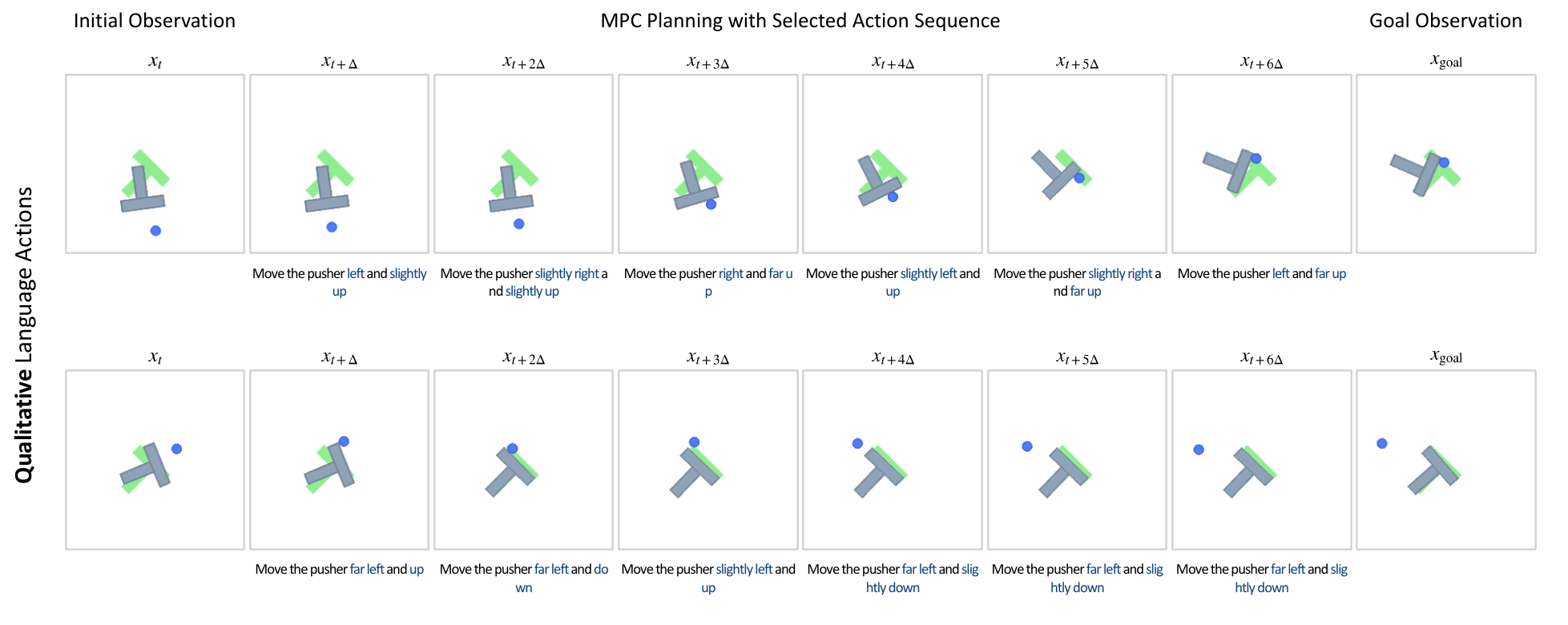}
    \caption{
    \textbf{Qualitative language actions on Push-T.}
    Initial and goal observations bracket simulator states during MPC, with the pusher's direction and relative displacement described below each frame.
    }
    \label{fig:wm-qual-pusht}
\end{figure}

%% file: figure-qualitative/world-model/qual-meta.tex
\begin{figure}[!htbp]
    \centering
    \includegraphics[width=\linewidth]{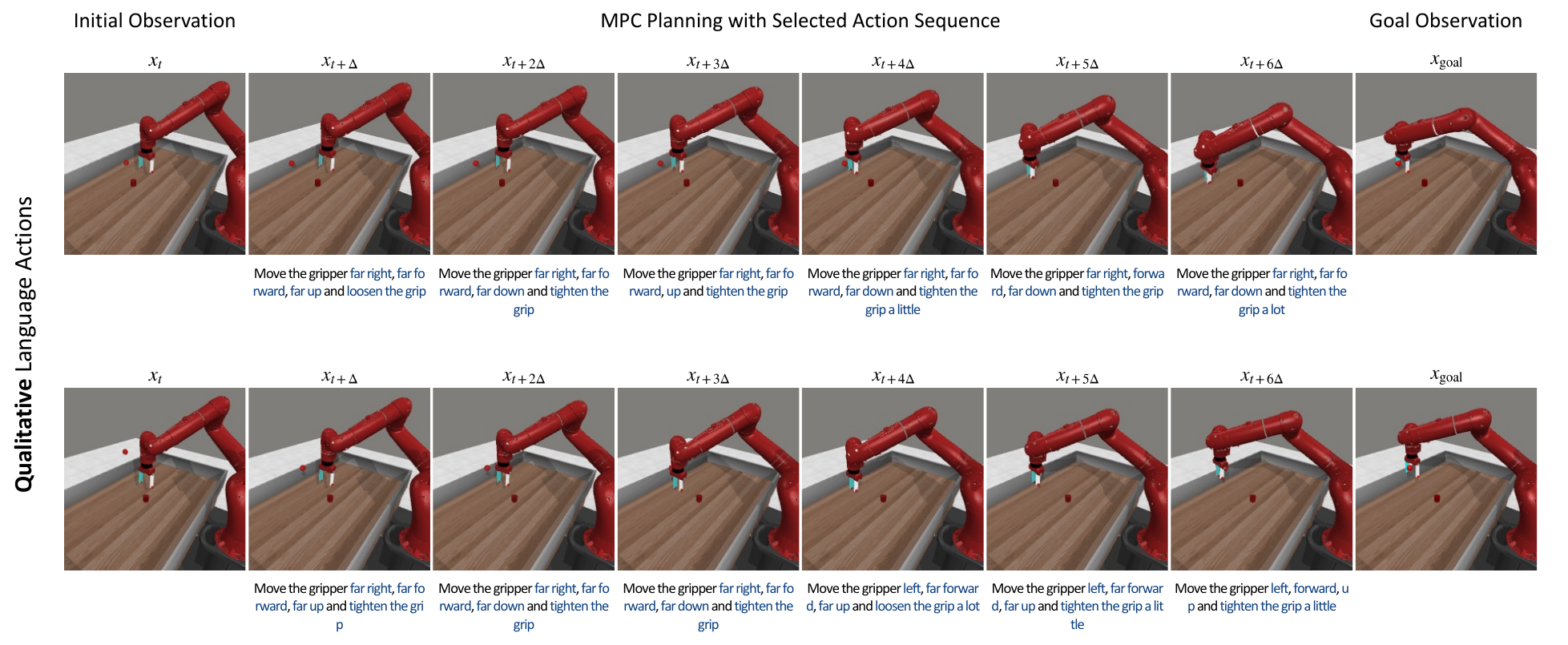}
    \caption{
    \textbf{Qualitative language actions on MetaWorld.}
    Simulator observations show reaching trajectories planned using descriptions of gripper motion and grip adjustment without exact numerical magnitudes.
    }
    \label{fig:wm-qual-meta}
\end{figure}

%% file: figure-qualitative/world-model/qual-robo.tex
\begin{figure}[!htbp]
    \centering
    \includegraphics[width=\linewidth]{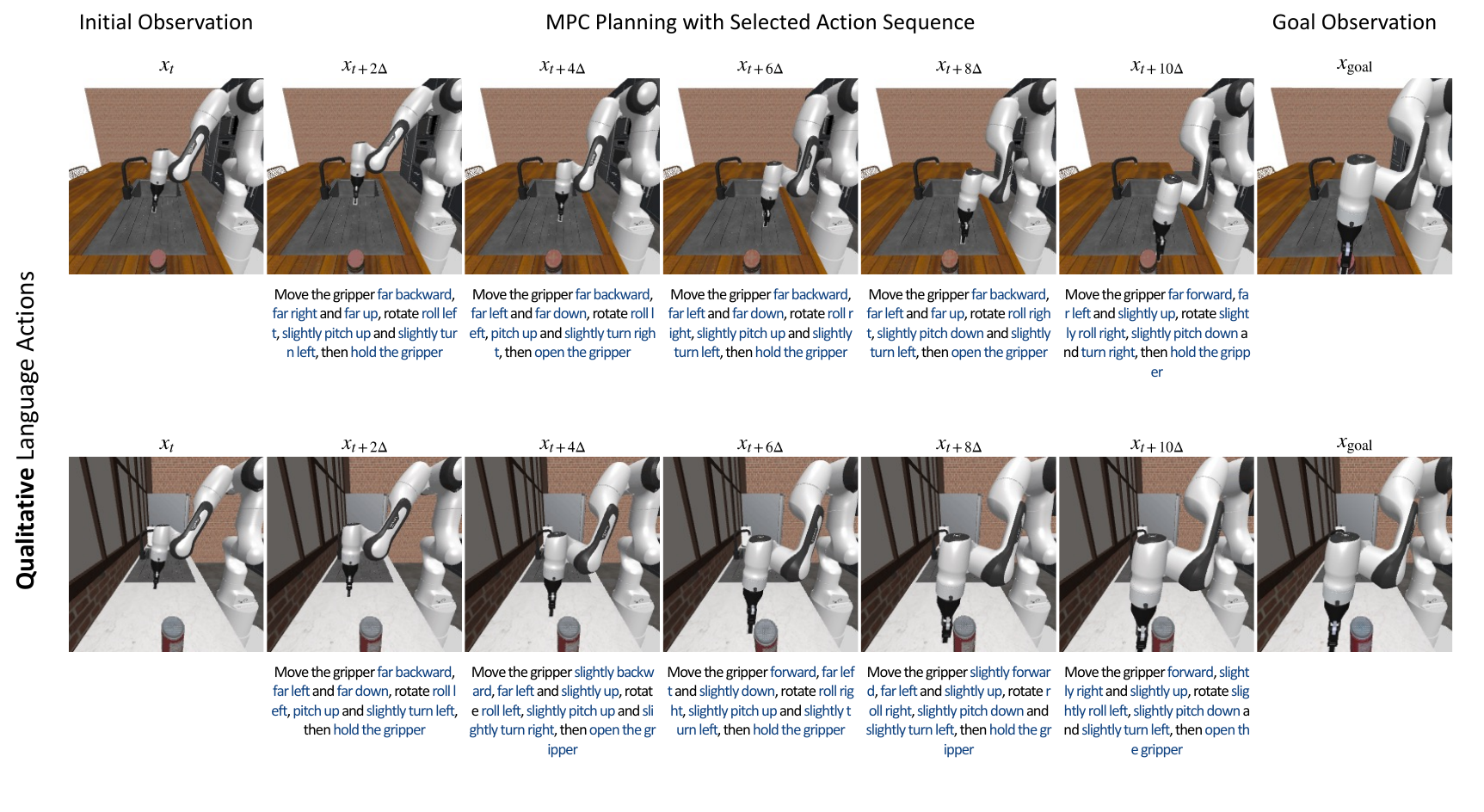}
    \caption{
    \textbf{Qualitative language actions on RoboCasa Reach.}
    Simulator observations show zero-shot MPC execution toward the goal, conditioned on qualitative descriptions of gripper translation, rotation, and opening or closing.
    }
    \label{fig:wm-qual-robo}
\end{figure}

%% file: figure-qualitative/world-model/qual-robo-p.tex
\begin{figure}[!htbp]
    \centering
    \includegraphics[width=\linewidth]{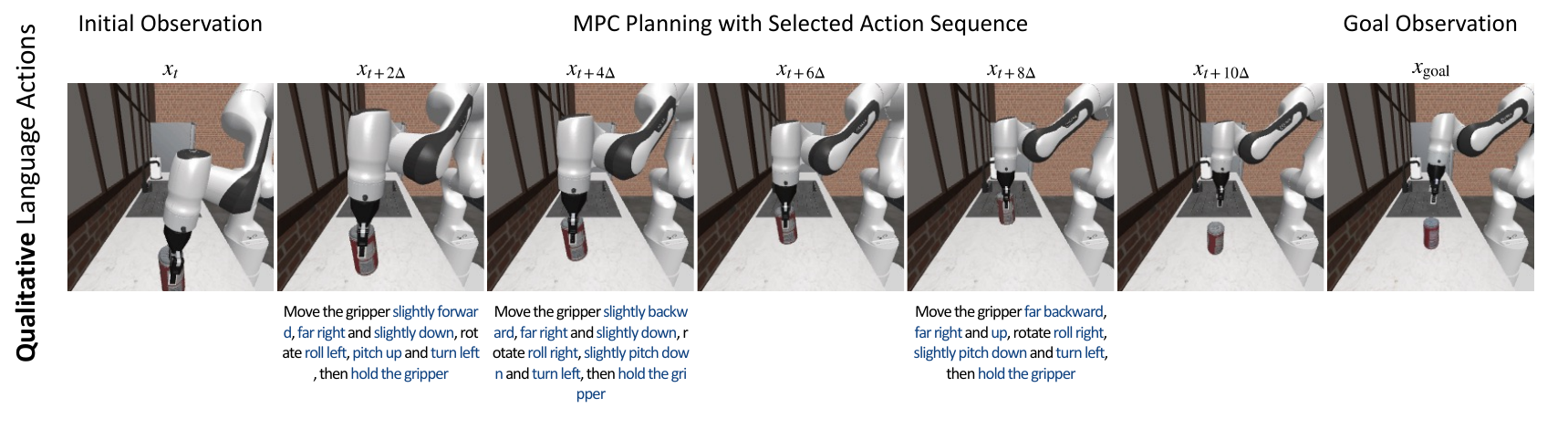}
    \caption{
    \textbf{Qualitative language actions on RoboCasa Place.}
    Simulator observations show an object-placement trajectory under MPC, with qualitative gripper actions and the goal observation for reference.
    }
    \label{fig:wm-qual-robo-p}
\end{figure}

%% file: figure-qualitative/policy/all.tex
\input{figure-qualitative/policy/libero-spatial-1}
\input{figure-qualitative/policy/libero-spatial-2}
\input{figure-qualitative/policy/libero-object-1}
\input{figure-qualitative/policy/libero-object-2}
\input{figure-qualitative/policy/libero-goal-1}
\input{figure-qualitative/policy/libero-goal-2}
\input{figure-qualitative/policy/libero-long-1}
\input{figure-qualitative/policy/libero-long-2}

\clearpage
\input{figure-qualitative/policy/robocasa-coffee-1}
\input{figure-qualitative/policy/robocasa-coffee-2}
\input{figure-qualitative/policy/robocasa-open-1}
\input{figure-qualitative/policy/robocasa-open-2}
\input{figure-qualitative/policy/robocasa-pick-1}
\input{figure-qualitative/policy/robocasa-pick-2}
\input{figure-qualitative/policy/robocasa-turn-1}

%% file: figure-qualitative/policy/libero-spatial-1.tex
\begin{figure}[p]
    \centering
    \includegraphics[width=\linewidth,height=0.88\textheight,keepaspectratio]{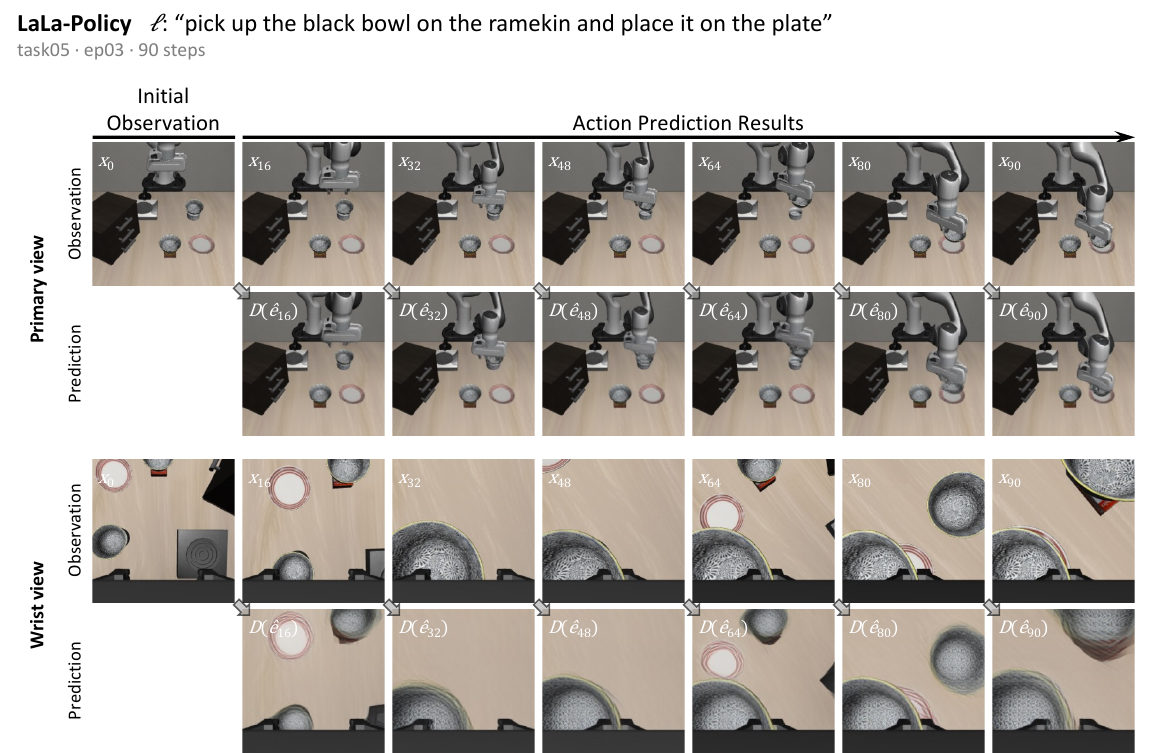}
    \caption{
    \textbf{LIBERO Spatial: bowl on a ramekin.}
    \ourpolicy\ moves the bowl from the ramekin onto the plate. Primary and wrist views compare simulator observations (upper rows) with decoded future predictions (lower rows).
    }
    \label{fig:policy-libero-spatial-1}
\end{figure}

%% file: figure-qualitative/policy/libero-spatial-2.tex
\begin{figure}[p]
    \centering
    \includegraphics[width=\linewidth,height=0.88\textheight,keepaspectratio]{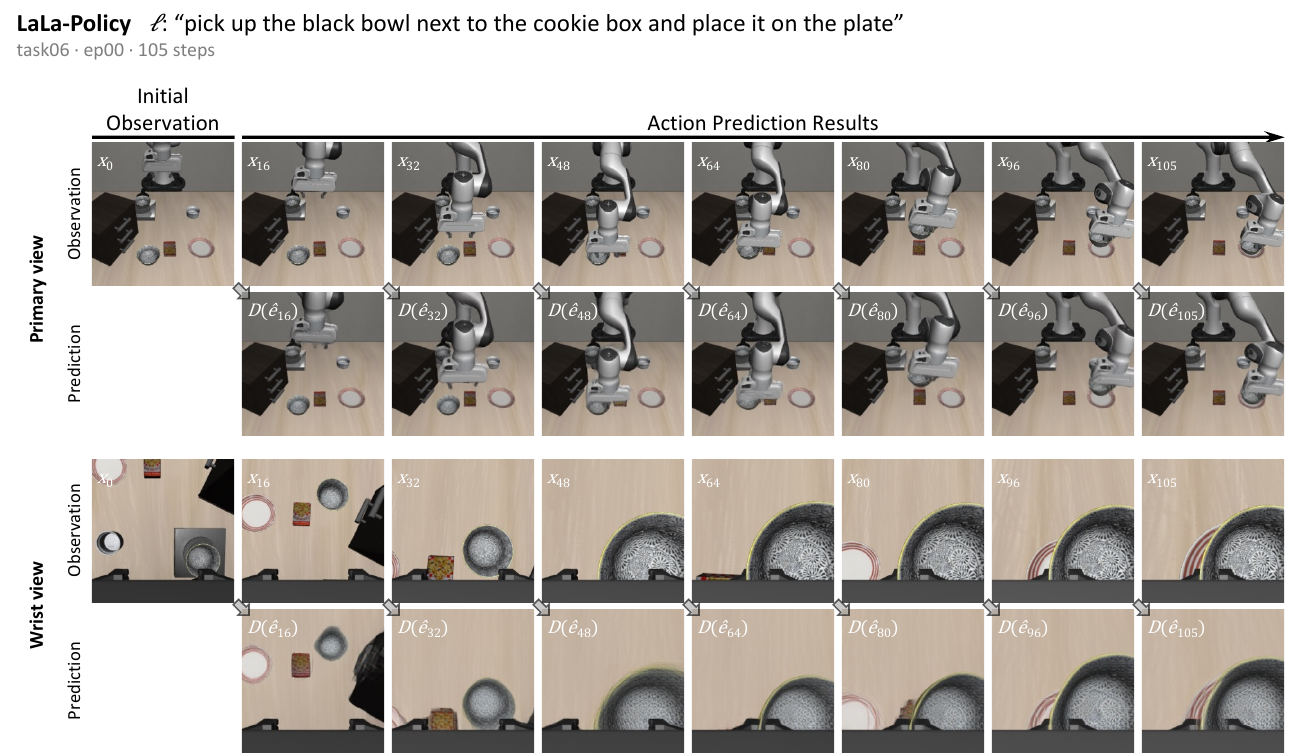}
    \caption{
    \textbf{LIBERO Spatial: bowl next to a cookie box.}
    \ourpolicy\ moves the specified bowl onto the plate. Simulator observations and decoded future predictions are paired for the primary and wrist views.
    }
    \label{fig:policy-libero-spatial-2}
\end{figure}

%% file: figure-qualitative/policy/libero-object-1.tex
\begin{figure}[p]
    \centering
    \includegraphics[width=\linewidth,height=0.88\textheight,keepaspectratio]{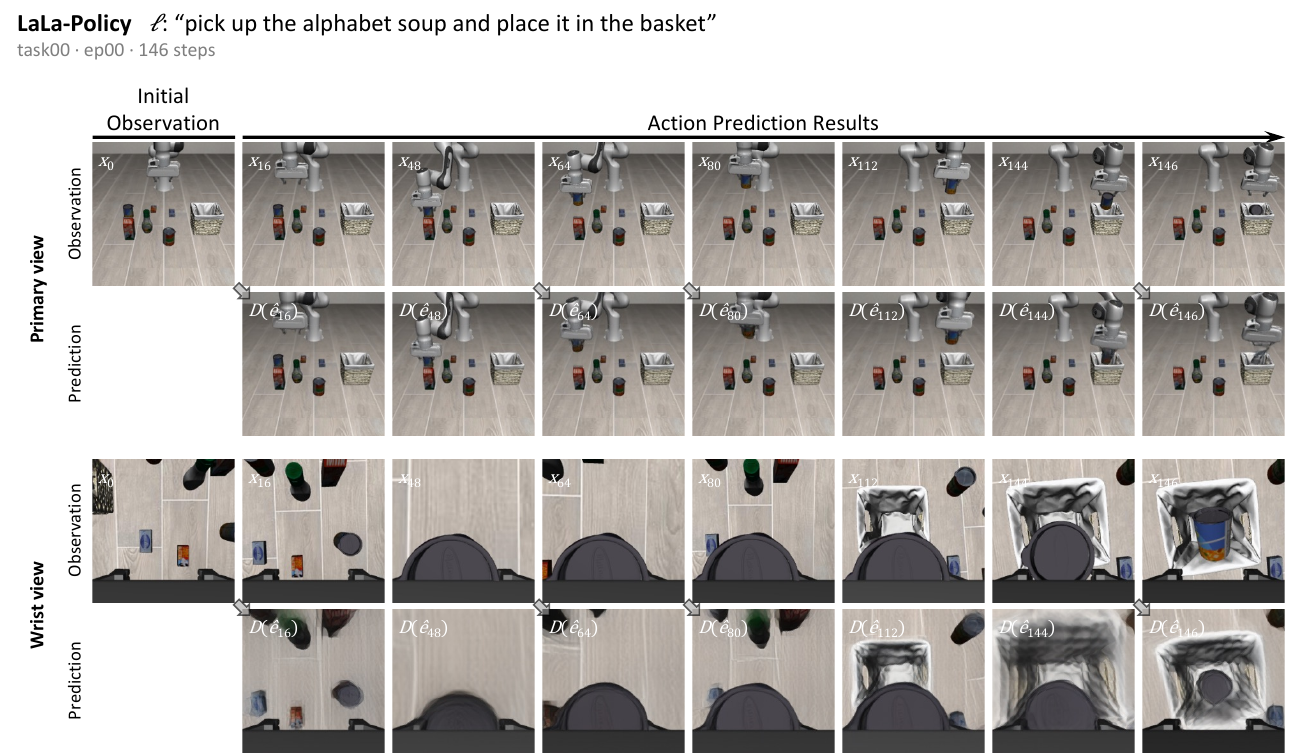}
    \caption{
    \textbf{LIBERO Object: alphabet soup.}
    \ourpolicy\ picks up the alphabet soup and places it in the basket. Each camera view pairs observed states with decoded future predictions.
    }
    \label{fig:policy-libero-object-1}
\end{figure}

%% file: figure-qualitative/policy/libero-object-2.tex
\begin{figure}[p]
    \centering
    \includegraphics[width=\linewidth,height=0.88\textheight,keepaspectratio]{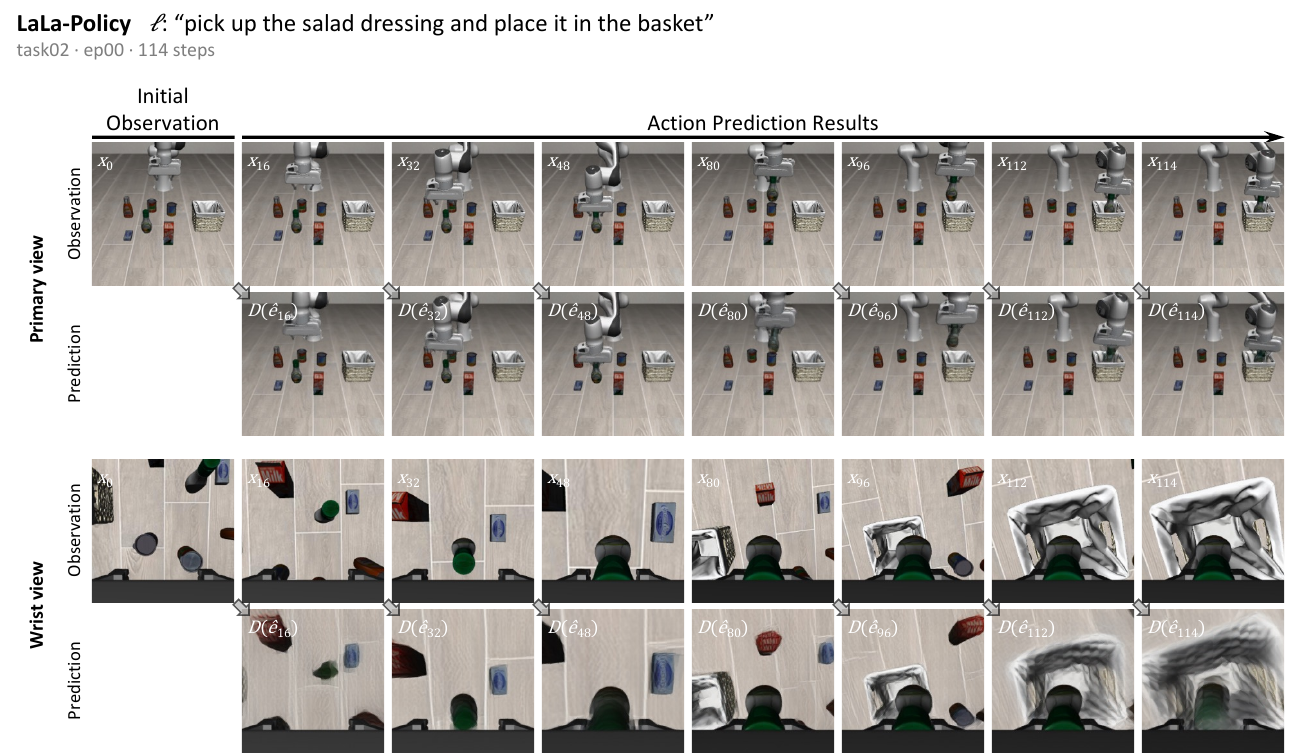}
    \caption{
    \textbf{LIBERO Object: salad dressing.}
    \ourpolicy\ transfers the salad dressing to the basket. Primary and wrist views show simulator observations above the corresponding decoded future predictions.
    }
    \label{fig:policy-libero-object-2}
\end{figure}

%% file: figure-qualitative/policy/libero-goal-1.tex
\begin{figure}[p]
    \centering
    \includegraphics[width=\linewidth,height=0.88\textheight,keepaspectratio]{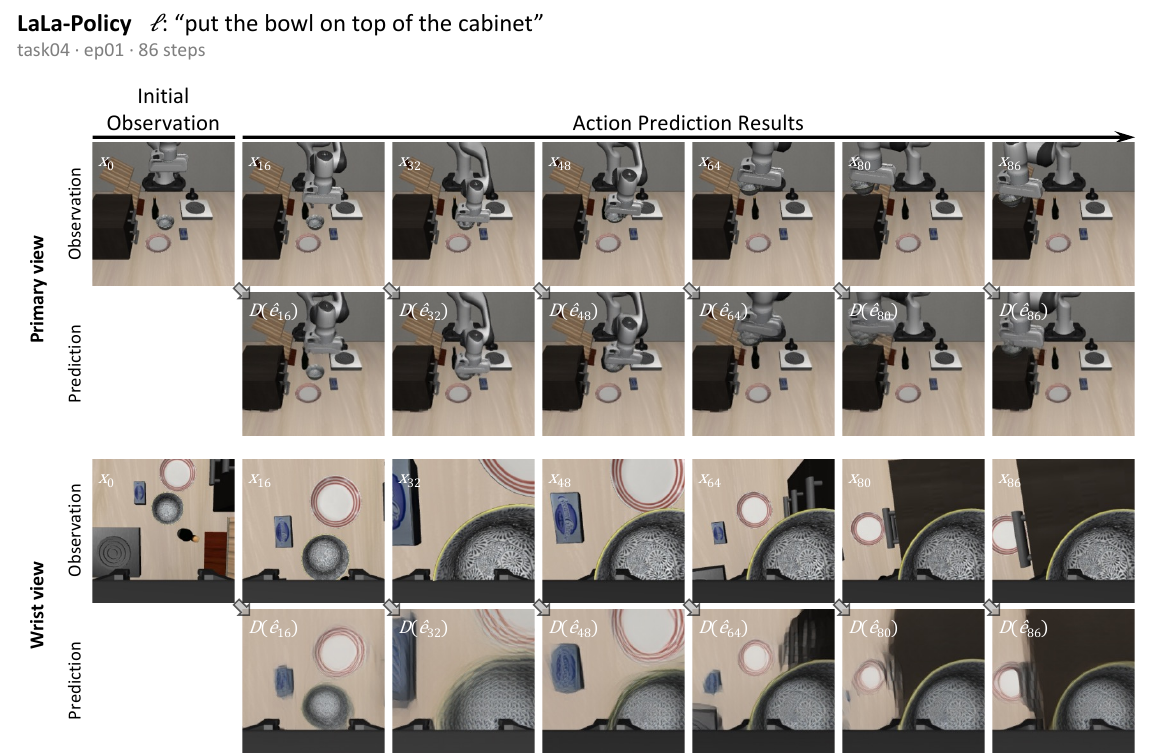}
    \caption{
    \textbf{LIBERO Goal: bowl on the cabinet.}
    \ourpolicy\ places the bowl on top of the cabinet. Primary and wrist views compare observed states with decoded future predictions throughout execution.
    }
    \label{fig:policy-libero-goal-1}
\end{figure}

%% file: figure-qualitative/policy/libero-goal-2.tex
\begin{figure}[p]
    \centering
    \includegraphics[width=\linewidth,height=0.88\textheight,keepaspectratio]{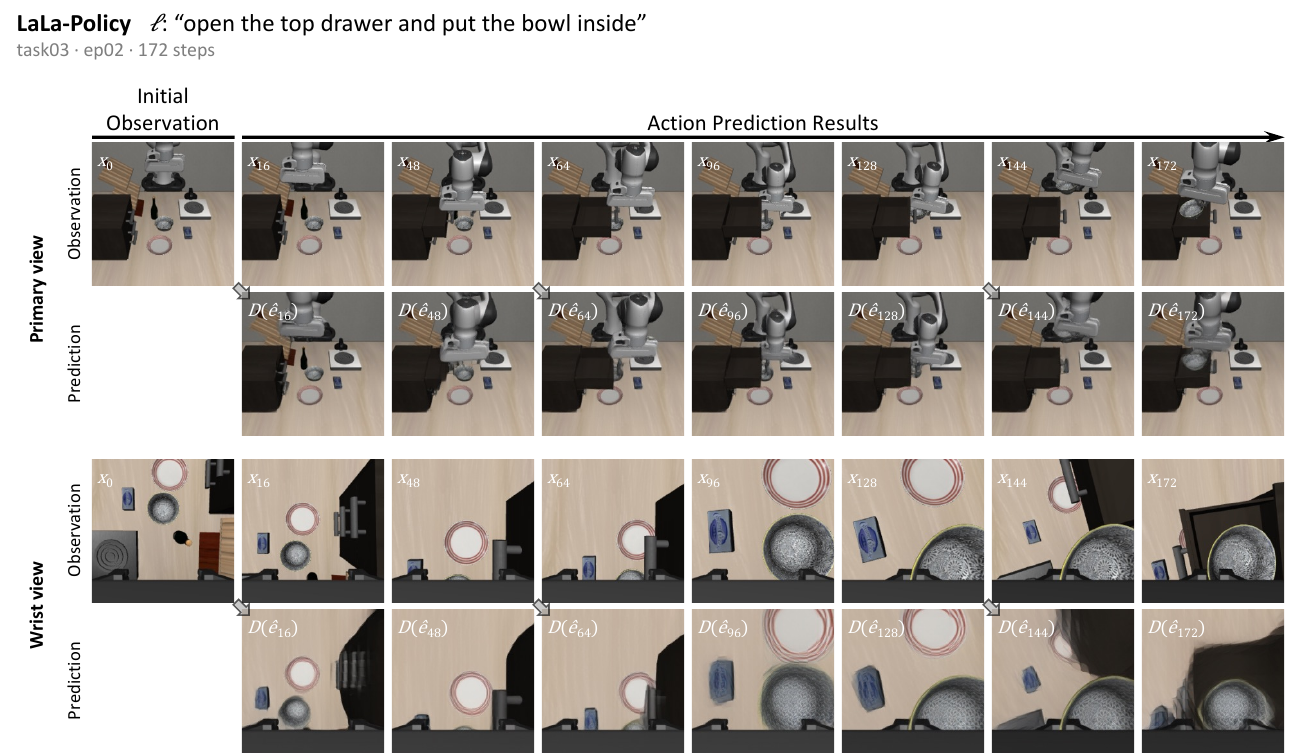}
    \caption{
    \textbf{LIBERO Goal: bowl inside the drawer.}
    \ourpolicy\ opens the top drawer and places the bowl inside. Both camera views pair simulator observations with decoded future predictions.
    }
    \label{fig:policy-libero-goal-2}
\end{figure}

%% file: figure-qualitative/policy/libero-long-1.tex
\begin{figure}[p]
    \centering
    \includegraphics[width=\linewidth,height=0.88\textheight,keepaspectratio]{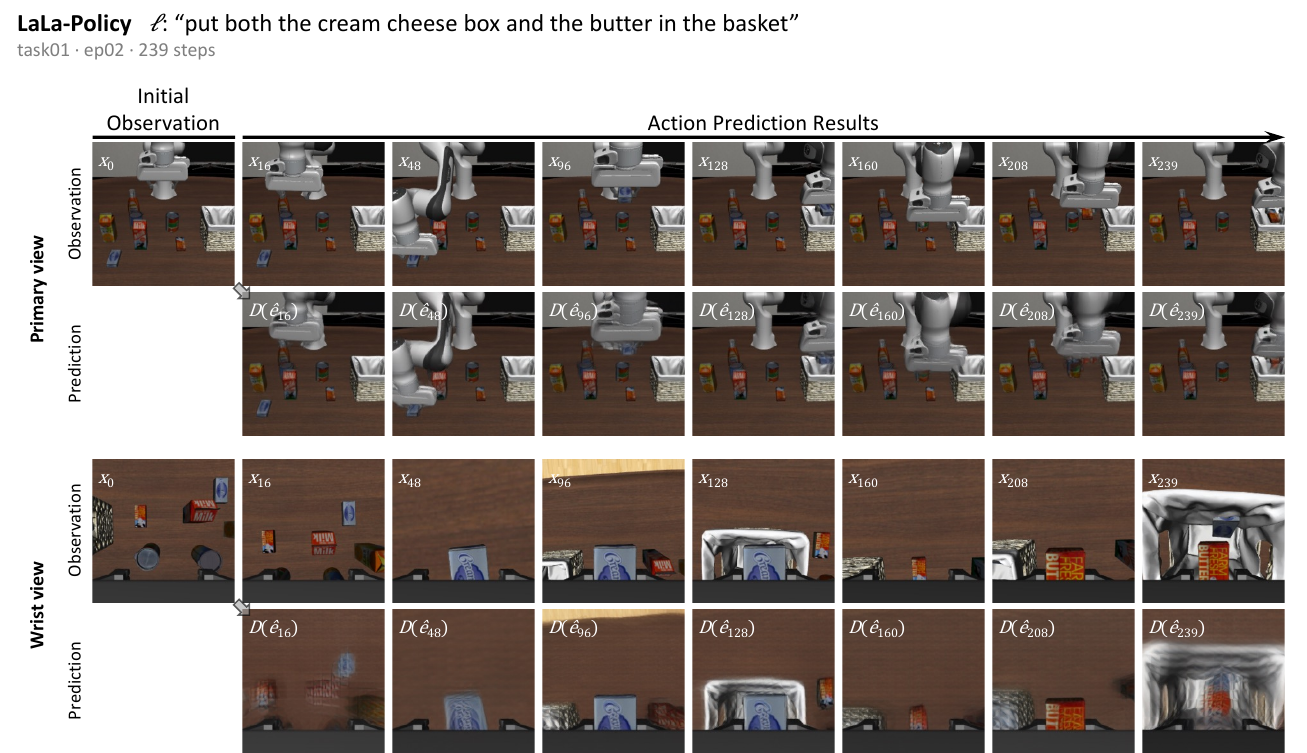}
    \caption{
    \textbf{LIBERO Long: placing two objects in a basket.}
    \ourpolicy\ transfers the cream cheese box and butter to the basket. Observations and decoded future predictions are shown for both camera views.
    }
    \label{fig:policy-libero-long-1}
\end{figure}

%% file: figure-qualitative/policy/libero-long-2.tex
\begin{figure}[p]
    \centering
    \includegraphics[width=\linewidth,height=0.88\textheight,keepaspectratio]{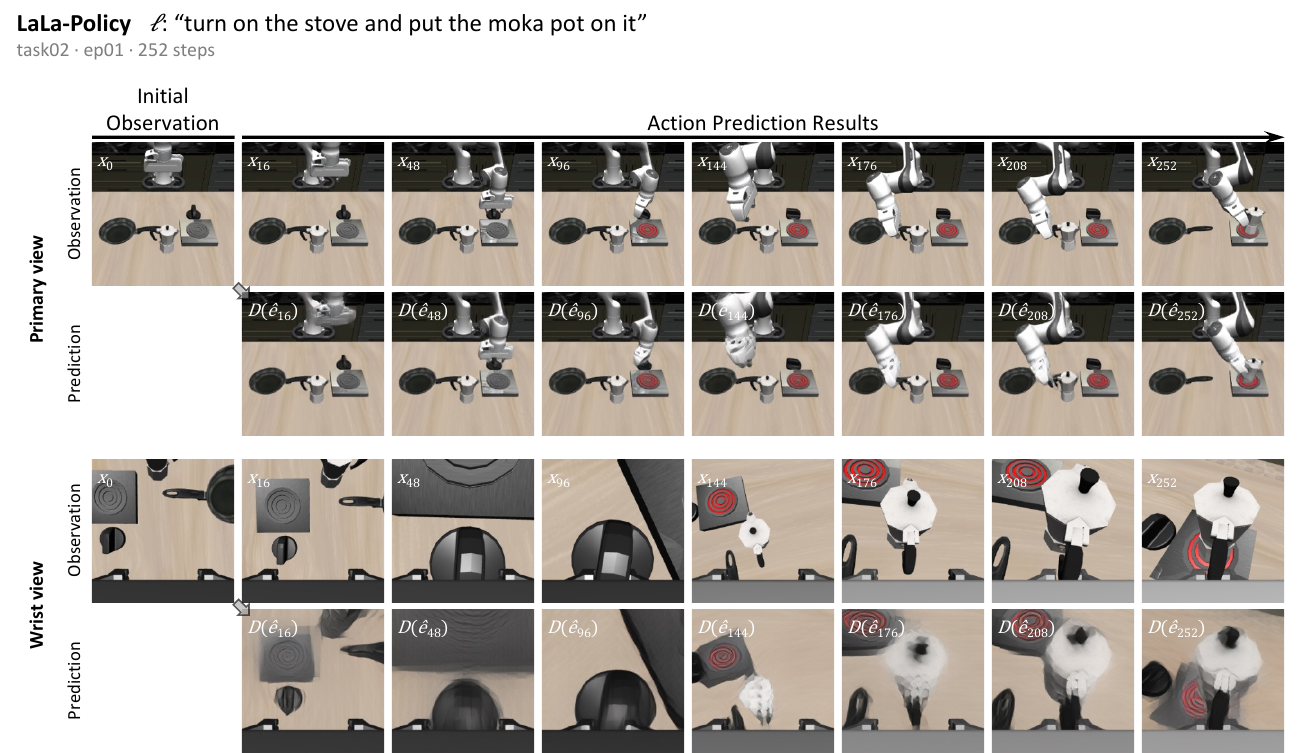}
    \caption{
    \textbf{LIBERO Long: stove and moka pot.}
    \ourpolicy\ turns on the stove and places the moka pot on it. Primary and wrist views compare simulator observations with decoded future predictions.
    }
    \label{fig:policy-libero-long-2}
\end{figure}

%% file: figure-qualitative/policy/robocasa-coffee-1.tex
\begin{figure}[!htbp]
    \centering
    \includegraphics[width=0.7\linewidth,height=0.88\textheight,keepaspectratio]{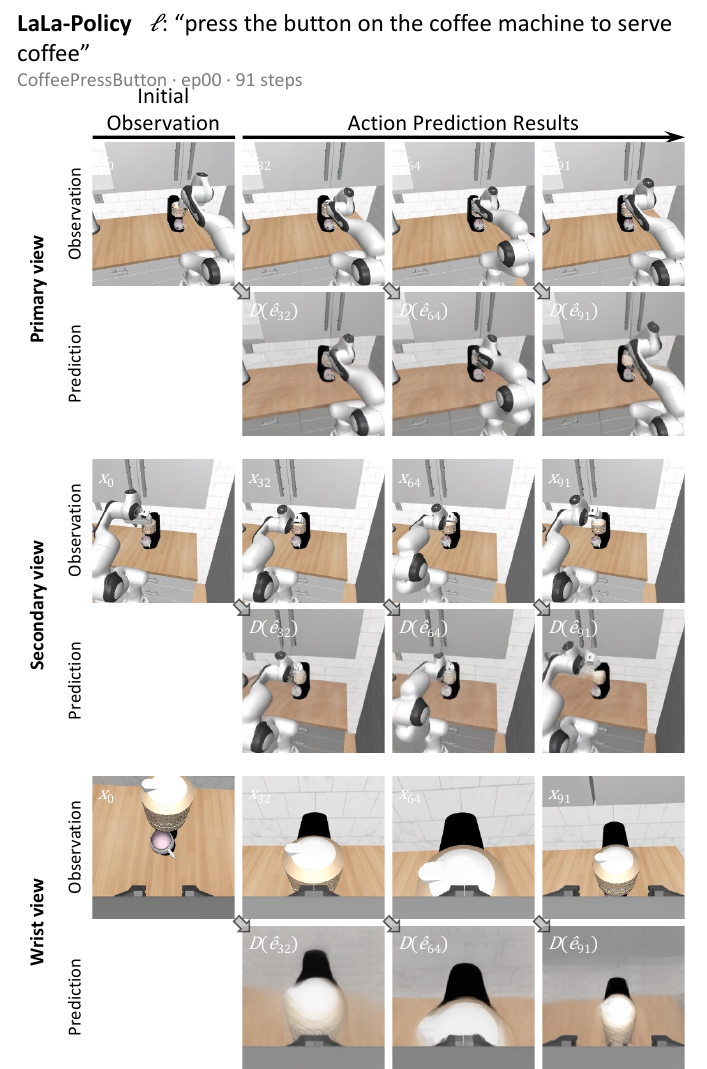}
    \caption{
    \textbf{RoboCasa: pressing the coffee-machine button.}
    Simulator observations (upper rows) and decoded future predictions (lower rows) show \ourpolicy\ pressing the button, from two external views and the wrist view.
    }
    \label{fig:policy-robocasa-coffee-1}
\end{figure}

%% file: figure-qualitative/policy/robocasa-coffee-2.tex
\begin{figure}[!htbp]
    \centering
    \includegraphics[width=\linewidth,height=0.88\textheight,keepaspectratio]{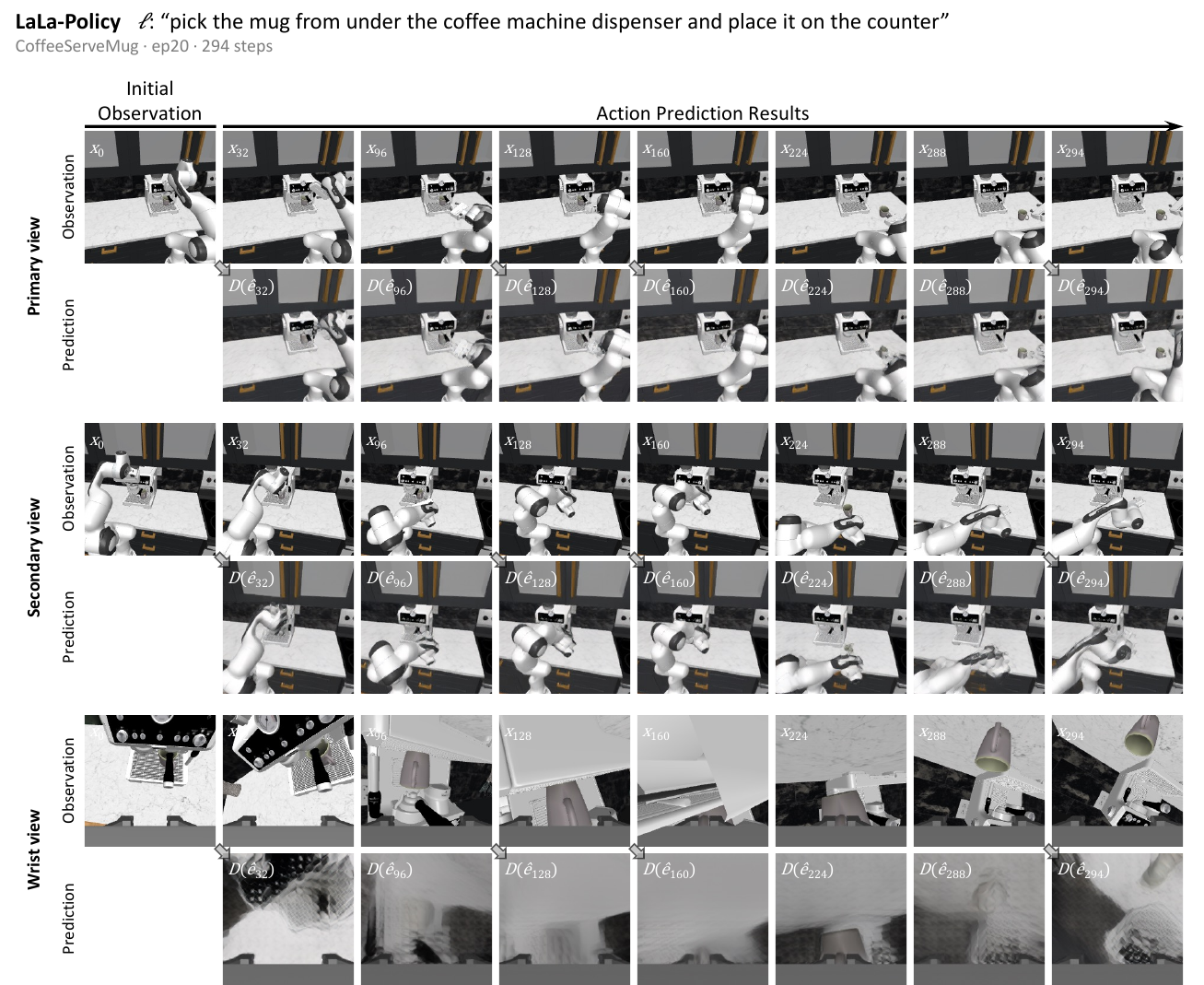}
    \caption{
    \textbf{RoboCasa: serving a mug.}
    \ourpolicy\ moves the mug from beneath the coffee-machine dispenser to the counter. Each of the three views compares observed states with decoded future predictions.
    }
    \label{fig:policy-robocasa-coffee-2}
\end{figure}

%% file: figure-qualitative/policy/robocasa-open-1.tex
\begin{figure}[!htbp]
    \centering
    \includegraphics[width=\linewidth,height=0.88\textheight,keepaspectratio]{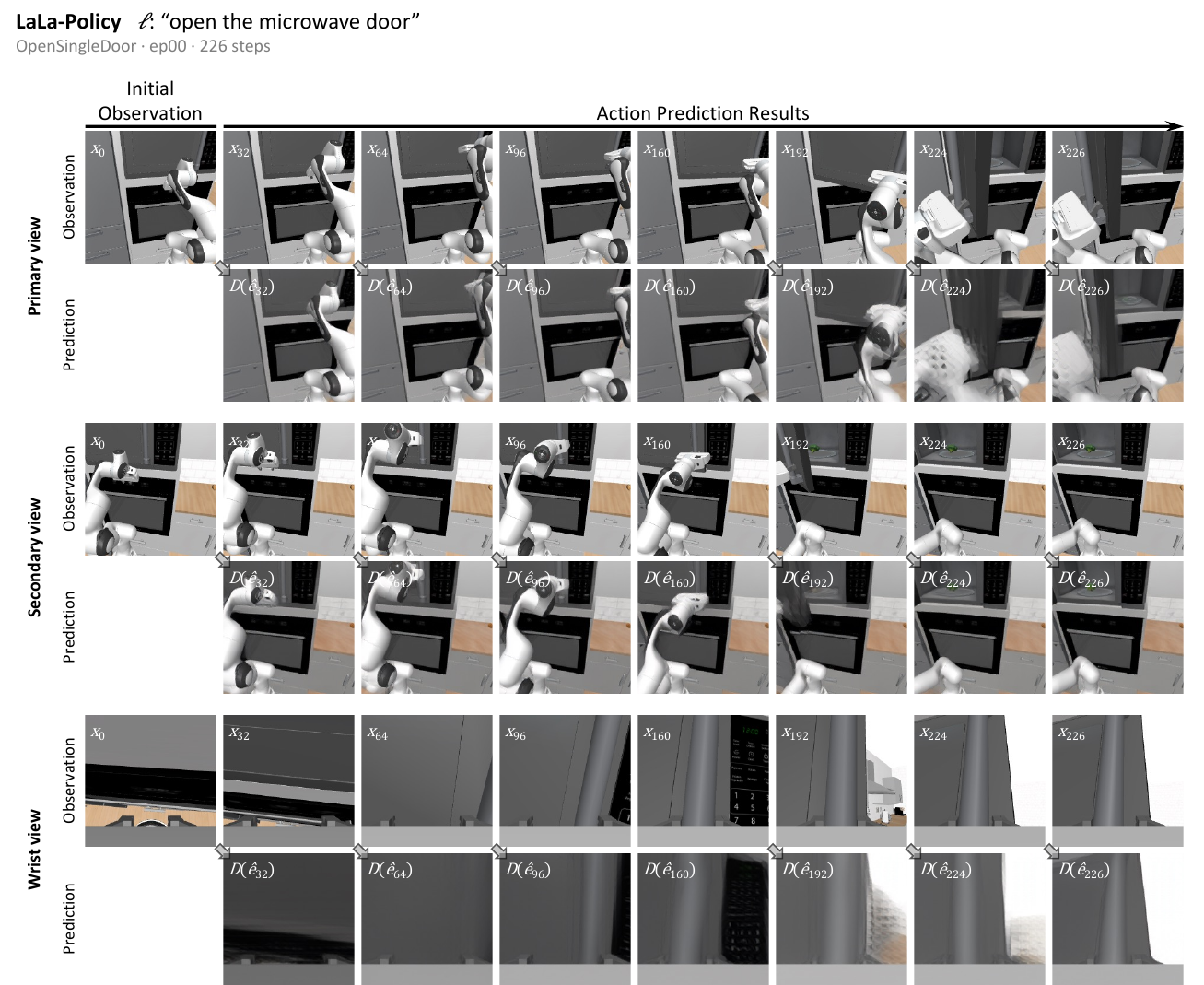}
    \caption{
    \textbf{RoboCasa: opening a microwave.}
    Simulator observations and decoded future predictions show \ourpolicy\ opening the microwave door, from the primary, secondary, and wrist views.
    }
    \label{fig:policy-robocasa-open-1}
\end{figure}

%% file: figure-qualitative/policy/robocasa-open-2.tex
\begin{figure}[!htbp]
    \centering
    \includegraphics[width=\linewidth,height=0.88\textheight,keepaspectratio]{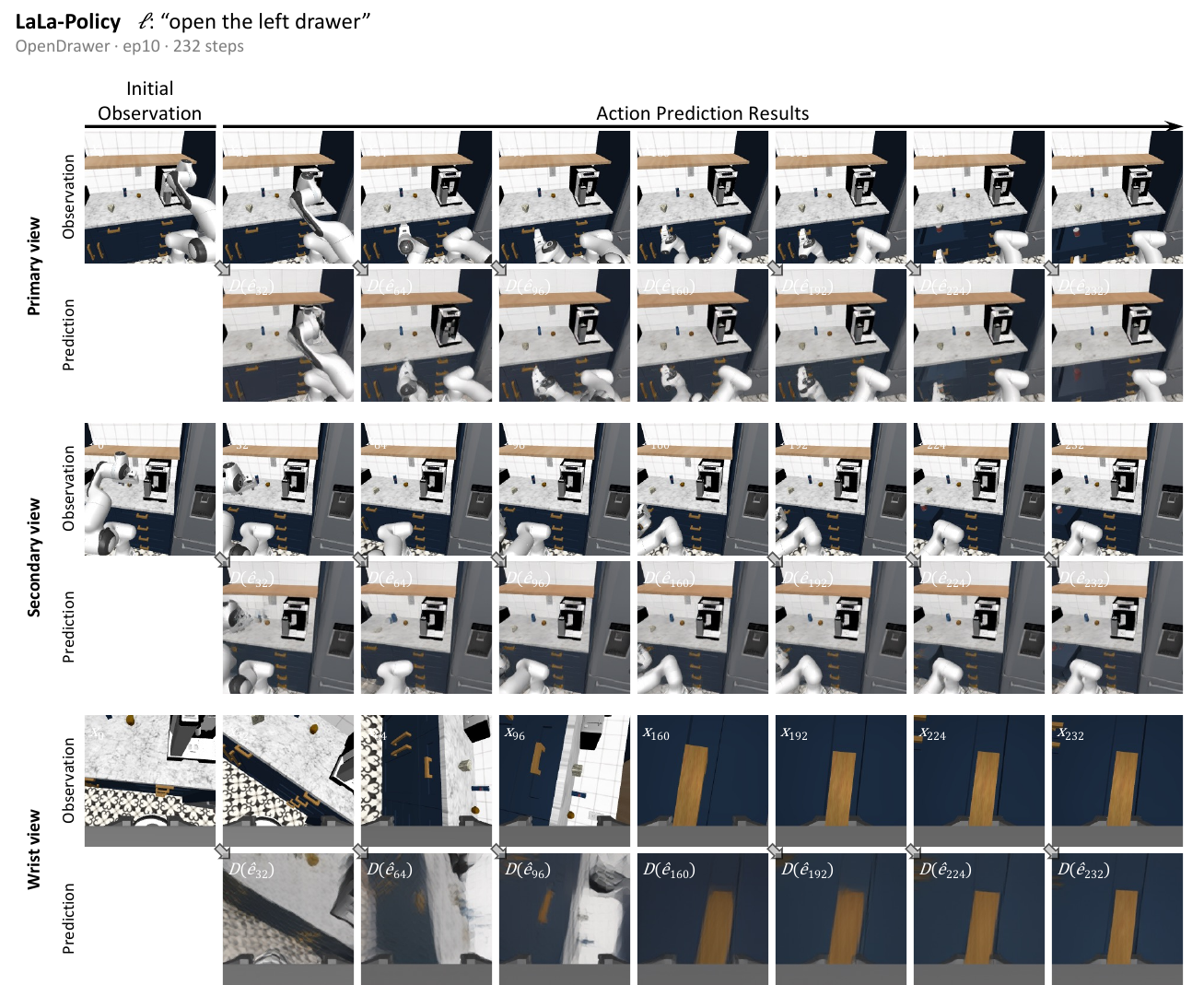}
    \caption{
    \textbf{RoboCasa: opening a drawer.}
    \ourpolicy\ opens the left drawer. The three camera views pair simulator observations with decoded future predictions during reaching and drawer opening.
    }
    \label{fig:policy-robocasa-open-2}
\end{figure}

%% file: figure-qualitative/policy/robocasa-pick-1.tex
\begin{figure}[!htbp]
    \centering
    \includegraphics[width=\linewidth,height=0.88\textheight,keepaspectratio]{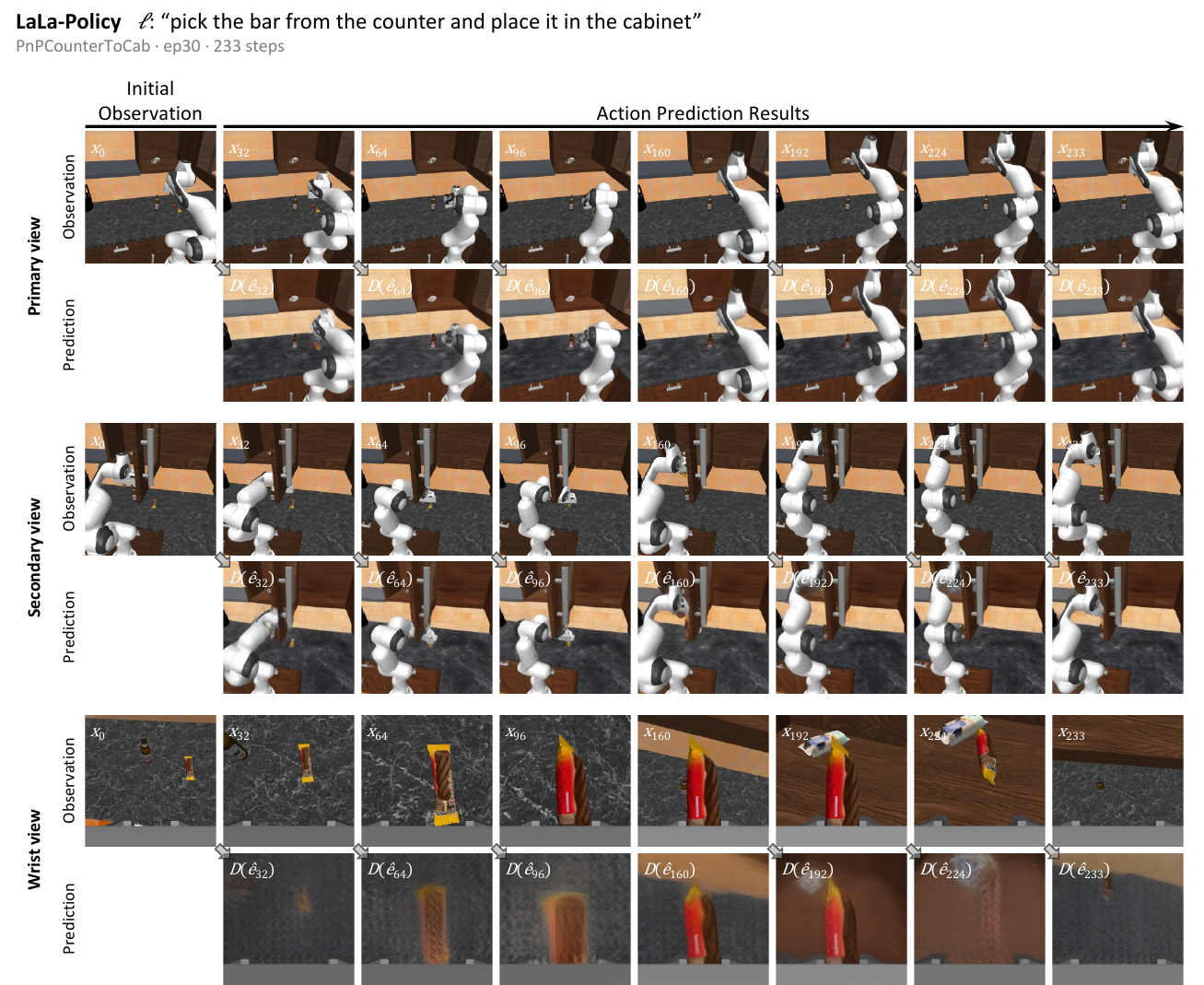}
    \caption{
    \textbf{RoboCasa: counter-to-cabinet transfer.}
    \ourpolicy\ picks up the bar from the counter and places it in the cabinet. Each camera view compares observed states with decoded future predictions.
    }
    \label{fig:policy-robocasa-pick-1}
\end{figure}

%% file: figure-qualitative/policy/robocasa-pick-2.tex
\begin{figure}[!htbp]
    \centering
    \includegraphics[width=\linewidth,height=0.88\textheight,keepaspectratio]{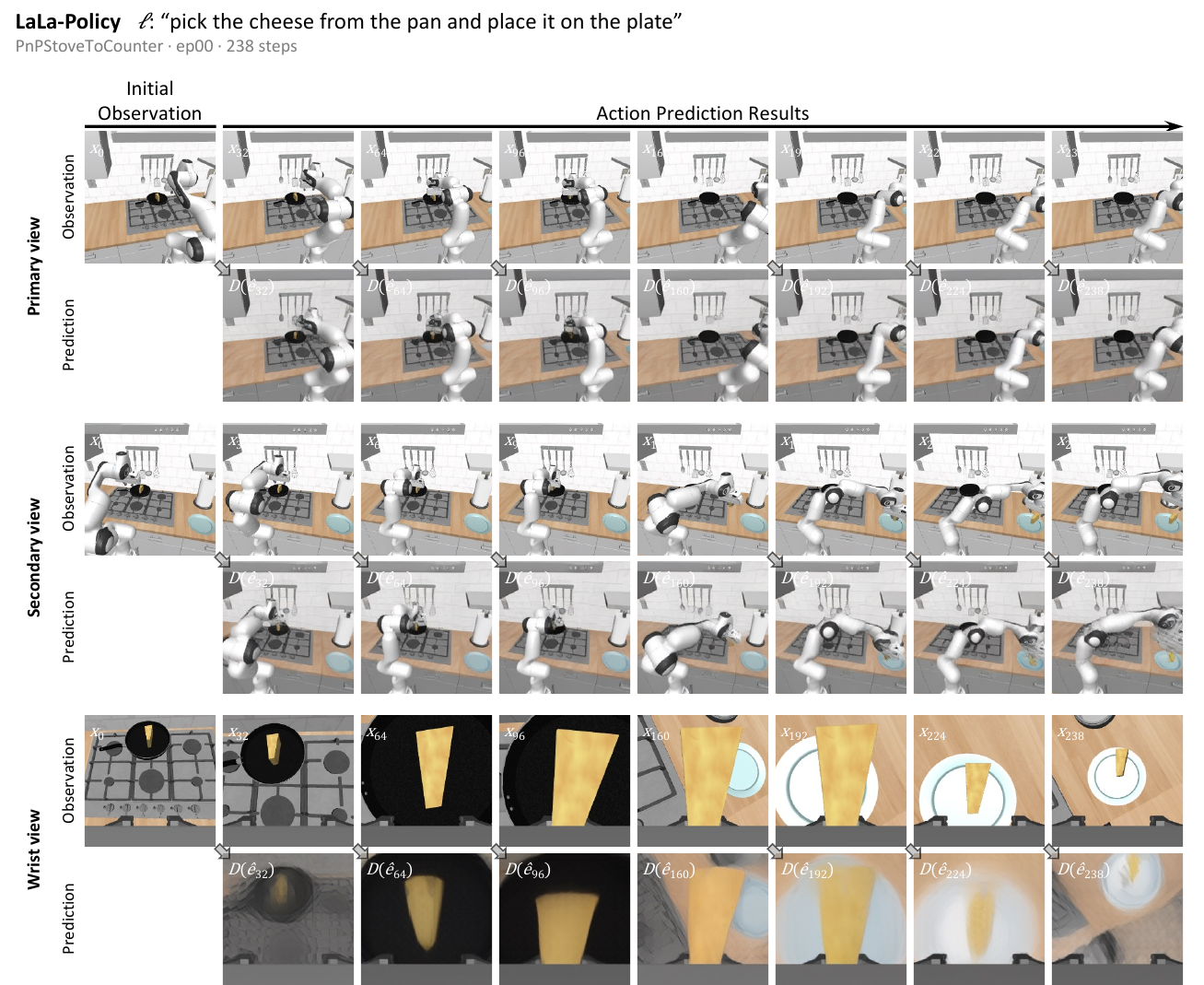}
    \caption{
    \textbf{RoboCasa: pan-to-plate transfer.}
    \ourpolicy\ moves the cheese from the pan onto the plate. Observations and decoded future predictions are shown for two external views and the wrist view.
    }
    \label{fig:policy-robocasa-pick-2}
\end{figure}

%% file: figure-qualitative/policy/robocasa-turn-1.tex
\begin{figure}[!htbp]
    \centering
    \includegraphics[width=.75\linewidth,height=0.88\textheight,keepaspectratio]{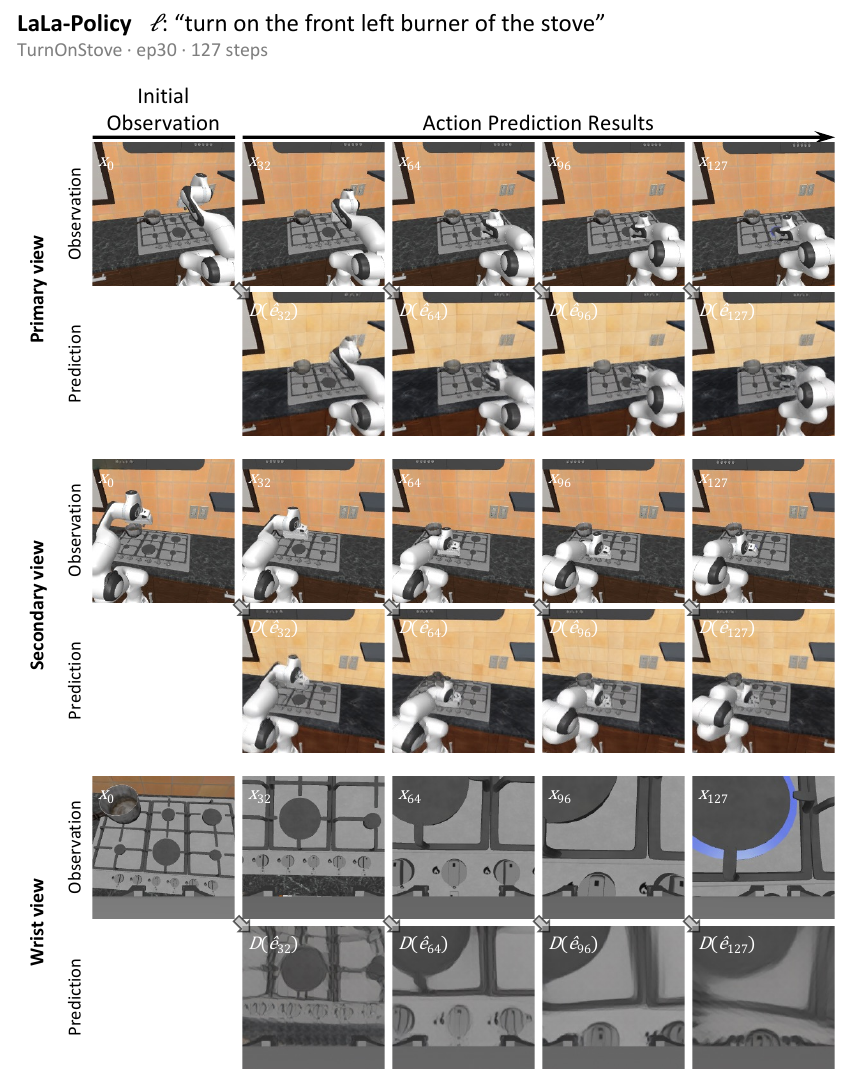}
    \caption{
    \textbf{RoboCasa: turning on a stove burner.}
    Simulator observations and decoded future predictions show \ourpolicy\ turning on the front-left burner, from the primary, secondary, and wrist views.
    }
    \label{fig:policy-robocasa-turn-1}
\end{figure}